\documentclass[10pt]{article} 
\usepackage[preprint]{tmlr}
\usepackage{graphicx}
\usepackage{amsmath}
\usepackage{booktabs}
\usepackage{cite}
\usepackage{comment}
\usepackage{multirow}
\usepackage{rotating}
\usepackage{textcomp} 
\usepackage{gensymb} 

\usepackage{etoolbox,siunitx}
\usepackage{adjustbox}
\usepackage{tabularx}
\usepackage[capitalize,nameinlink]{cleveref}

\usepackage{amsfonts}

\robustify\bfseries

\usepackage{fancyhdr}
\begin{document}
\title{Leveraging Road Area Semantic Segmentation \\ with Auxiliary Steering Task}
%
%
\author{\name Jyri Maanp{\"a}{\"a} \email jyri.maanpaa@nls.fi \\
      \addr Department of Remote Sensing and Photogrammetry, \\ 
      Finnish Geospatial Research Institute FGI, National Land Survey of Finland \\
      Department of Computer Science, \\
      Aalto University, Finland
      \AND
      \name Iaroslav Melekhov \email iaroslav.melekhov@aalto.fi \\
      \addr Department of Computer Science, \\
      Aalto University, Finland
      \AND
      \name Josef Taher \\
      \addr Department of Remote Sensing and Photogrammetry, \\ Finnish Geospatial Research Institute FGI, National Land Survey of Finland \\
      Department of Computer Science, \\
      Aalto University, Finland
      \AND
      \name Petri Manninen \\
      \addr Department of Remote Sensing and Photogrammetry, \\ Finnish Geospatial Research Institute FGI, National Land Survey of Finland
      \AND
      \name Juha Hyypp{\"a} \\
      \addr Department of Remote Sensing and Photogrammetry, \\ Finnish Geospatial Research Institute FGI, National Land Survey of Finland
      }

\maketitle              
\begin{abstract}
    Robustness of different pattern recognition methods is one of the key challenges in autonomous driving, especially when driving in the  high variety of road environments and weather conditions, such as gravel roads and snowfall. Although one can collect data from these adverse conditions using cars equipped with sensors, it is quite tedious to annotate the data for training. In this work, we address this limitation and propose a CNN-based method that can leverage the steering wheel angle information to improve the road area semantic segmentation. As the steering wheel angle data can be easily acquired with the associated images, one could improve the accuracy of road area semantic segmentation by collecting data in new road environments without manual data annotation. We demonstrate the effectiveness of the proposed approach on two challenging data sets for autonomous driving and show that when the steering task is used in our segmentation model training, it leads to a 0.1--2.9\% gain in the road area mIoU (mean Intersection over Union) compared to the corresponding reference transfer learning model. 

\end{abstract}

\section{Introduction}

\thispagestyle{fancy}

One of the main challenges in fully autonomous driving is that there is a huge amount of different roads and weather conditions in which all perception methods should work within an acceptable accuracy. For example, several areas have gravel roads and regular wintertime, in which the road environment looks remarkably different.  
Deep learning based perception methods should also be trained with these conditions for safe operation within use in traffic.

It is relatively easy to collect data from these adverse conditions with cars equipped with sensors. However, it is rather difficult to train the current perception methods without annotated training data. As training data annotation is a laborious task, one could try to tackle this problem with semi-supervised~\citep{mittal2019semi} or weakly supervised~\citep{barnes2017find} learning methods as well as with transfer learning approaches, such as unsupervised domain adaptation~\citep{zou2018unsupervised, zheng2019unsupervised}. These methods benefit from the training data with no annotation increasing the accuracy in the original perception task.

\begin{figure}[!t]
\centering
\includegraphics[width=\linewidth]{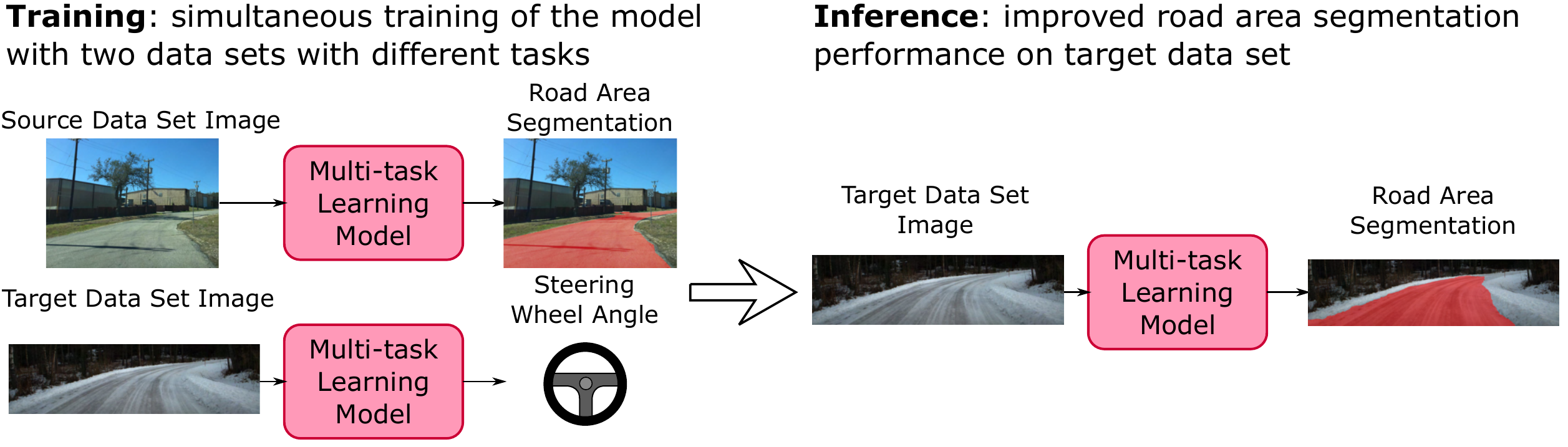}
\caption{A simplified summary of our method. We train a multi-task learning model simultaneously with two data sets which have separate road area segmentation and steering wheel angle prediction tasks. This improves the road area segmentation accuracy in a data set which has steering wheel angle information but no ground truth road area segmentation labels during training.} %
\label{fig:summary}
\end{figure}

One possible solution is to find an auxiliary task which is closely related to the target perception problem and has ground truth labels that can be easily acquired. A multi-task perception model is trained to perform both original and auxiliary tasks on the data set, increasing the accuracy in all tasks if they are related to each other. Autonomous cars have several sensor outputs, such as car control metrics, recorded trajectories, different imaging or lidar sensor modalities, that can be used for auxiliary task learning. If these auxiliary tasks could be used to benefit robust perception model training, 
this has a potential for large scale implementation as cars equipped with sensors could collect this auxiliary task data constantly from all conditions.

In this work, we improve road area semantic segmentation with steering wheel angle prediction as an auxiliary supportive task. We chose to segment the road area as it is one of the most important semantic classes for staying on the road in adverse environmental conditions. Our hypothesis is that the steering wheel angle relates to the road boundaries in a way that the steering task supports the road area segmentation problem. The results demonstrate that the proposed multi-task model has higher accuracy than the corresponding transfer learning baseline. A simplified summary of our method is illustrated in~\cref{fig:summary}.

\section{Related Work}

Semantic segmentation from RGB images is a widely researched topic in pattern recognition and most of the state-of-the-art methods based on convolutional neural networks (CNNs) are also trained and tested on road environment semantic segmentation data sets~\citep{Alhaija2018IJCV,Cordts2016Cityscapes,neuhold2017mapillary}. Some of the recently proposed data sets also include gravel road areas and different weather conditions~\citep{metzger2021fine,pitropov2021canadian}. DeepLab V3 Plus~\citep{chen2018encoder} is one of the most popular segmentation models as a starting point for different experiments, but more complex state-of-the-art models exist, such as the hierarchical multi-scale attention approach by~\citet{tao2005hierarchical}. We refer an interested reader to the great survey on semantic segmentation models by~\citet{lateef2019survey}.

There are several papers on transfer learning in road environment semantic segmentation, mostly related to the simulation-to-real problem. Some of them focus on the adversarial approach, in which a discriminator predicts from the operation of the model if the sample is from the source or from the target data set, thus decreasing the domain gap between the data sets~\citep{Tsai_adaptseg_2018,tsai2019domain}. Others may also include self-learning methods in which the target data set is self-labeled by the trained model to support the fine-tuning training to the target data set~\citep{zou2018unsupervised,zou2019confidence}. Some of the works also apply both approaches~\citep{zheng2019unsupervised}. 

Semantic segmentation accuracy can also be improved by using multi-task learning. These methods are used to 
decrease the computational cost of running several different models for different tasks, but the tasks can also support each other during training and therefore improve the overall accuracy.  
In road environment semantic segmentation, the depth prediction task has been shown to improve the semantic segmentation accuracy~\citep{xu2018pad,sener2018multi,chennupati2019auxnet}. 
There are several approaches to perform multi-task learning as described in the survey of~\citet{vandenhende2021multi}, but the most promising deep multi-task models are based on task-specific encoders in the network architecture, such as MTI-Net~\citep{vandenhende2020mti}.

In this paper our aim is to improve road area semantic segmentation accuracy with steering wheel angle prediction task as a supportive auxiliary task. ~\citet{bojarski2016end,bojarski2020nvidia} proposed a CNN-based approach for steering wheel angle estimation using front camera images. 
In our previous work~\citep{maanpaa2021multimodal} we extended the work by M. Bojarski by utilizing lidar data in addition to camera images and by testing the method in adverse road and weather conditions. 
Work by~\citet{wang2019end} and by~\citet{xu2017end} are relatively close to our method as they improve car control prediction accuracy by utilizing semantic segmentation or object detection as supportive auxiliary tasks. In contrast, our work applies this idea vice versa, as we leverage steering wheel angle estimates to improve road area semantic segmentation performance. 

\section{Method} 

Our model is greatly inspired by MTI-Net~\citep{vandenhende2020mti} - a neural network architecture for multi-task learning and the transfer learning scheme proposed by~\citet{zheng2019unsupervised}. 
MTI-Net is a decoder-based multi-scale task learning model showing remarkable results on dense prediction problems~\citep{vandenhende2021multi}. We utilize the idea of transfer learning setup~\citep{zheng2019unsupervised} since a part of it can be straightforwardly applied in our problem and decreasing the domain gap with transfer learning supports our multi-task approach. It also offers a transfer learning baseline so that the effect of auxiliary steering task could be measured. In this section we describe the details of our implementation and otherwise follow the implementations in~\citet{vandenhende2020mti} and~\citet{zheng2019unsupervised}.

\subsection{Model Architecture}

\begin{figure}[!t]
\centering
\includegraphics[width=\linewidth]{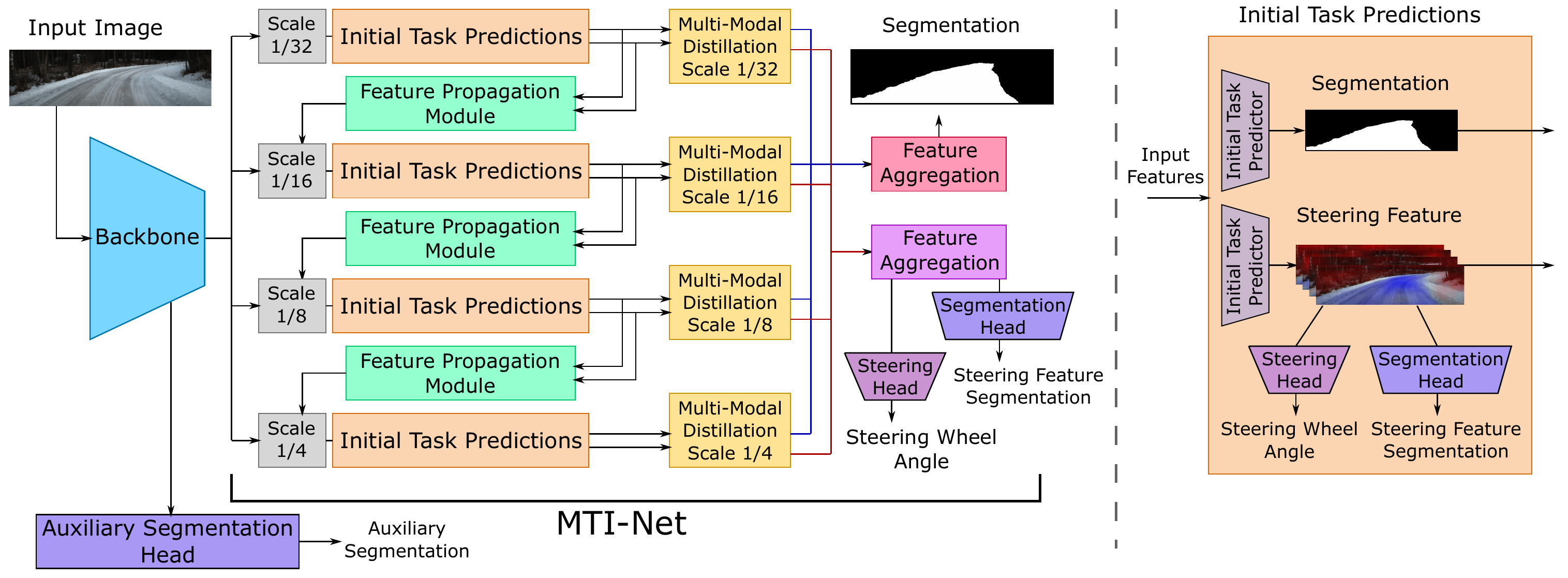}
\caption{Our model based on MTI-Net. On the left is the overall model and on the right a graph of a initial task prediction module. The main difference to the original MTI-Net is that one of the auxiliary tasks is predicting a steering feature, which is used to predict steering wheel angle and steering feature based segmentation in different scales. The auxiliary segmentation output is used for transfer learning purposes.}
\label{fig:model}
\end{figure}

Our model performs two tasks: road area semantic segmentation and steering wheel angle prediction. As presented in~\cref{fig:model}, 
MTI-Net predicts initial estimates for each task output within each scale. This is done using \textit{initial task prediction} modules, each of which uses one feature scale from the backbone network. 
The sum of these loss predictions for task~$t$ is called deep supervision loss~$L_{Deep,t}$.

The initial task predictions are converted to task-specific features with \textit{feature propagation modules} and forwarded to the next higher scale in the network structure. This provides information for the next scale initial task predictions so that the predictions could adapt to different task features obtained from the previous scale. Next, the task predictions from each scale are fused with \textit{multi-modal distillation} modules that impose attention for initial task predictions based on all other task outputs. Finally, the output from different scales is combined with \textit{feature aggregation} modules to obtain final output predictions. 

However, the steering wheel angle prediction is not an image-spatial task like semantic segmentation as steering wheel angle is a scalar value. Therefore, the initial task prediction module actually outputs a steering feature instead of a steering wheel angle. This feature has the same width and height as the initial prediction of road area segmentation so that they could be fused together. When evaluating deep supervision loss, we apply a corresponding steering head in each of the steering features to predict an actual steering wheel angle. A steering head is also applied in the steering task feature aggregation module output to obtain a final steering wheel angle prediction.

The steering feature could be seen as a prior of the road shape, which is used in the steering. However, we noticed in our initial experiments that the steering feature could not correspond to road area well if it is only directly affected by the steering loss leading to a lower accuracy. Therefore, we also applied separate segmentation heads to each steering feature in the model to produce road area segmentations based on them. Within this way steering features correspond more to road area, even though they mostly contain information for steering prediction. This increases the effectiveness of the transfer from the steering task to segmentation. We call this particular segmentation output as steering feature segmentation in this work.

In our experiments, we decided to use 4 steering feature channels based on our initial tests. Examples of these feature masks are presented in the supplementary material, 
which also explains the architectures of the steering heads and other specific details of the model. 

Note that as the input samples have images and either segmentation or steering ground truth depending on which data set gives the sample, the model is not trained simultaneously to both tasks with a same image. Each input batch during training contain samples from both data sets and the segmentation and steering losses are evaluated depending on the sample type.

\subsection{Transfer Learning Scheme}

Our transfer learning scheme adopted the adversarial domain adaptation and memory regularisation approaches that were introduced in the work of~\citet{zheng2019unsupervised}.  
Our method resembles the 'Stage-I' of the original work~\citep{zheng2019unsupervised} as we omitted the presented self-learning approach. We consider the Mapillary Vistas data set with road area segmentation masks as the source data set and the FGI autonomous steering data set from the work of~\citet{maanpaa2021multimodal} as the target data set.

As a part of the transfer learning method, we predict an auxiliary segmentation output with a segmentation head, that operates on the three highest-resolution features from the backbone. This auxiliary segmentation head has otherwise similar architecture to the segmentation feature aggregation module. 

We train two discriminators that operate on the primary segmentation output from the MTI-Net and on the auxiliary segmentation output, trying to separate the samples from different data sets from each other. These impose adversarial losses on the segmentation outputs from both segmentation heads. This should improve the model operation on the target set with no ground truth segmentation. The discriminators are trained similarly as in the previous work of~\citet{zheng2019unsupervised}, with the exception that the 'real' output used in the real-fake comparison is the primary segmentation on the source data set. The primary segmentation output is also used as the segmentation in the validation and test set results. 

\begin{figure}[!t]
\centering
\makebox[\textwidth][c]{
\includegraphics[width=1.0\linewidth]{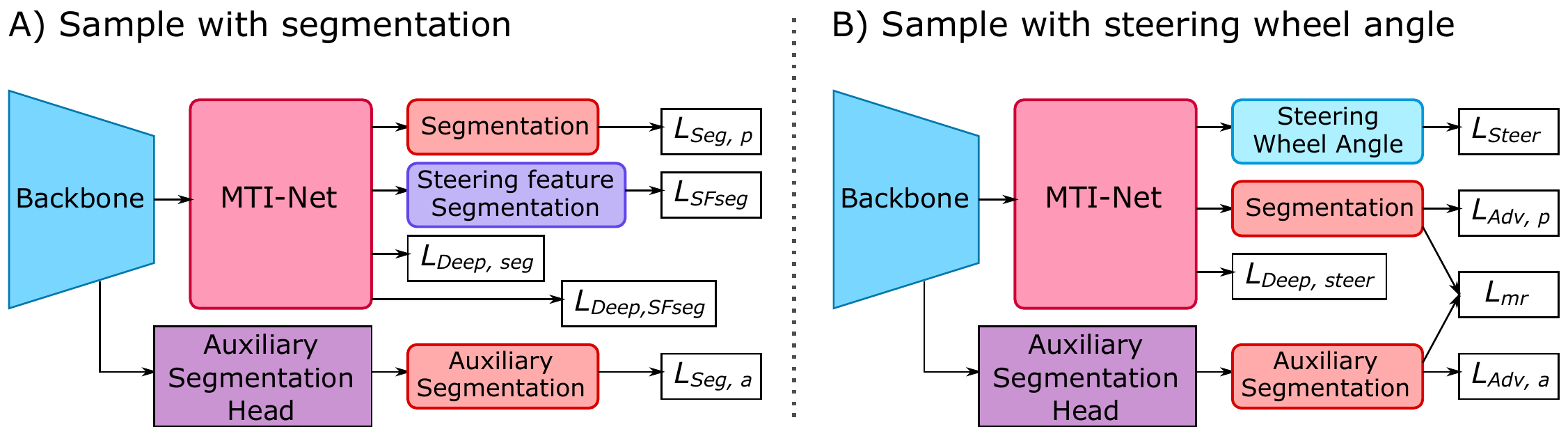}
}
\caption{Our overall training scheme. When the sample is from the Mapillary data set with segmentation (case A) we evaluate the segmentation loss for primary, auxiliary and steering feature segmentation outputs and similarly the deep supervision loss for all segmentations in different MTI-Net scales. When the sample is from the FGI autonomous steering data set with steering wheel angle (case B) we evaluate the primary steering loss, deep supervision loss for steering, adversarial losses for primary and auxiliary segmentation outputs and memory regularisation loss between primary and auxiliary segmentation outputs.}
\label{fig:transfer}
\end{figure}

\Cref{fig:transfer} shows our overall training scheme for a single batch that has samples from both of our data sets. Different loss functions are evaluated on the model, depending on the input data set. 
When the input sample has a road area segmentation mask, we evaluate the binary cross entropy loss for all segmentation outputs in the model: the primary segmentation loss~($L_{Seg,p}$), the auxiliary segmentation loss~($L_{Seg,a}$), and steering feature based segmentation loss~($L_{SFseg}$). The same loss is also applied for the deep supervision, in which the overall loss is the sum of segmentation losses in each initial prediction scale ($L_{Deep,seg}$ for initial segmentation prediction and $L_{Deep,SFseg}$ for initial steering feature based segmentation). 
These losses are fully supervised as segmentation masks are available for the Mapillary Vistas data set.

When the input sample is from the FGI autonomous steering data set, 
we evaluate the mean square error of the steering wheel angle prediction~$L_{Steer}$ and a deep supervision loss~$L_{Deep,steer}$ that is a sum of the mean squared errors evaluated on initial steering predictions from different scales. Following the current multi-scale discriminator approach in the source code of~\citep{zheng2019unsupervised}, the predicted segmentation outputs are supervised with the following LSGAN~\citep{mao2017least} adversarial loss, applied only for the target set:
\begin{equation}
    L_{adv,i} = 2\cdot\mathbb{E}\left[(D_i(F_i(x_t))-1)^2\right].
\end{equation}
Here~$x_t$ is the target input image batch, $F_i$~is the segmentation head output model (primary or auxiliary) and $D_i$~is the corresponding discriminator model. Therefore we get the adversarial losses $L_{adv,p}$ for the primary segmentation from the MTI-Net and $L_{adv,a}$ for the auxiliary segmentation.

In addition, we use the memory regularisation loss as presented in~\citep{zheng2019unsupervised}, which is the pixel-wise KL-divergence loss:
\begin{align}
    L_{mr} = &-\sum_{h=1}^H\sum_{w=1}^W F_a(x_t)\log (F_p(x_t)) - \sum_{h=1}^H\sum_{w=1}^W F_p(x_t)\log (F_a(x_t)).
\end{align}
Here the sums act pixel-wise and $F_p$ and $F_a$ produce the primary and secondary segmentation masks from the target set input image~$x_t$. This loss enforces the model to be consistent between primary and auxiliary predictions in the target set, also acting in a 
self-supervising manner.

As a summary, our loss in the source data set with segmentation is
\begin{align}\label{eq:source_loss}
    L_{source} = & L_{Seg,p} + \lambda_{aux} L_{Seg,a} + \lambda_{deep} L_{Deep,seg} \\
    & + \lambda_{SFseg}\left( L_{SFseg} + \lambda_{deep} L_{Deep,SFseg}\right) \nonumber
\end{align}
and in the target set with steering wheel angles the loss is
\begin{align}\label{eq:target_loss}
    L_{target} = & \lambda_{steer}\left( L_{Steer} + \lambda_{deep} L_{Deep,steer} \right) \\
    &+ \lambda_{adv,p}L_{adv,p} + \lambda_{adv,a}L_{adv,a} + \lambda_{mr}L_{mr}. \nonumber
\end{align}

Here $\lambda_{aux}$, $\lambda_{deep}$, $\lambda_{SFseg}$, $\lambda_{steer}$, $\lambda_{adv,p}$, $\lambda_{adv,a}$, and $\lambda_{mr}$ are the weighting coefficients for the auxiliary segmentation, deep supervision, steering feature segmentation, steering, adversarial for the primary and auxiliary segmentation, and the memory regularisation loss respectively. The final loss is the sum of these losses, as each mini-batch contains data from both source and target data sets.

\section{Experiments}

\subsection{Data Sets}

We use two data sets: Mapillary Vistas~\citep{neuhold2017mapillary} and a data set for autonomous steering in adverse weather conditions from our previous work~\citep{maanpaa2021multimodal}, called the FGI autonomous steering data set in this work. The Mapillary data set has multi-class ground truth segmentations for diverse driving scenarios and the FGI autonomous steering data set has 28 hours of camera image sequences with corresponding steering wheel angles in a variety of road and weather conditions. We combined drivable-area classes to one road area class in the Mapillary data set. 
After preprocessing, the Mapillary data set has 17074 samples and the FGI autonomous steering data set has 990436 samples. More information about the data set preprocessing is provided in the supplementary material. We also annotated the drivable area in 100 images for model validation and 100 images for performance testing in the FGI autonomous steering data set. Most of these samples focus on winter conditions, there are also samples from gravel roads and during night. 
No training samples are used within 5~seconds before and after each validation or test set sample in the data set.

\subsection{Training Setup}

We trained three models for a performance evaluation: {\em (1)}~\textbf{single-task (ST) model}: trained only with the segmentation task on Mapillary data set; {\em (2)}~\textbf{transfer learning (TL) model}:  utilizes the transfer learning setup, MTI-Net and deep supervision on segmentation and steering feature segmentation, but is not trained with the steering task; {\em (3)}~\textbf{our multi-task model}: similar to {\em (2)} but is also supervised to predict the steering wheel angle, both with primary output and with deep supervision.

The similar structure of the last two models allow the investigation of the additional impact of the auxiliary steering task. Although the steering segmentation loss is evaluated in the transfer learning model, all actual steering related losses are not evaluated during its training. 

We repeated the experiments with two backbones: High-Resolution Network (HRNet)~\citep{wang2020deep} and Feature Pyramid Network~\citep{lin2017feature} applied to ResNet features (FPN ResNet) as in~\citep{vandenhende2020mti}. This was done to confirm the increase in accuracy due to the auxiliary steering task. 
In addition, we train a stage I model from~\citep{zheng2019unsupervised} as a comparison to MTI-Net based architectures, with the exceptions that our implementation has ordinary ResNet instead of dilated ResNet and does not have a dropout layer. For loss function in~\cref{eq:source_loss} and~\cref{eq:target_loss} 
we used the weights $\lambda_{aux}=0.5$, $\lambda_{deep}=1.0$, $\lambda_{SFseg}=0.3$, $\lambda_{steer}=0.5$, $\lambda_{adv,p}=0.001$ and $\lambda_{adv,a}=0.0002$. The memory regularisation loss was applied after 15000 first training steps with $\lambda_{mr}=0.1$. Otherwise our training setup mostly follows the work in~\citep{zheng2019unsupervised}, more details on model training are provided in the supplementary.

\subsection{Results}

The performance of the models is evaluated on validation and test splits from the FGI autonomous steering data set. The results are reported as road area mean Intersection-over-Union (mIoU) values which mean the fraction of the intersection and union between the predicted and actual road area pixels, averaged over validation or test set samples. The results show that although our multi-task model performs best with both backbones, the difference to the transfer learning baseline is not always significant. This is especially seen in the test set mIoU of the HRNet backbone models, in which the steering task causes less than 0.1\% increase in mIoU. However, the multi-task model with FPN ResNet backbone has a 0.9\% increase in test set mIoU to the HRNet transfer learning model and even a 2.9\% increase to the corresponding FPN ResNet transfer learning model~(\cref{tab:valtest}).

\begin{table}[t!]
\fontsize{9}{10}\selectfont
\centering
\renewcommand{\tabcolsep}{10pt}

\caption{We report road area semantic segmentation performance on the validation and test set in terms of mIoU, precision ($\mathcal{P}$), and recall ($\mathcal{R}$) for different models and backbones. The proposed approach outperforms other methods by a noticeable margin.}
\label{tbl:test}
\begin{tabularx}{\textwidth}{@{} l | c | S[table-format=2.2,detect-weight,detect-shape,detect-mode] S[table-format=2.2,detect-weight,detect-shape,detect-mode] S[table-format=2.2,detect-weight,detect-shape,detect-mode] | S[table-format=2.2,detect-weight,detect-shape,detect-mode] S[table-format=2.2,detect-weight,detect-shape,detect-mode] S[table-format=2.2,detect-weight,detect-shape,detect-mode]}
\toprule
 \multirow{2}{*}{Model} & \multirow{2}{*}{Backbone} & \multicolumn{3}{c|}{Validation} & \multicolumn{3}{c}{Test} \\
 & & {mIoU} & {$\mathcal{P}$} & {$\mathcal{R}$} & {mIoU} & {$\mathcal{P}$} & {$\mathcal{R}$}\\
 \midrule
 {\citet{zheng2019unsupervised}} & {ResNet-18} & 85.96 & 92.05 & 92.67 & 87.18 & 94.06 & 92.47 \\
 \midrule
 {Single Task (ST)} & \multirow{3}{*}{HRNetV2-W18} & 84.70 & 88.21 & \bfseries 94.28 & 87.43 & 91.73 & 94.64 \\
 {Transfer Learning (TL)} & & 89.04 & 95.45 & 92.70 & 89.71 & \bfseries 97.36 & 92.19 \\
 {Proposed} & & \bfseries 90.44 & \bfseries 95.68 & 94.06 & \bfseries 89.79 & 97.21 & \bfseries 92.22 \\
 \midrule
 {Single Task (ST)} & \multirow{3}{*}{FPN ResNet-18} & 80.13 & 84.62 & 92.99 & 83.84 & 87.56 & \bfseries 94.56 \\
 {Transfer Learning (TL)} & & 87.11 & 93.80 & 92.25 & 87.70 & 94.97 & 92.01 \\
 {Proposed} & & \bfseries 88.92 & \bfseries 94.12 & \bfseries 93.94 & \bfseries 90.61 & \bfseries 95.92 & 94.38 \\
 
 \bottomrule
\end{tabularx}\label{tab:valtest}
\end{table}

The difference between validation and test set results is most probably caused by the small amount of data in validation and test sets. Each set contains 100 sample images and one poorly classified sample leads to almost 1\% difference in the mIoU result. It is also possible that the model development process overfitted the HRNet based multi-task model to the validation set with hyperparameter and architecture decisions. In addition, the interpretation of the road area boundaries is often ambiguous even for humans, and therefore it is difficult to observe performance differences when the mIoU accuracy is close to 90\%. However, when considering both validation and test set results with both backbones, we can argue that the steering task supports the road area segmentation task, as the mIoU increases systematically and notably when utilizing steering task with the FPN ResNet backbone. We also conclude that using MTI-Net in the transfer learning baseline is justified as the reference transfer model ('Stage-I' from the work of~\citet{zheng2019unsupervised}) has smaller or equal accuracy to the FPN ResNet based transfer learning model with a similar ResNet part.
Example road area segmentations for each model with FPN ResNet backbone can be seen in~\cref{fig:segmentation_examples2}. 
We observe that utilizing steering information makes the segmentation mask follow the road borders more accurately, reducing false positive area outside the road. The segmentation is also more accurate far away and does not contain gaps.

\section{Conclusion}

We trained a multi-task model to perform both road area segmentation and steering wheel angle prediction in order to transfer road area segmentation performance to a new data set with adverse road and weather conditions. 
We found out that when the steering task is used in the model training, we gained a 0.1--2.9\% percentage point increase in the road area mIoU when compared to the corresponding reference transfer learning model, reaching 90.6\% mIoU in the test set with our best multi-task model. 
This means that steering wheel angle prediction could be used as an useful auxiliary task for improving road area segmentation in a new data set which only has steering wheel angle ground truth. However, there can be plenty of random variation in the results due to small test data set size. Furthermore, one should note that our approach is mostly tested in rural environment and more experiments should be made to optimize the performance of this multi-task setup and to implement it to a multi-class semantic segmentation case. One should also consider if there are other auxiliary tasks that could support road area segmentation more effectively and if there is a limit in which higher road area segmentation accuracy could not be reached due to road area interpretation ambiguity.

\newcommand{\figWidthFinal}{0.26}

\begin{figure}[t!]
\vspace{-15mm}
\centering
\begin{adjustbox}{angle=90}
\begin{tabular}{ccccc} 

Original image & Ground truth & ST model & TL model & \textbf{Proposed model} \\
\includegraphics[width=\figWidthFinal\linewidth]{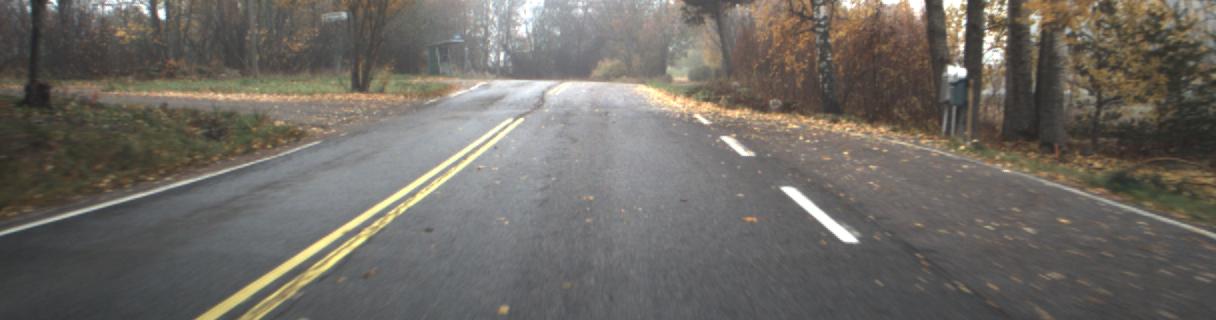} & \includegraphics[width=\figWidthFinal\linewidth]{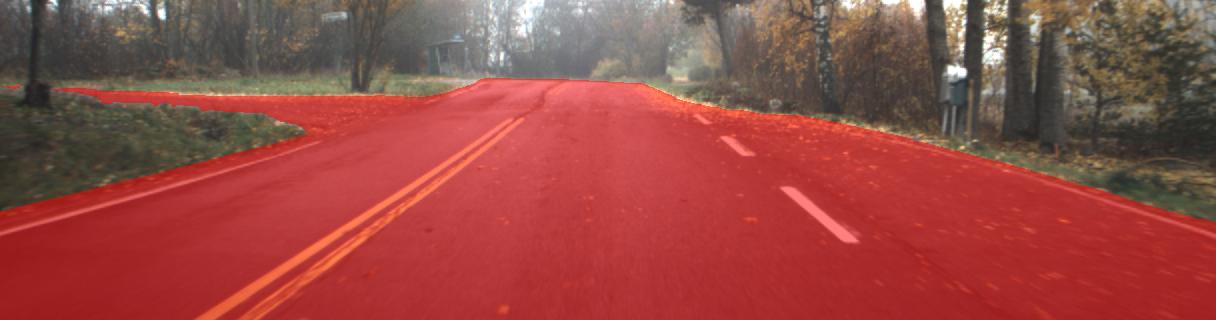} & \includegraphics[width=\figWidthFinal\linewidth]{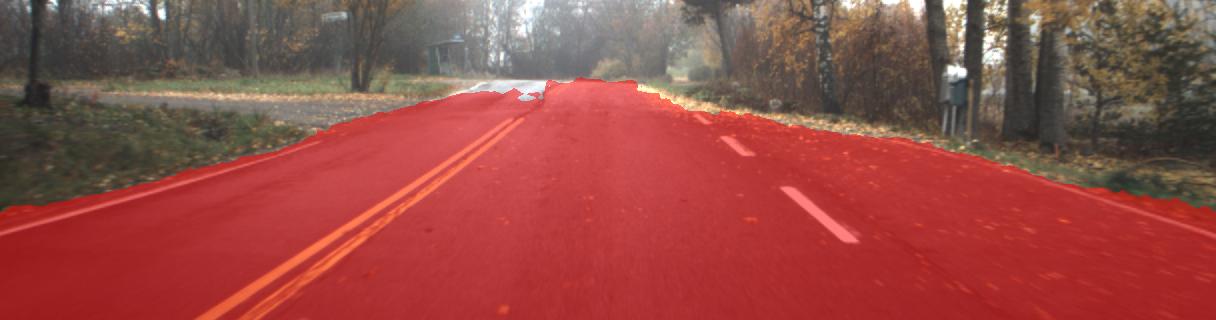} & \includegraphics[width=\figWidthFinal\linewidth]{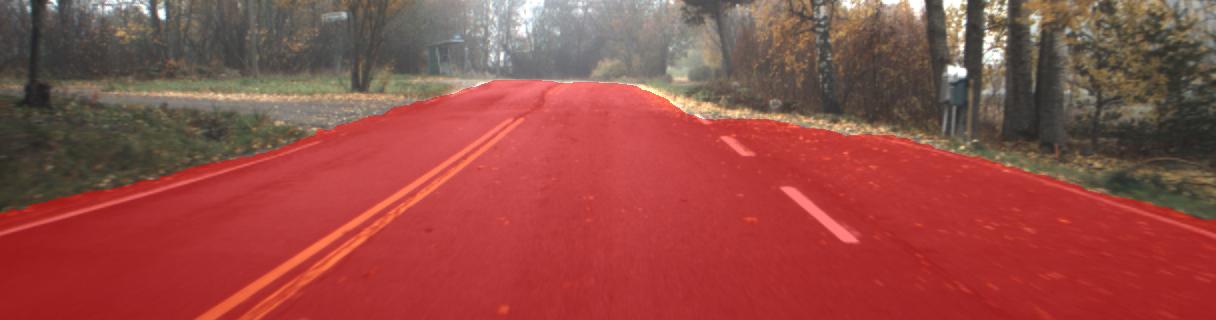} & \includegraphics[width=\figWidthFinal\linewidth]{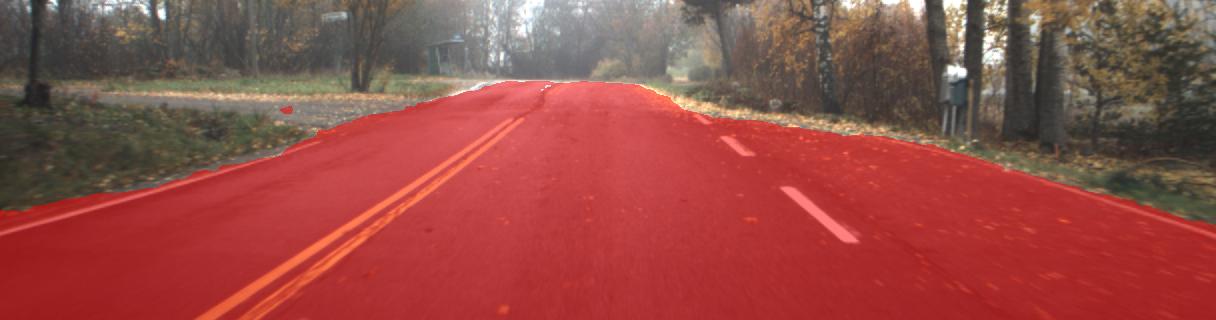} \\
\includegraphics[width=\figWidthFinal\linewidth]{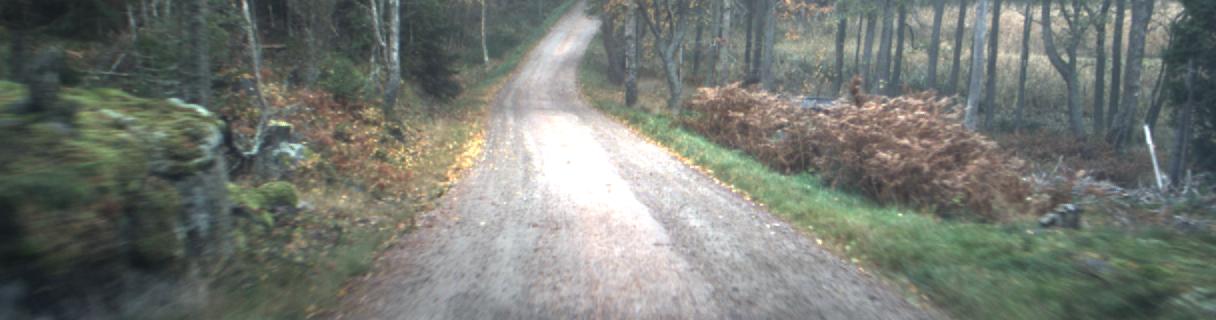} & \includegraphics[width=\figWidthFinal\linewidth]{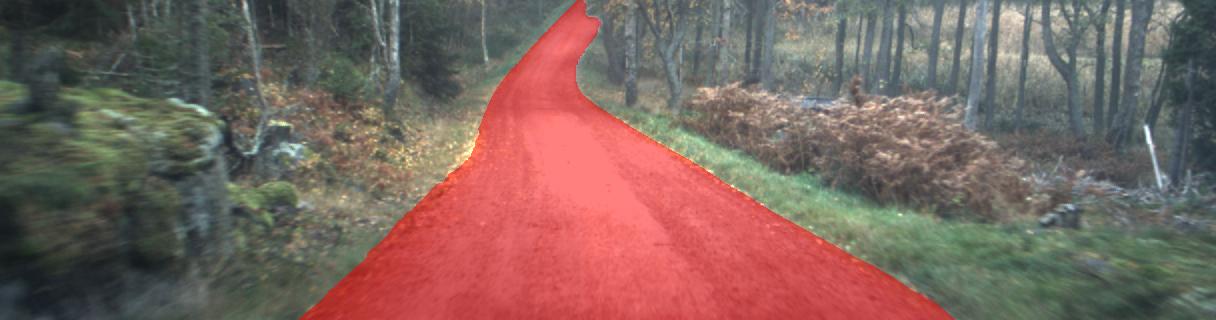} & \includegraphics[width=\figWidthFinal\linewidth]{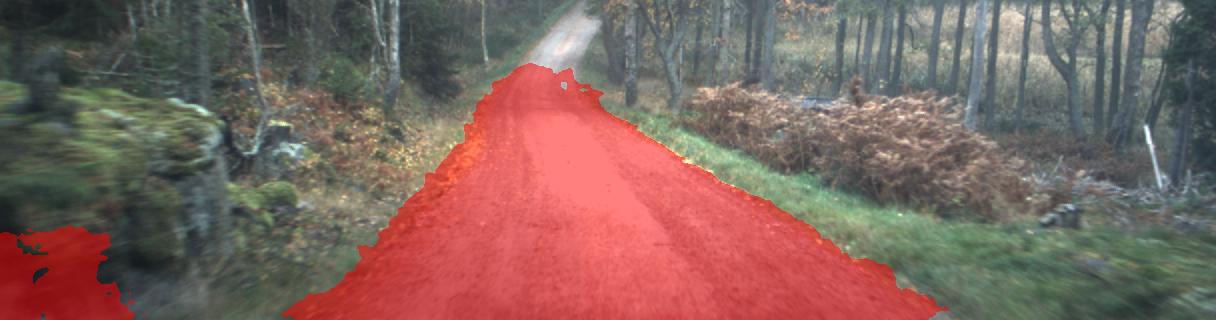} & \includegraphics[width=\figWidthFinal\linewidth]{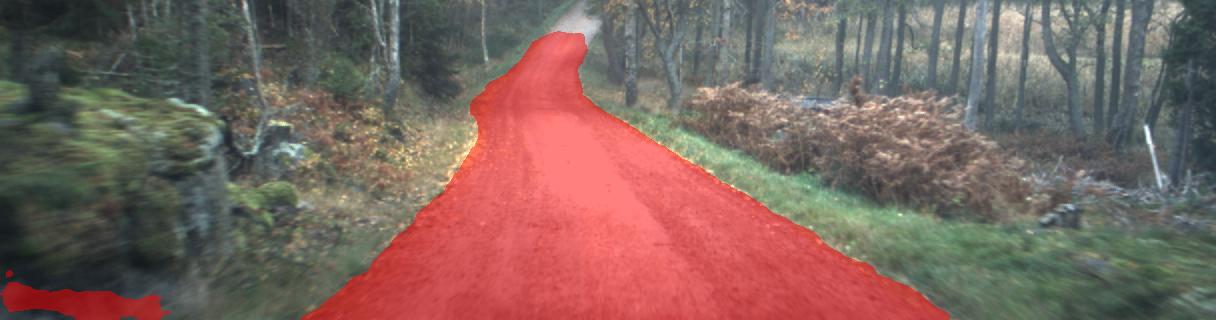} & \includegraphics[width=\figWidthFinal\linewidth]{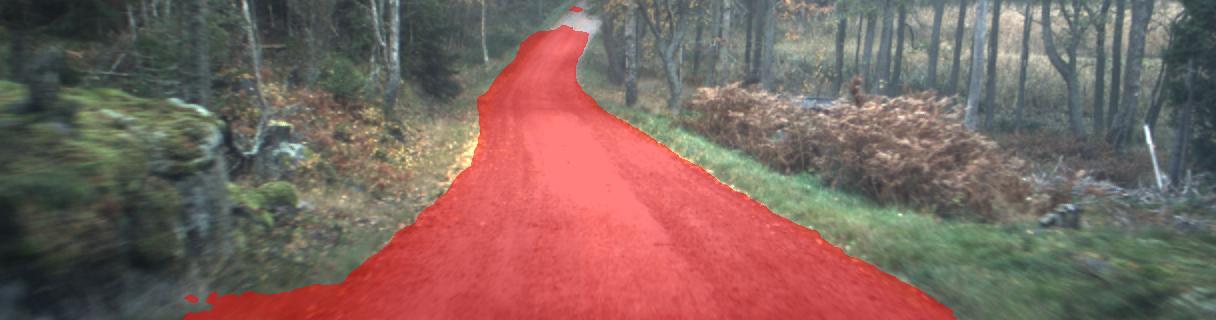} \\
\includegraphics[width=\figWidthFinal\linewidth]{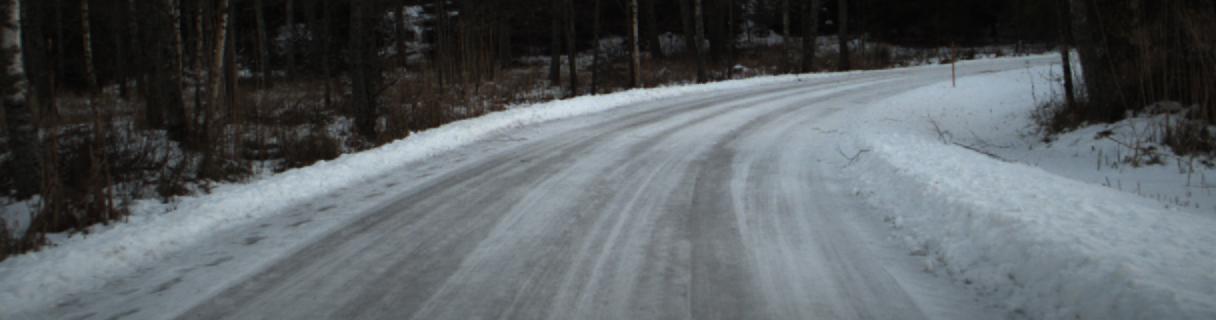} & \includegraphics[width=\figWidthFinal\linewidth]{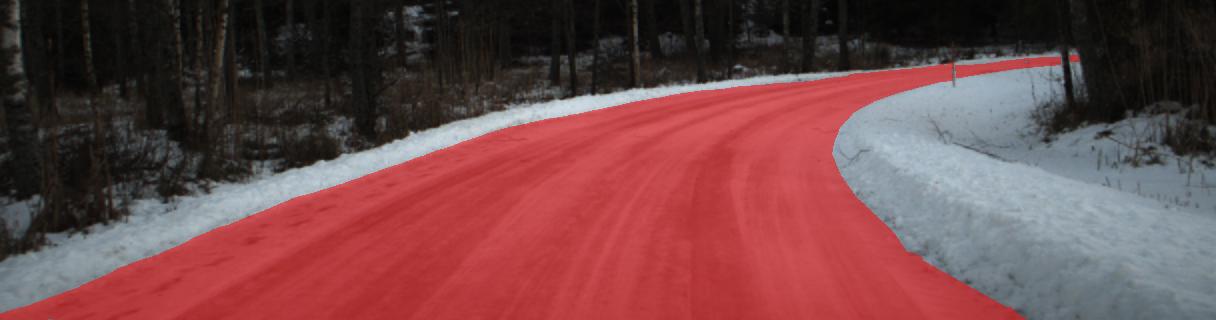} & \includegraphics[width=\figWidthFinal\linewidth]{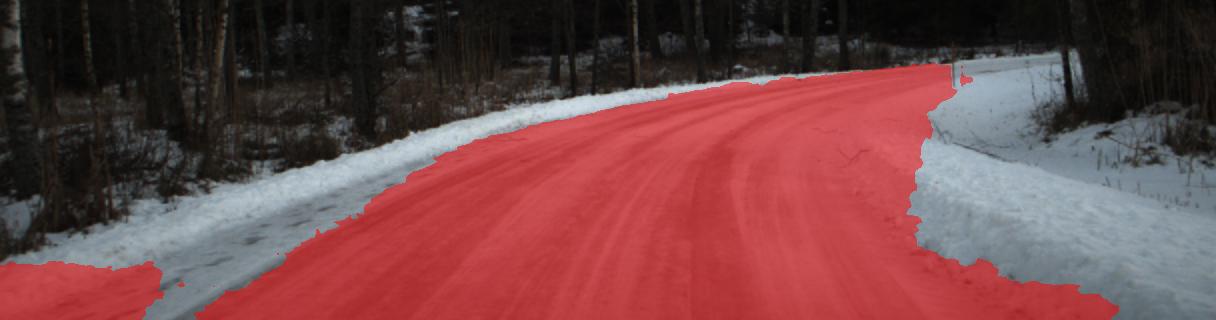} & \includegraphics[width=\figWidthFinal\linewidth]{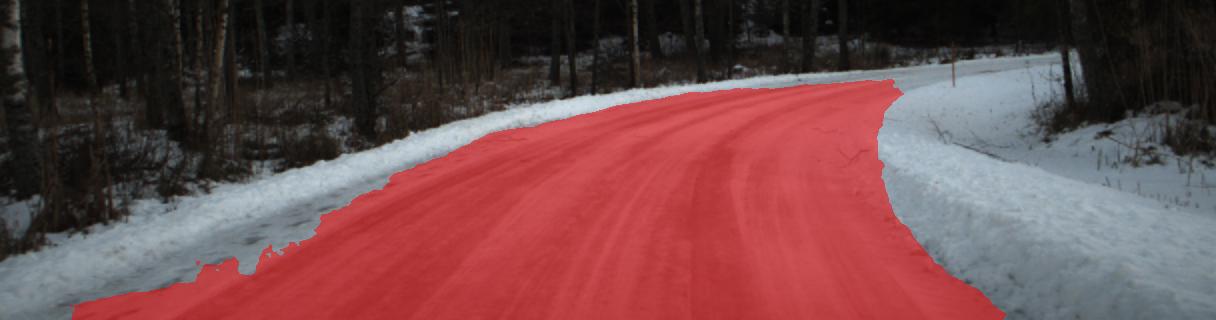} & \includegraphics[width=\figWidthFinal\linewidth]{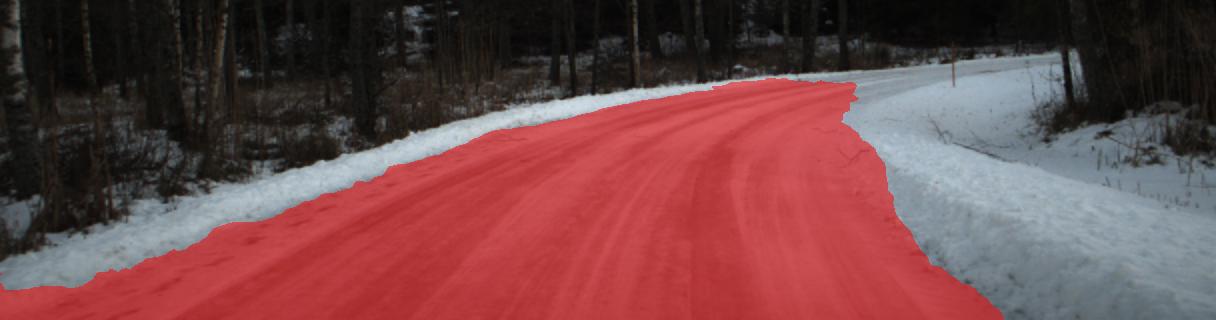} \\
\includegraphics[width=\figWidthFinal\linewidth]{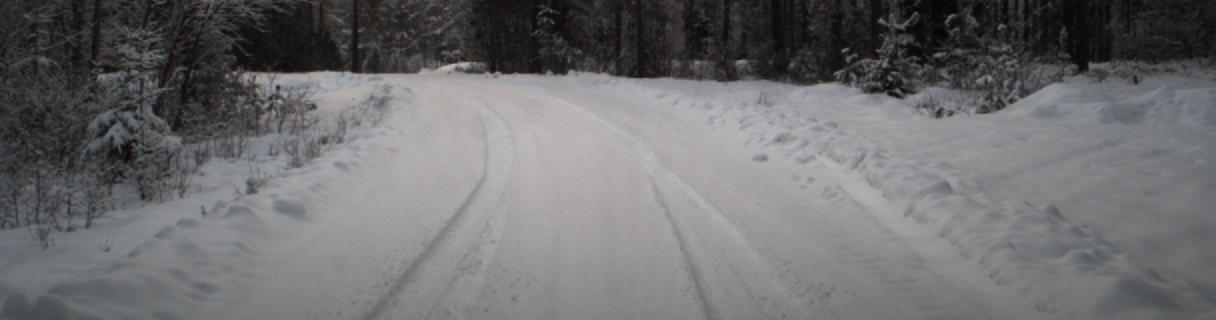} & \includegraphics[width=\figWidthFinal\linewidth]{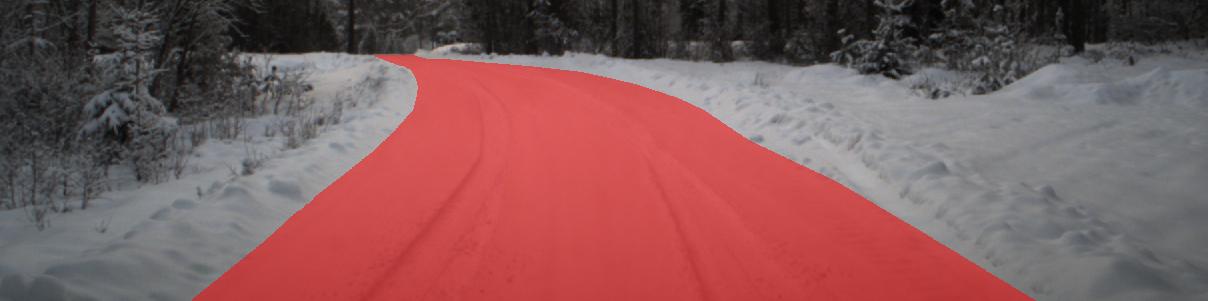} & \includegraphics[width=\figWidthFinal\linewidth]{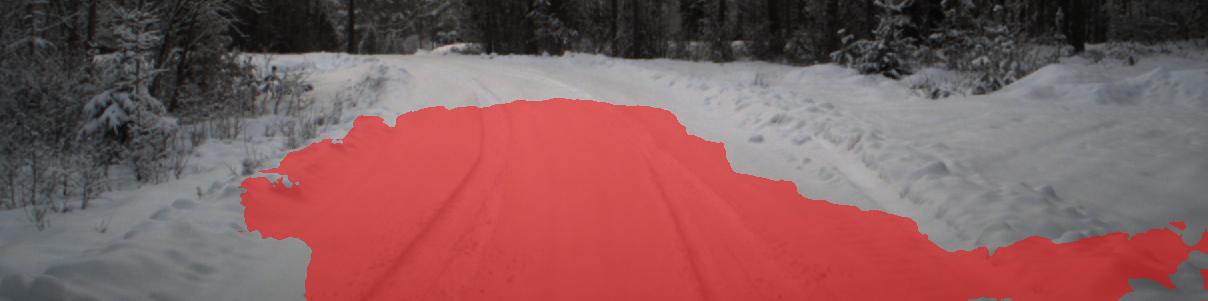} & \includegraphics[width=\figWidthFinal\linewidth]{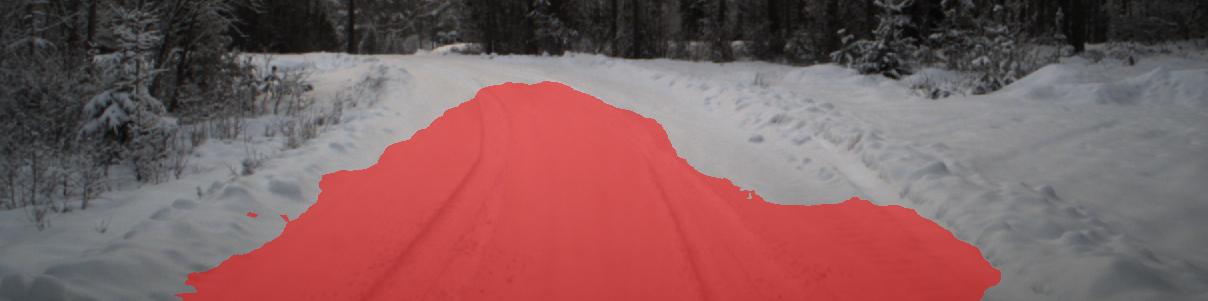} & \includegraphics[width=\figWidthFinal\linewidth]{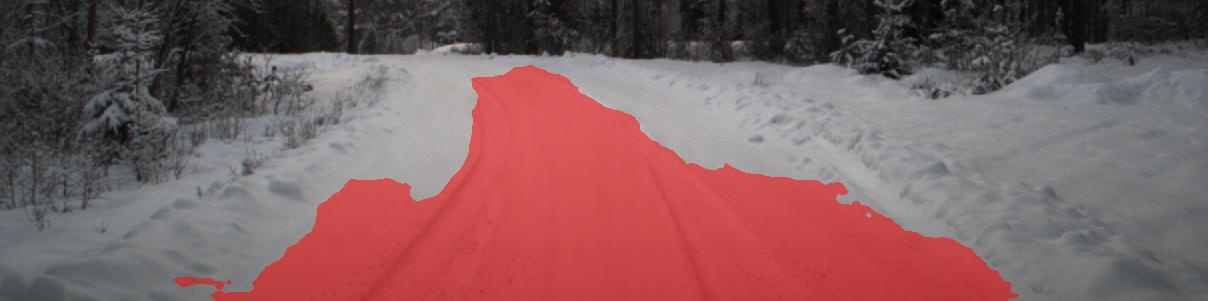} \\
\includegraphics[width=\figWidthFinal\linewidth]{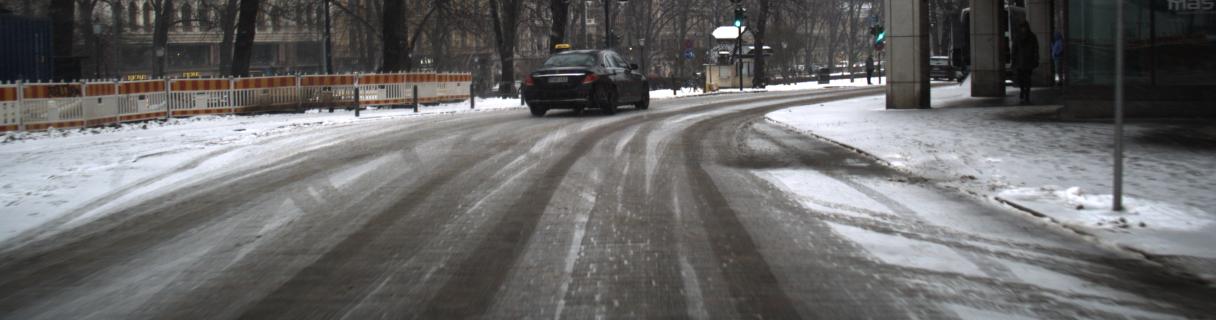} & \includegraphics[width=\figWidthFinal\linewidth]{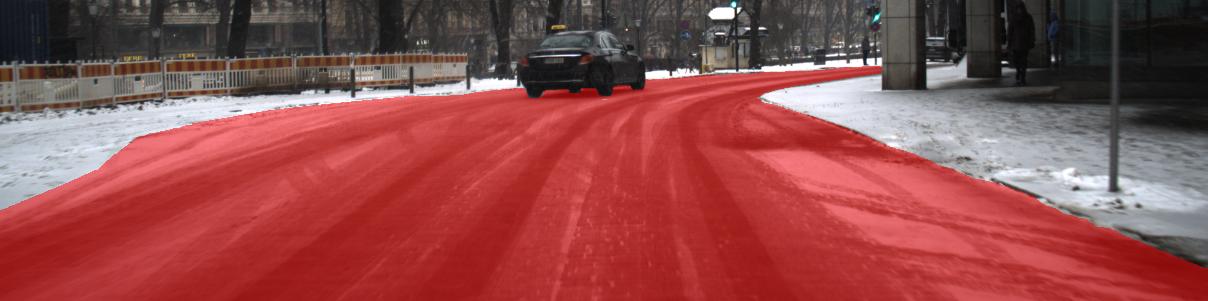} & \includegraphics[width=\figWidthFinal\linewidth]{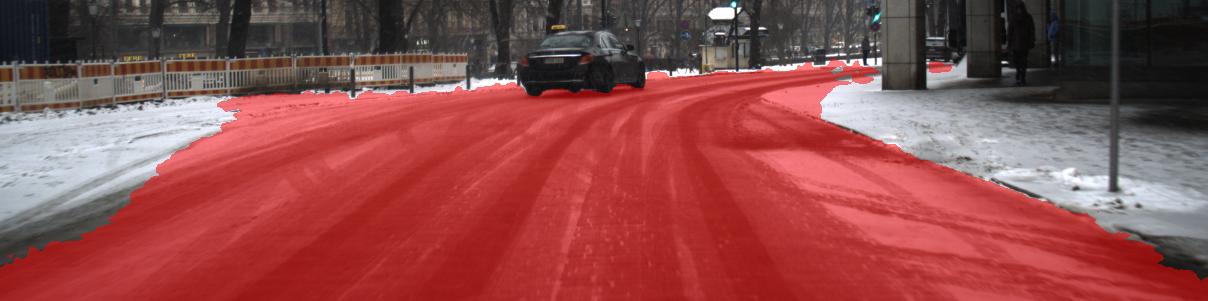} & \includegraphics[width=\figWidthFinal\linewidth]{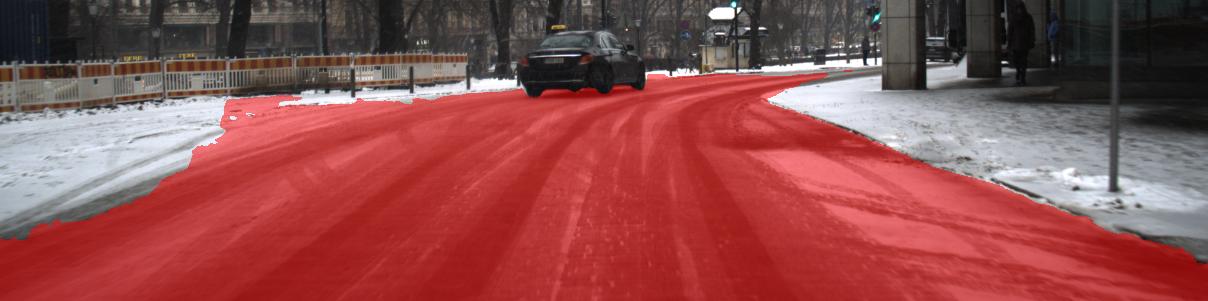} & \includegraphics[width=\figWidthFinal\linewidth]{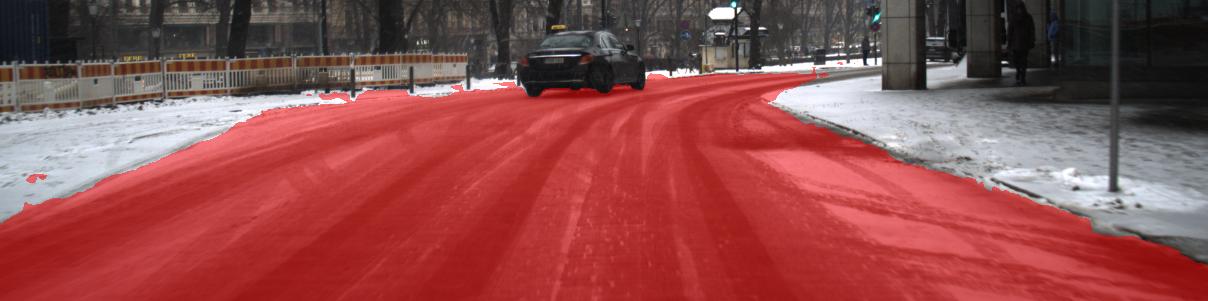} \\
\includegraphics[width=\figWidthFinal\linewidth]{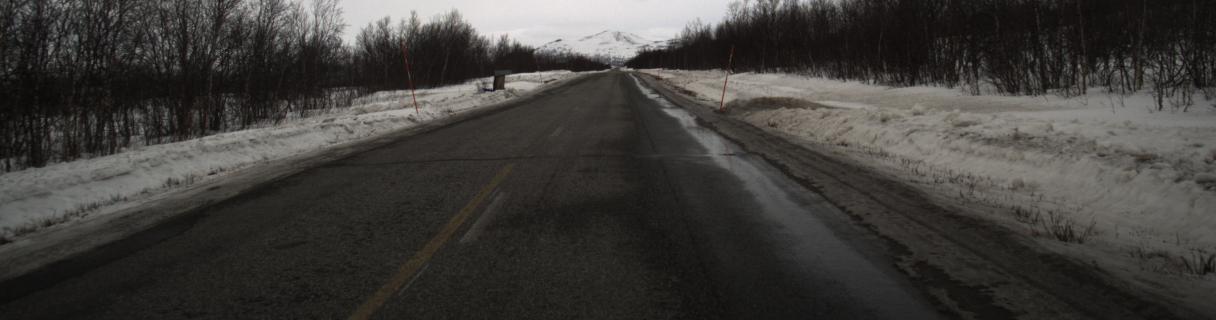} & \includegraphics[width=\figWidthFinal\linewidth]{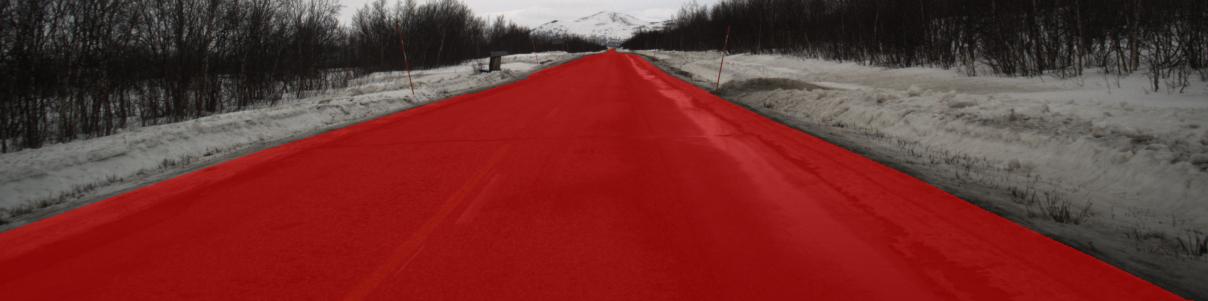} & \includegraphics[width=\figWidthFinal\linewidth]{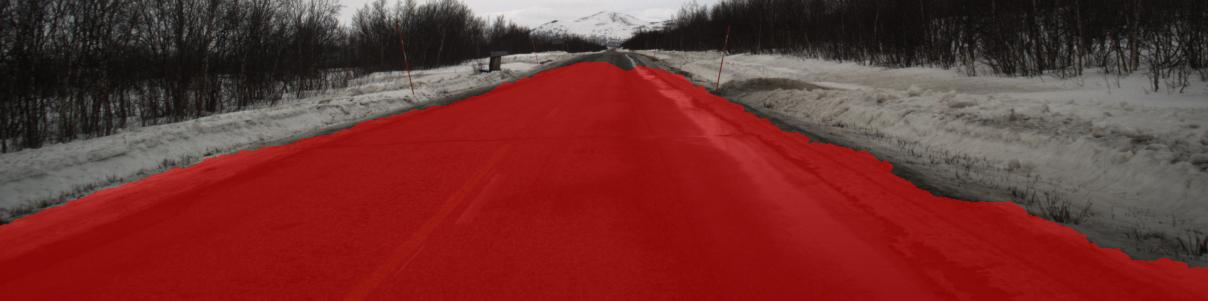} & \includegraphics[width=\figWidthFinal\linewidth]{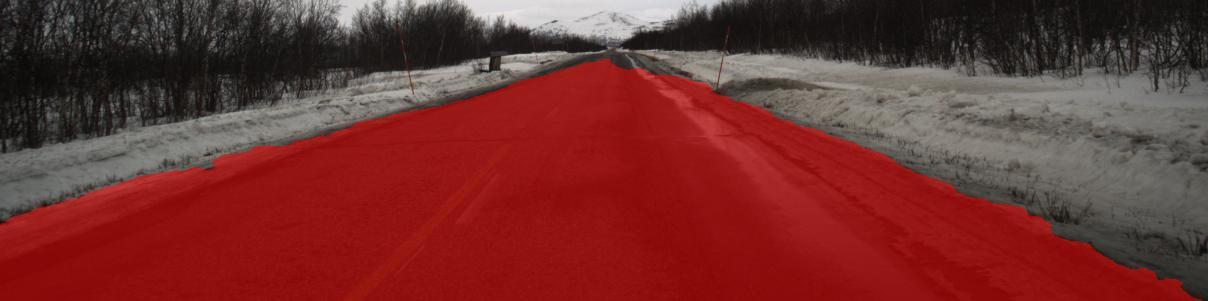} & \includegraphics[width=\figWidthFinal\linewidth]{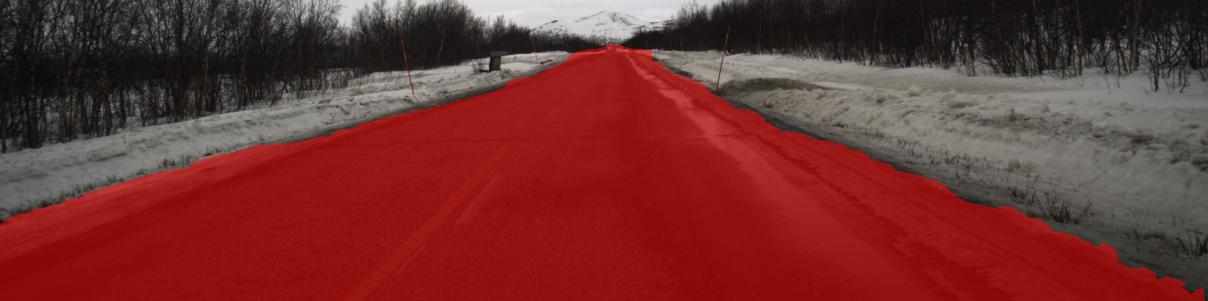} \\
\includegraphics[width=\figWidthFinal\linewidth]{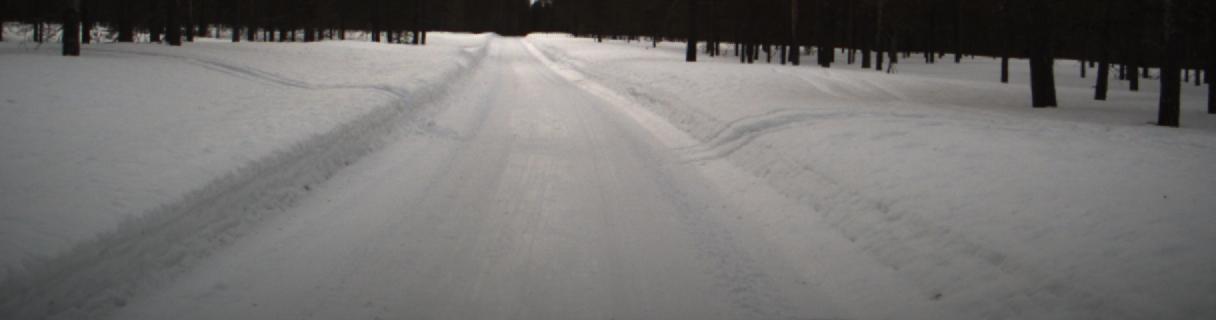} & \includegraphics[width=\figWidthFinal\linewidth]{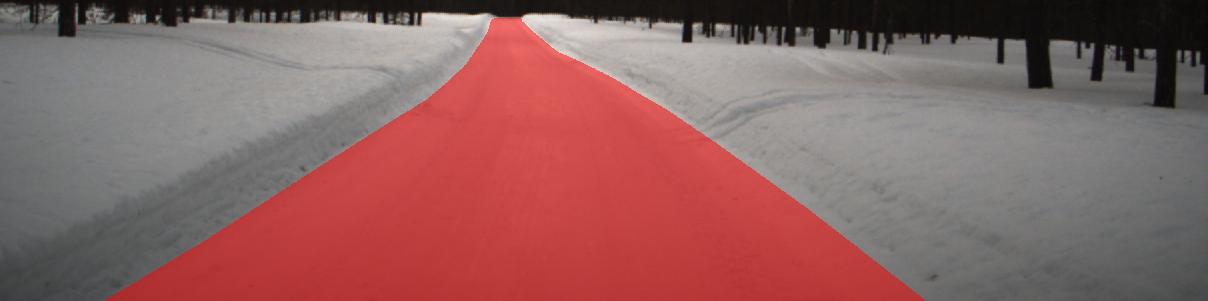} & \includegraphics[width=\figWidthFinal\linewidth]{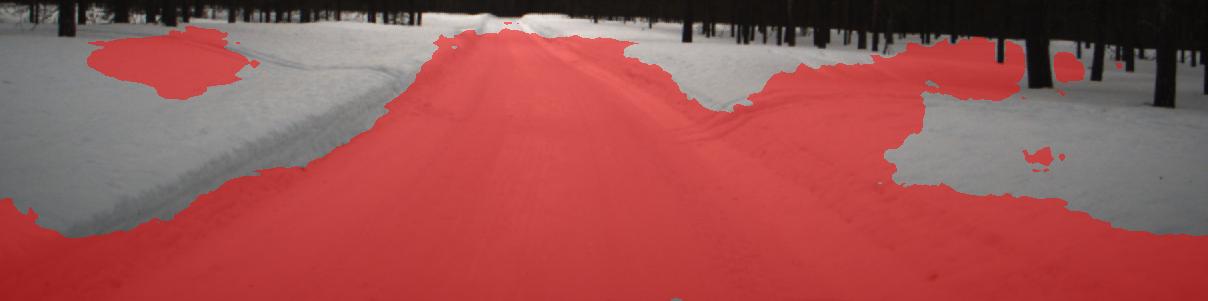} & \includegraphics[width=\figWidthFinal\linewidth]{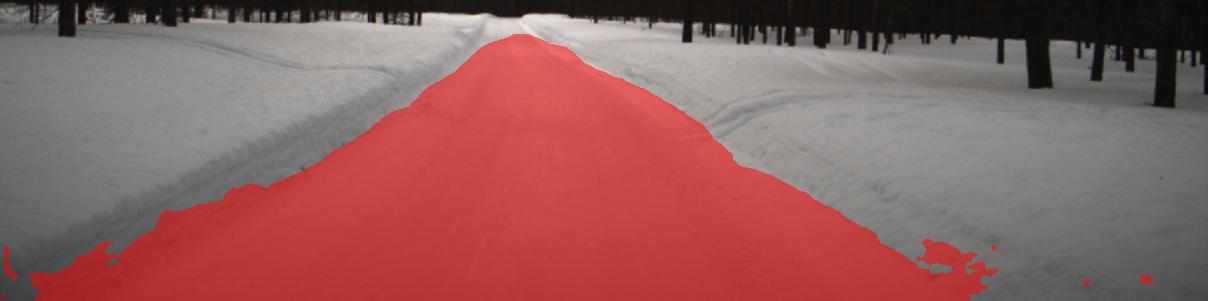} & \includegraphics[width=\figWidthFinal\linewidth]{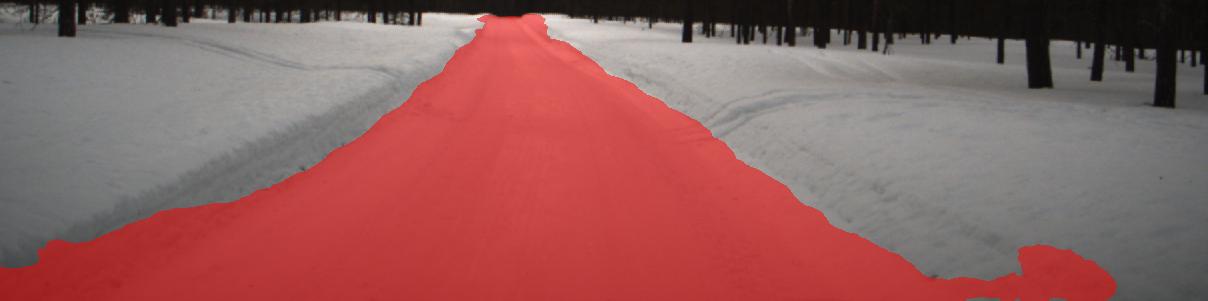} \\
\includegraphics[width=\figWidthFinal\linewidth]{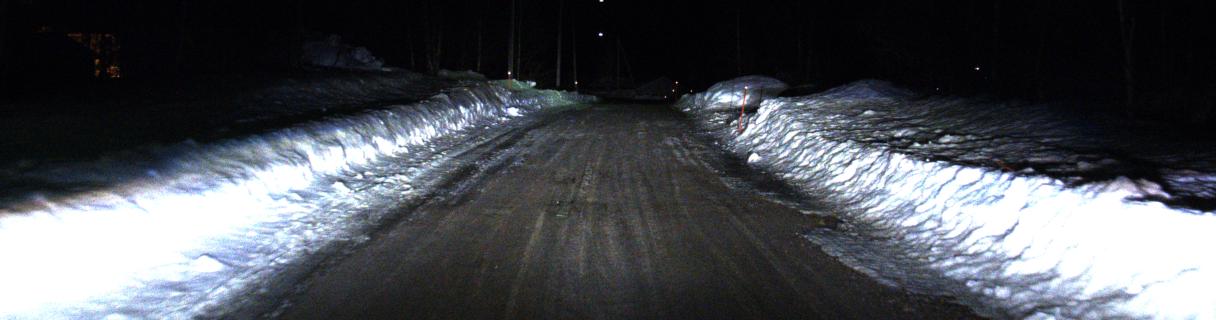} & \includegraphics[width=\figWidthFinal\linewidth]{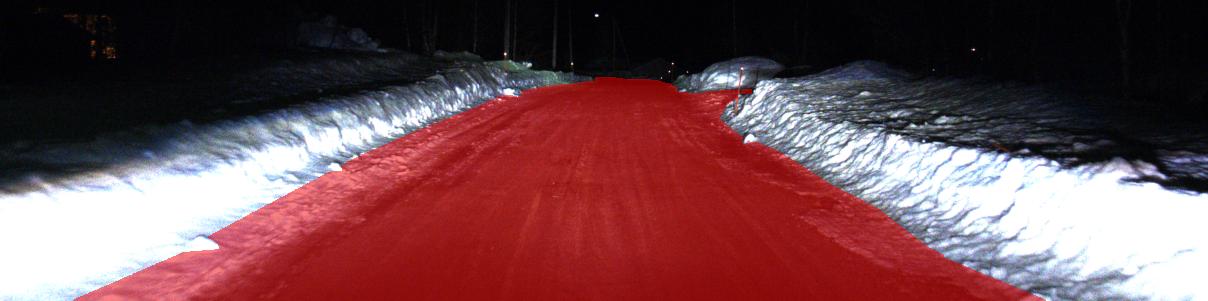} & \includegraphics[width=\figWidthFinal\linewidth]{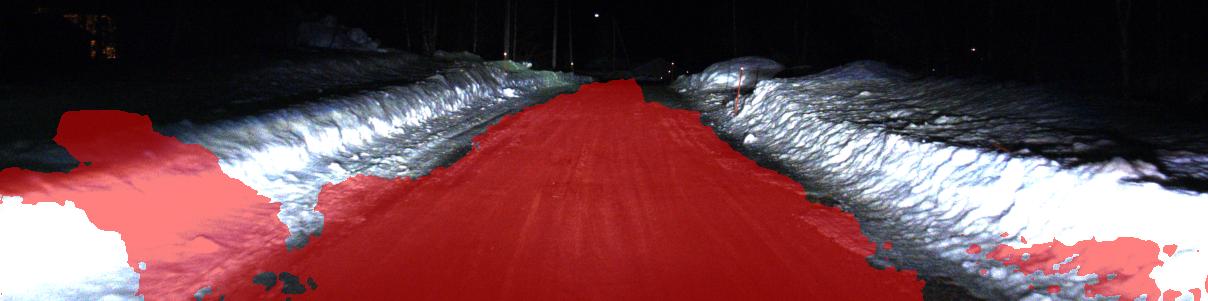} & \includegraphics[width=\figWidthFinal\linewidth]{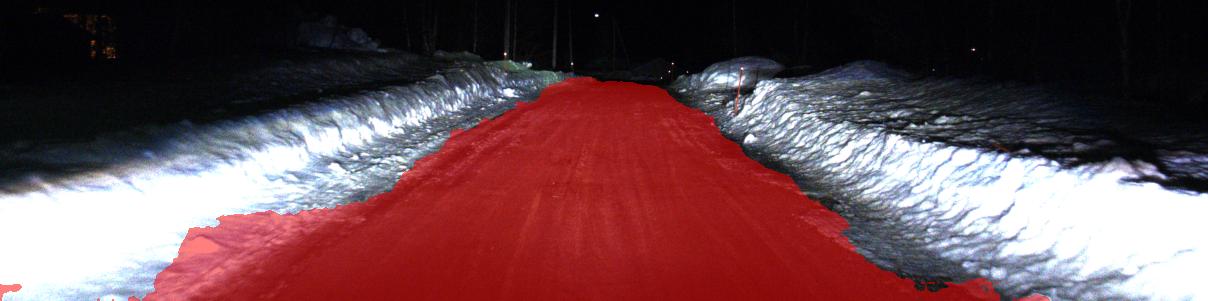} & \includegraphics[width=\figWidthFinal\linewidth]{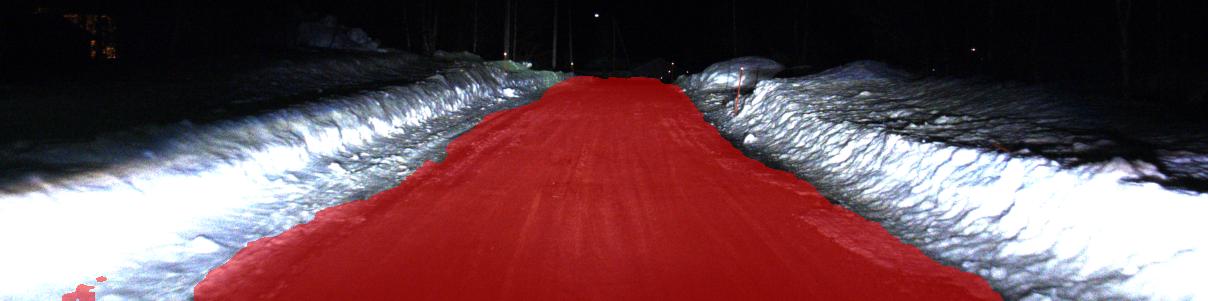} \\
\includegraphics[width=\figWidthFinal\linewidth]{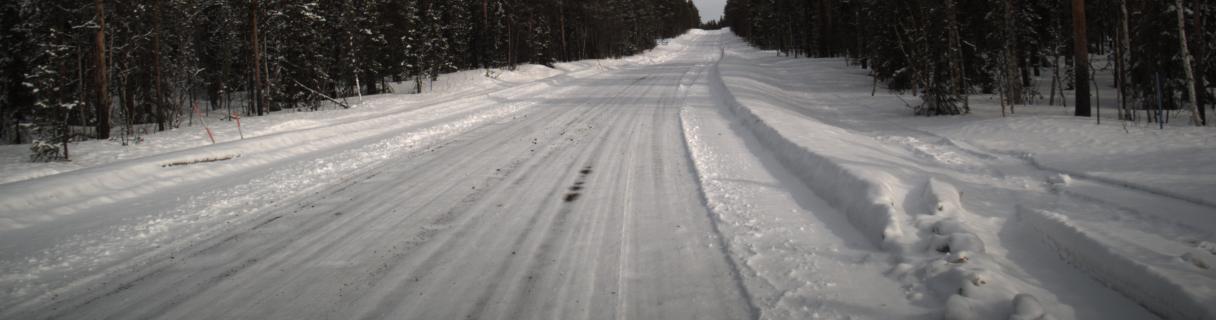} & \includegraphics[width=\figWidthFinal\linewidth]{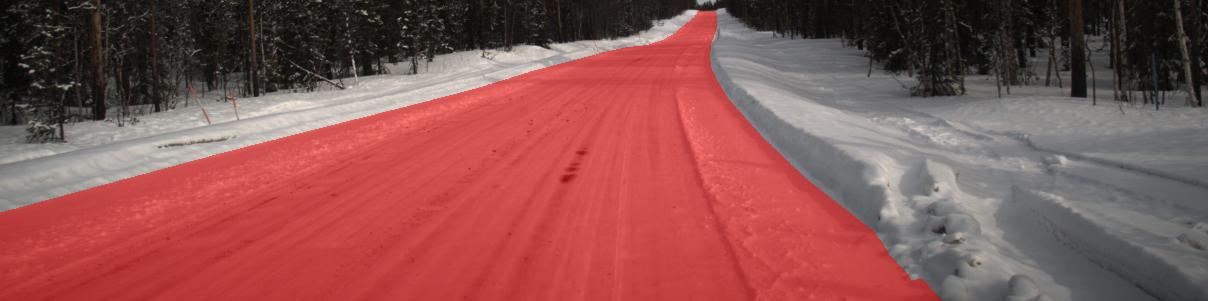} & \includegraphics[width=\figWidthFinal\linewidth]{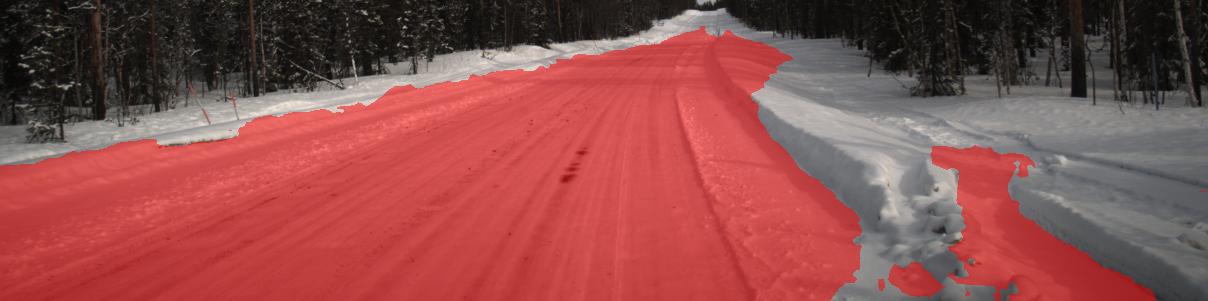} & \includegraphics[width=\figWidthFinal\linewidth]{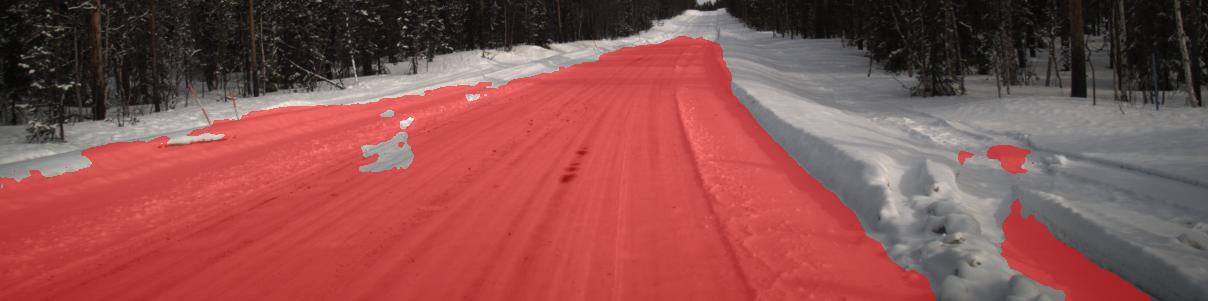} & \includegraphics[width=\figWidthFinal\linewidth]{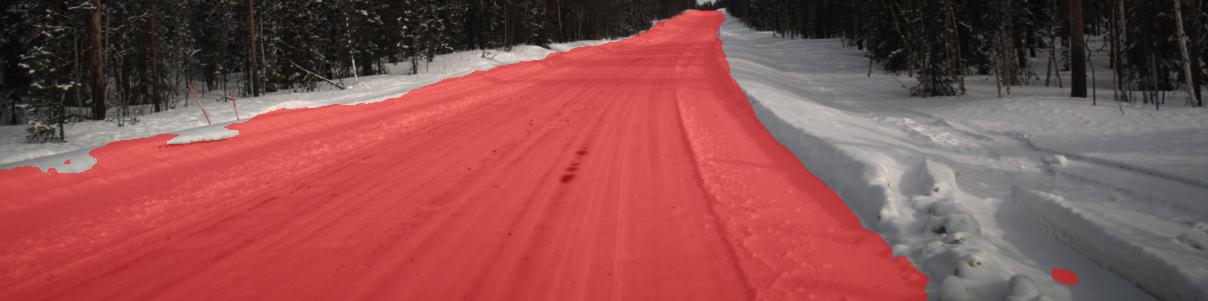} \\
\includegraphics[width=\figWidthFinal\linewidth]{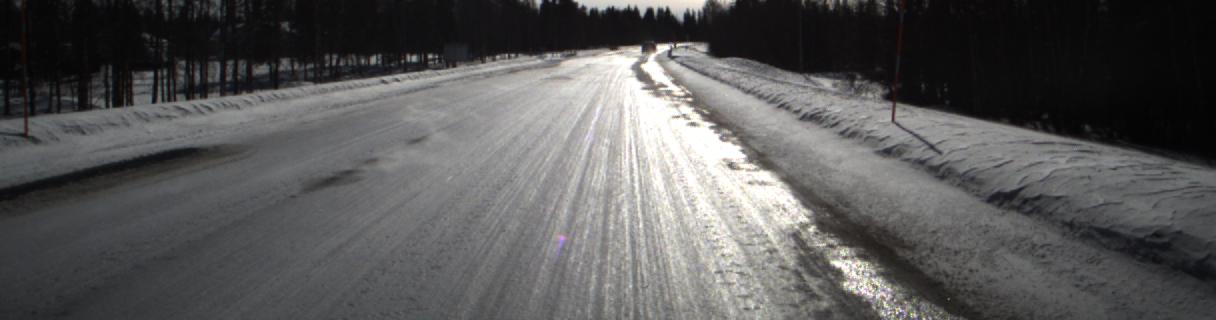} & \includegraphics[width=\figWidthFinal\linewidth]{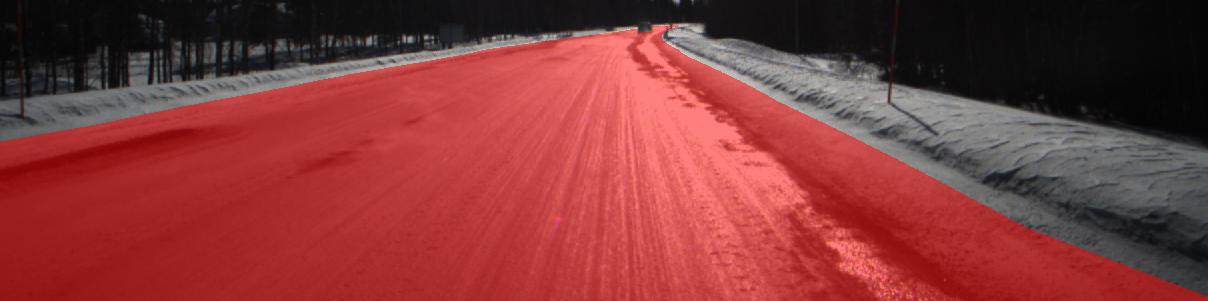} & \includegraphics[width=\figWidthFinal\linewidth]{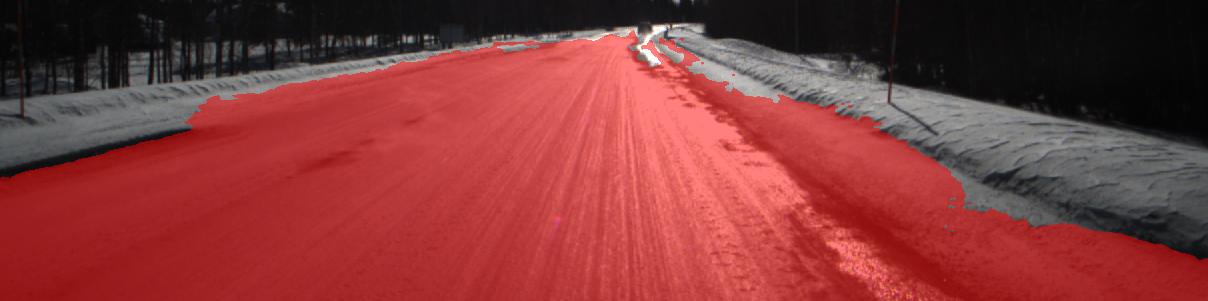} & \includegraphics[width=\figWidthFinal\linewidth]{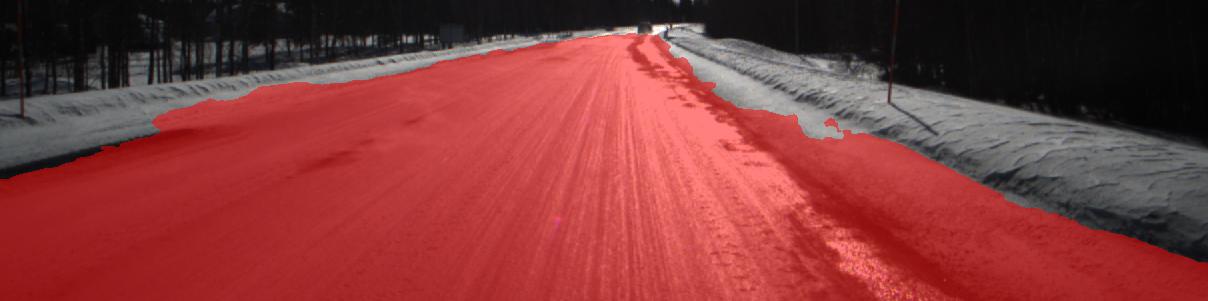} & \includegraphics[width=\figWidthFinal\linewidth]{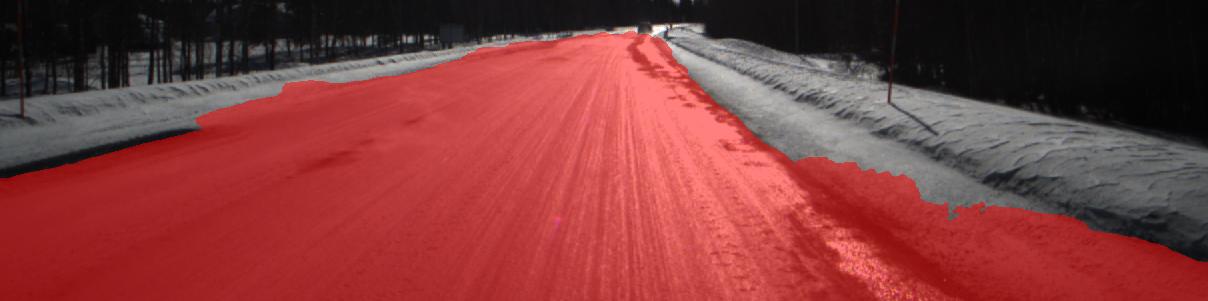} \\
\includegraphics[width=\figWidthFinal\linewidth]{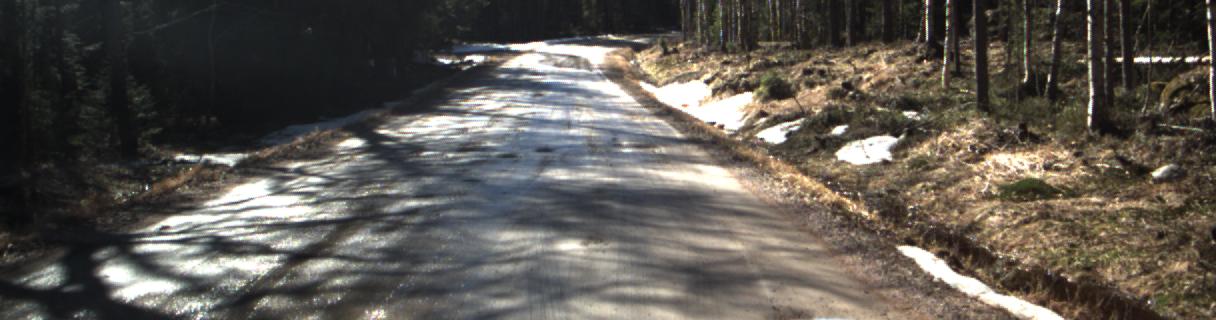} & \includegraphics[width=\figWidthFinal\linewidth]{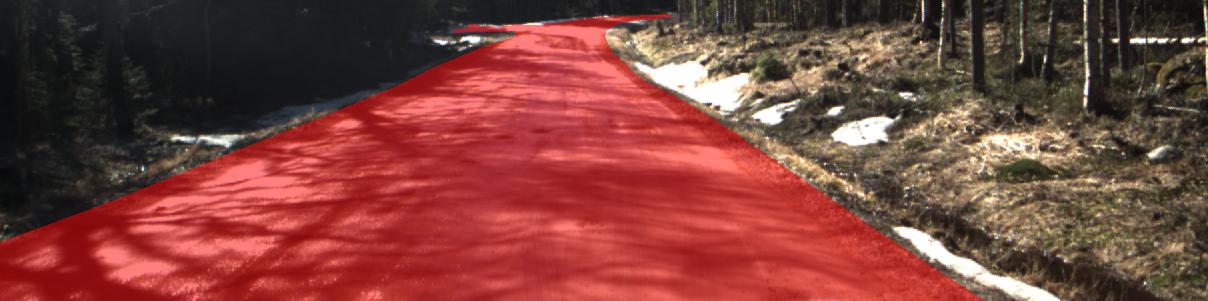} & \includegraphics[width=\figWidthFinal\linewidth]{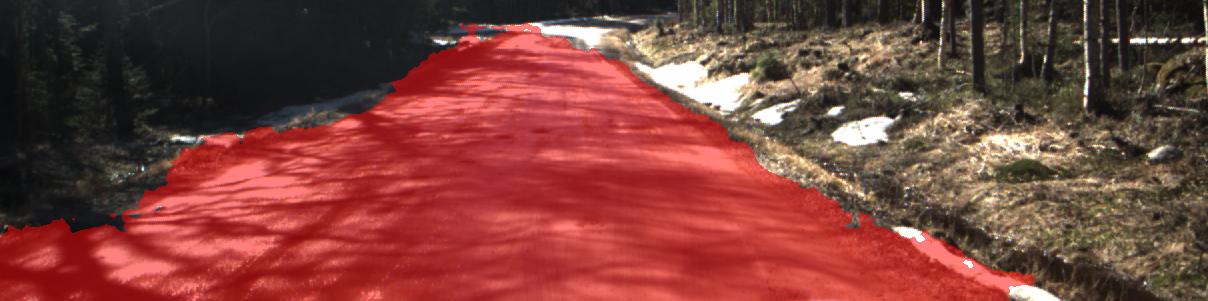} & \includegraphics[width=\figWidthFinal\linewidth]{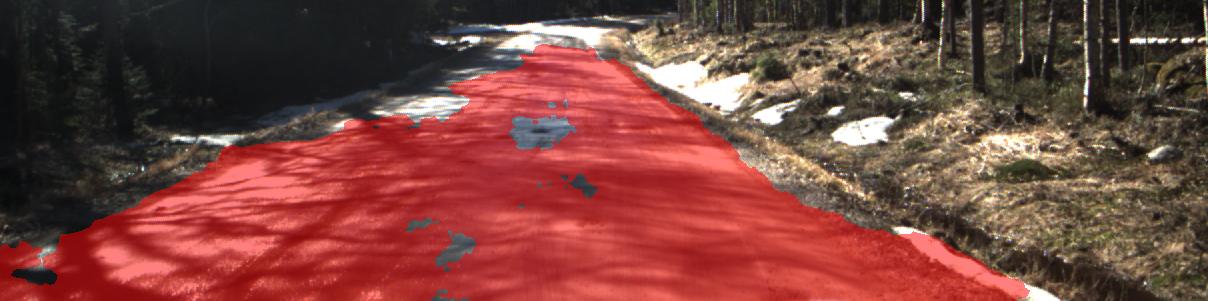} & \includegraphics[width=\figWidthFinal\linewidth]{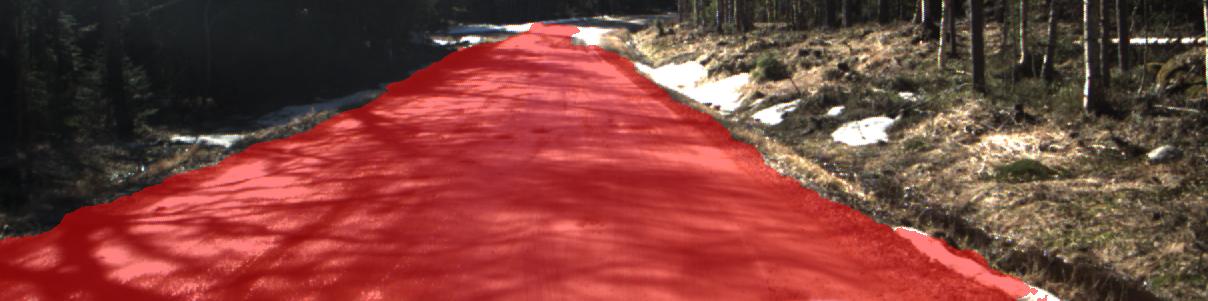}
\end{tabular}

\end{adjustbox}
\caption{Example road area segmentation masks evaluated with FPN ResNet based models on test set samples.}\label{fig:segmentation_examples2}
\end{figure}



\section*{Acknowledgement}

Academy of Finland projects (decisions 318437 and 319011) and Henry Ford Foundation are gratefully acknowledged for financial support.

The authors' contribution is the following: Maanp{\"a}{\"a} designed and performed the experiments and wrote the manuscript. Melekhov advised in the model development and provided feedback on the manuscript. Maanp{\"a}{\"a}, Taher, and Manninen took equal shares in instrumenting the autonomous driving platform and developing software. Maanp{\"a}{\"a} and Taher participated in data collection. Hyypp{\"a} supervised the project.

\clearpage

\bibliography{arXiv/bibliography_arxiv}
\bibliographystyle{arXiv/tmlr}

\end{document}


\section*{Supplementary Material}

\section{Examples of Steering Feature Maps}

\begin{figure}[p]
\centering
\makebox[\textwidth][c]{
\begin{tabular}{ccccc}
Original image & Channel 1 & Channel 2 & Channel 3 & Channel 4 \\
\includegraphics[width=0.2\linewidth]{testi/1539761851635811558.jpg} & \includegraphics[width=0.2\linewidth]{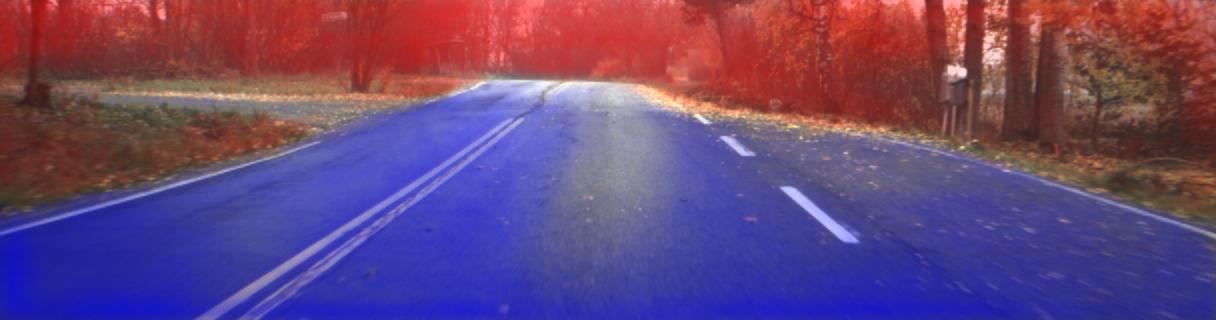} & \includegraphics[width=0.2\linewidth]{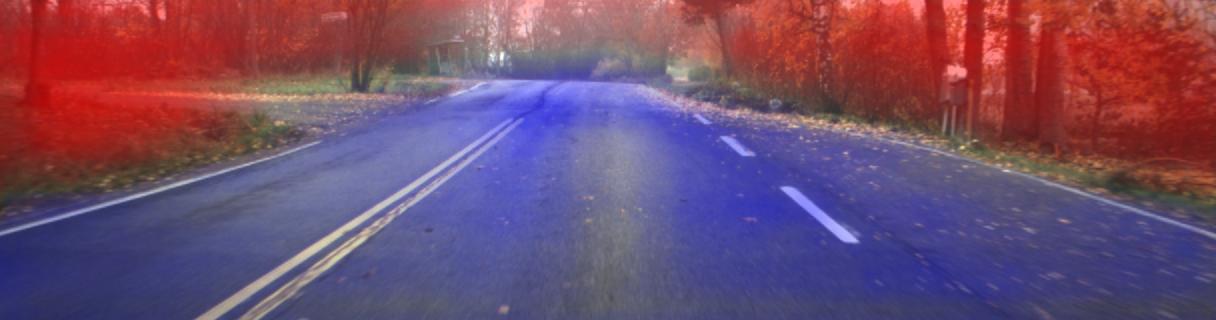} & \includegraphics[width=0.2\linewidth]{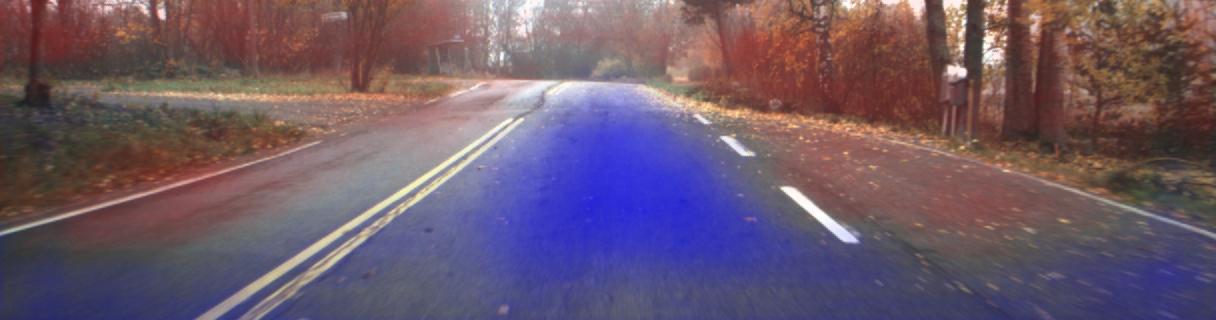} & \includegraphics[width=0.2\linewidth]{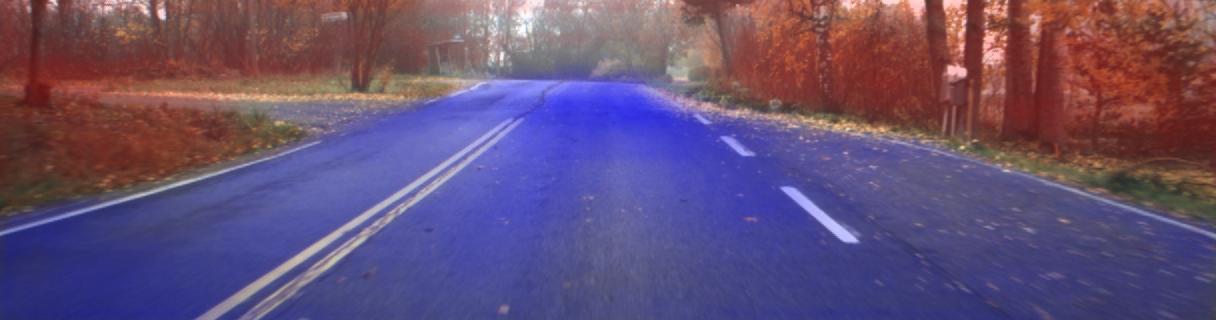} \\
\includegraphics[width=0.2\linewidth]{testi/1539765583750738068.jpg} & \includegraphics[width=0.2\linewidth]{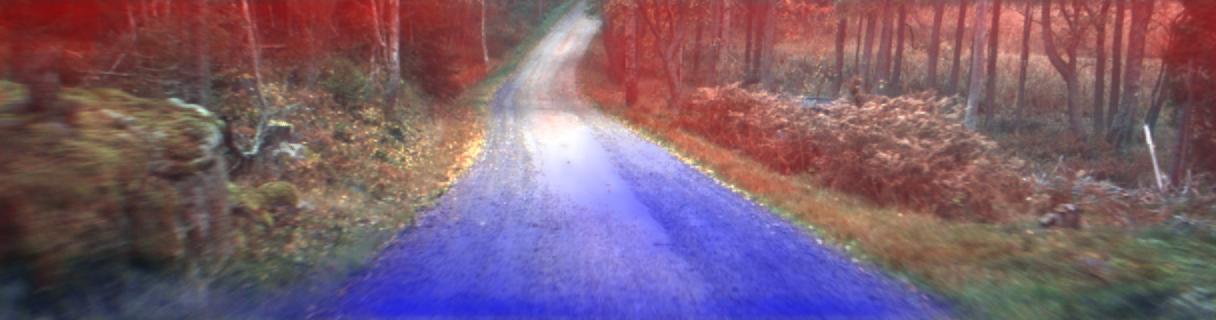} & \includegraphics[width=0.2\linewidth]{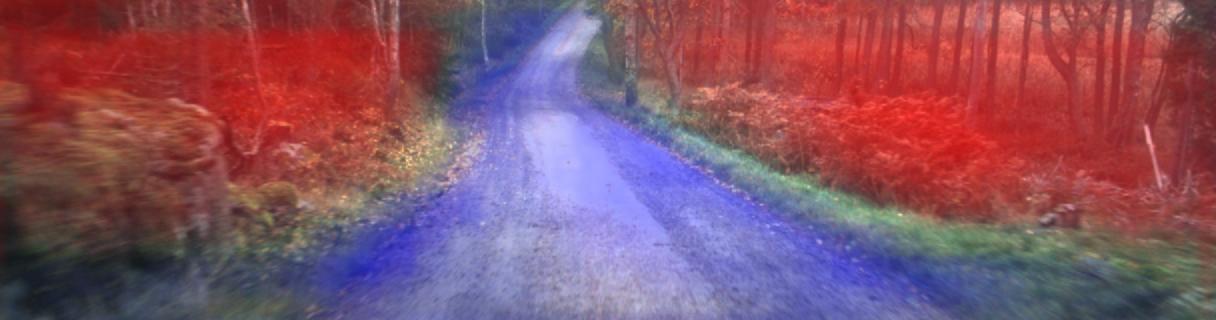} & \includegraphics[width=0.2\linewidth]{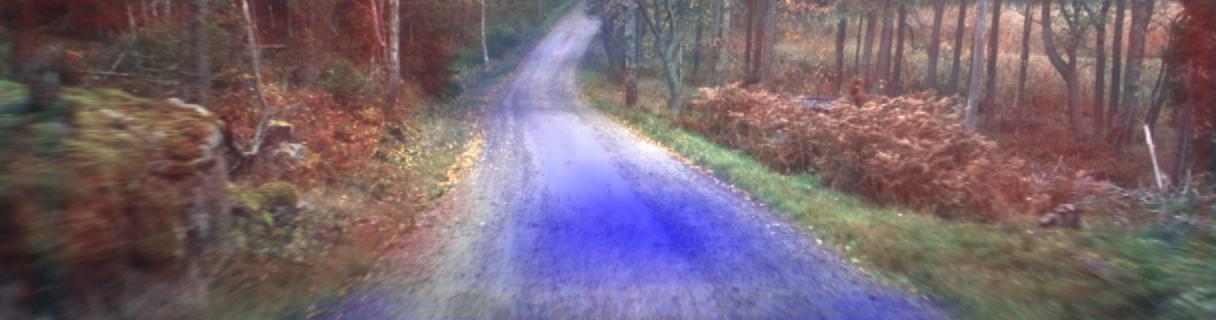} & \includegraphics[width=0.2\linewidth]{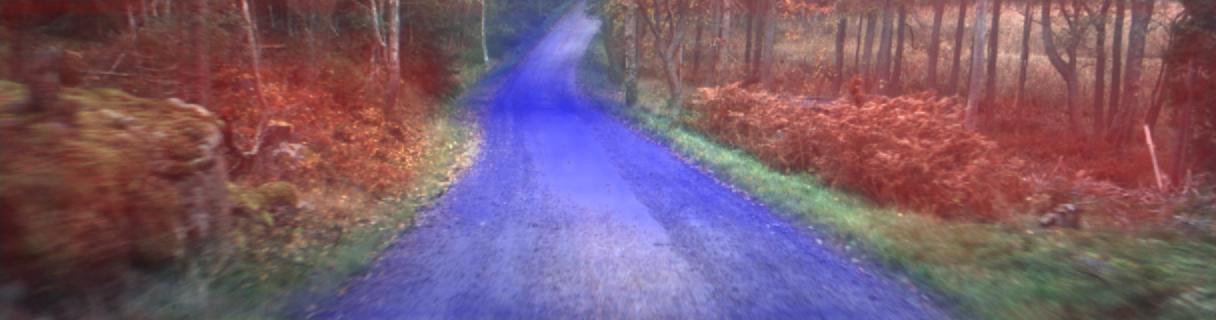} \\
\includegraphics[width=0.2\linewidth]{testi/1547210340470251952.jpg} & \includegraphics[width=0.2\linewidth]{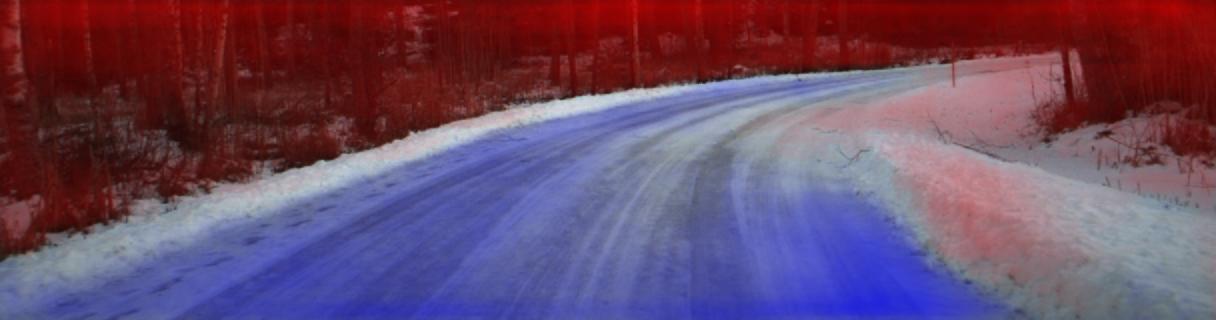} & \includegraphics[width=0.2\linewidth]{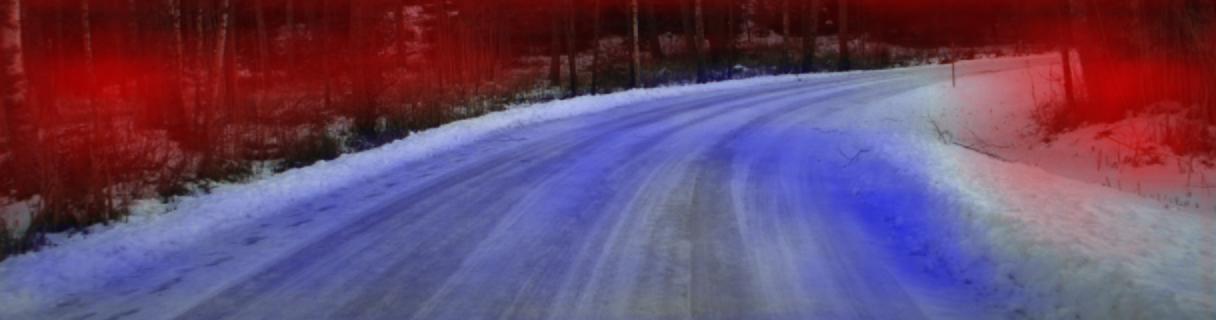} & \includegraphics[width=0.2\linewidth]{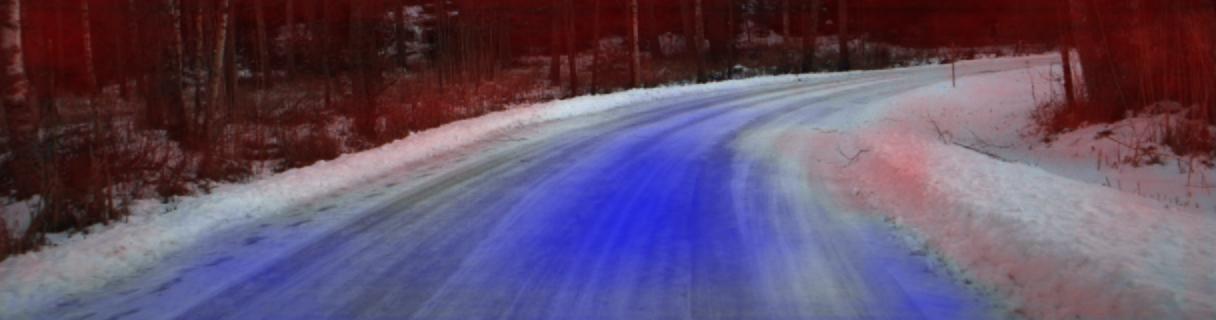} & \includegraphics[width=0.2\linewidth]{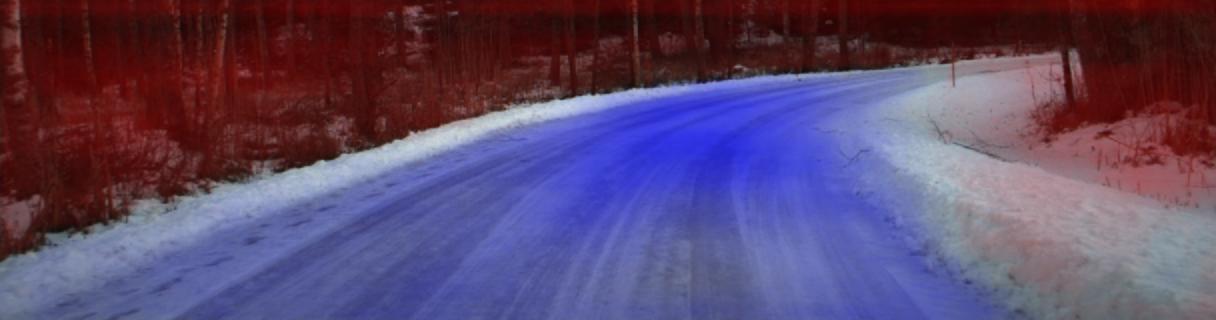} \\
\includegraphics[width=0.2\linewidth]{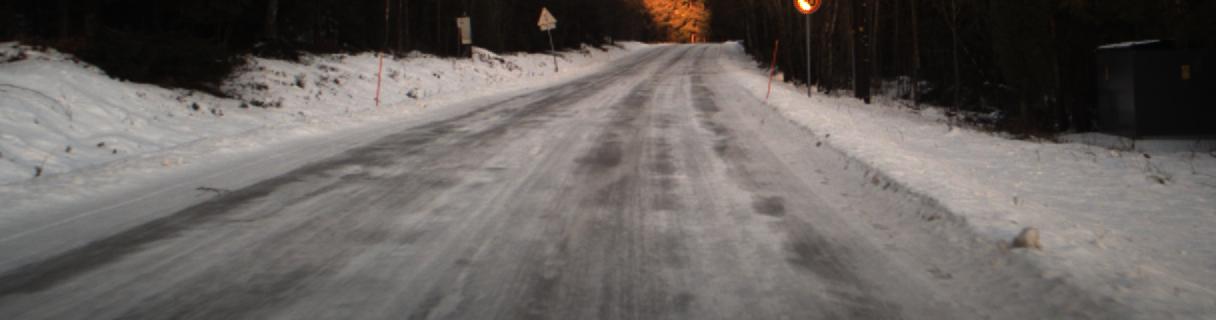} & \includegraphics[width=0.2\linewidth]{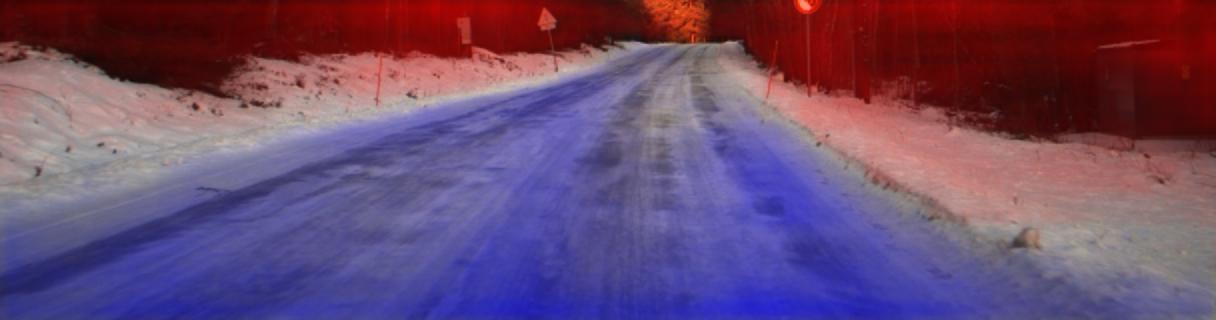} & \includegraphics[width=0.2\linewidth]{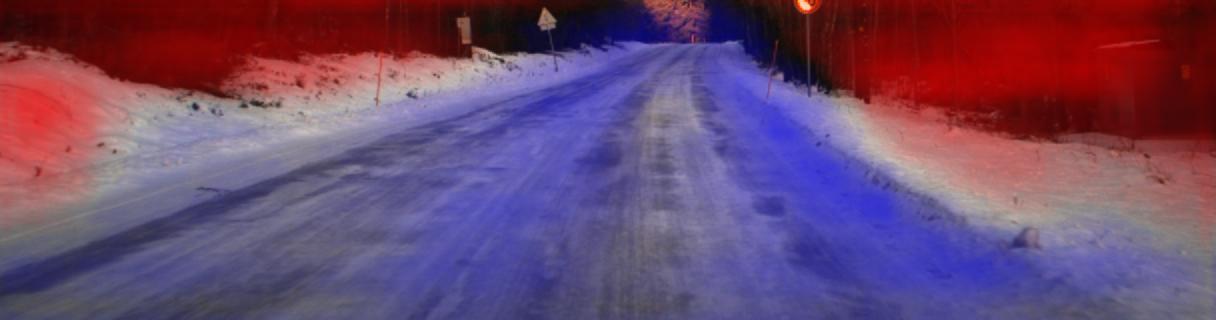} & \includegraphics[width=0.2\linewidth]{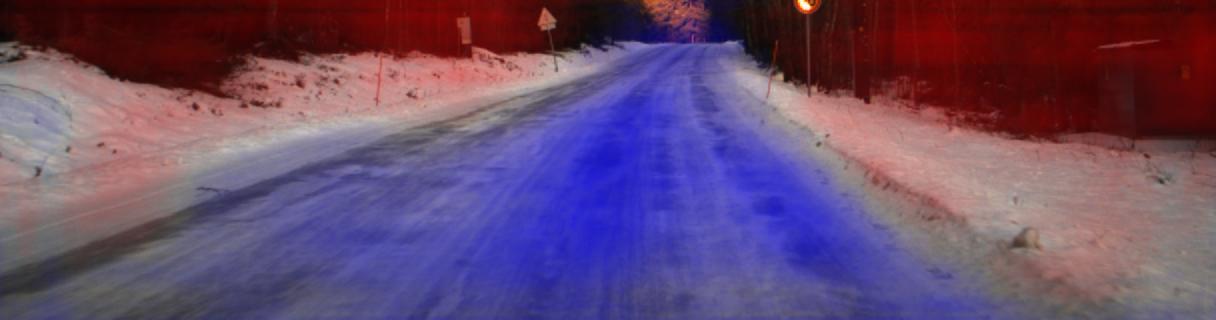} & \includegraphics[width=0.2\linewidth]{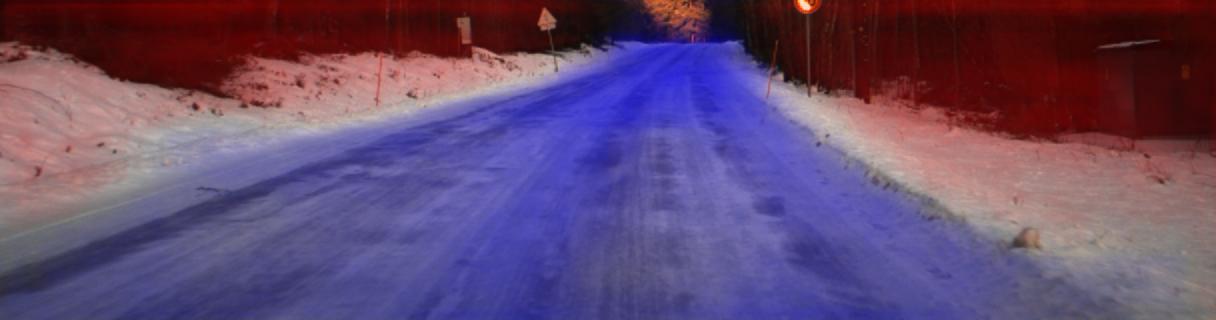} \\
\includegraphics[width=0.2\linewidth]{testi/1547463486843833753.jpg} & \includegraphics[width=0.2\linewidth]{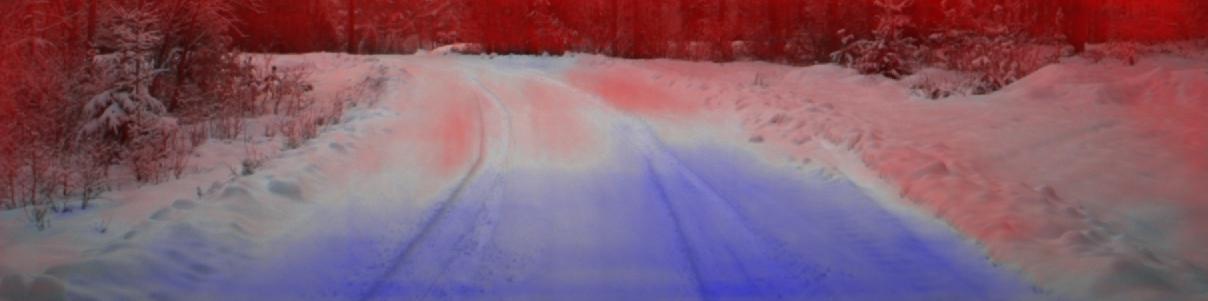} & \includegraphics[width=0.2\linewidth]{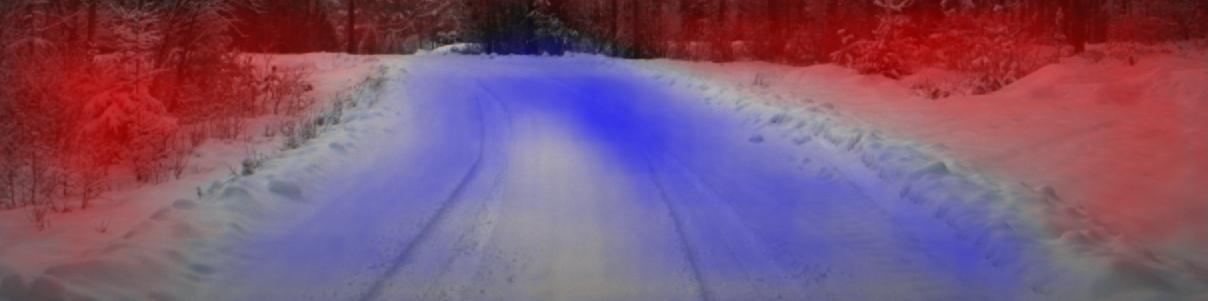} & \includegraphics[width=0.2\linewidth]{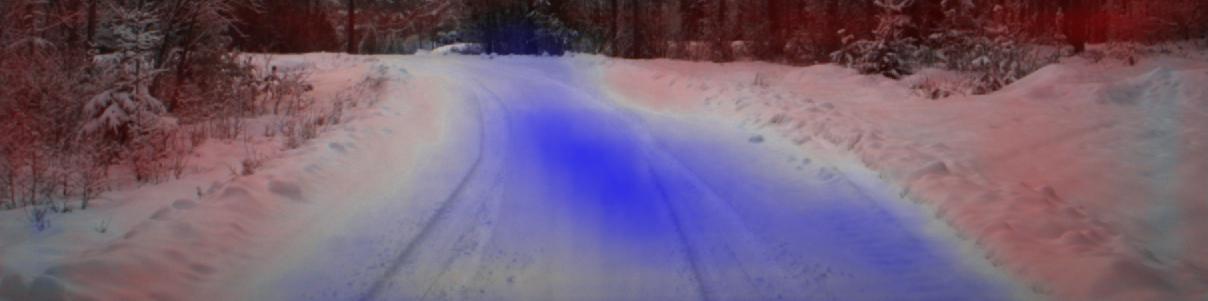} & \includegraphics[width=0.2\linewidth]{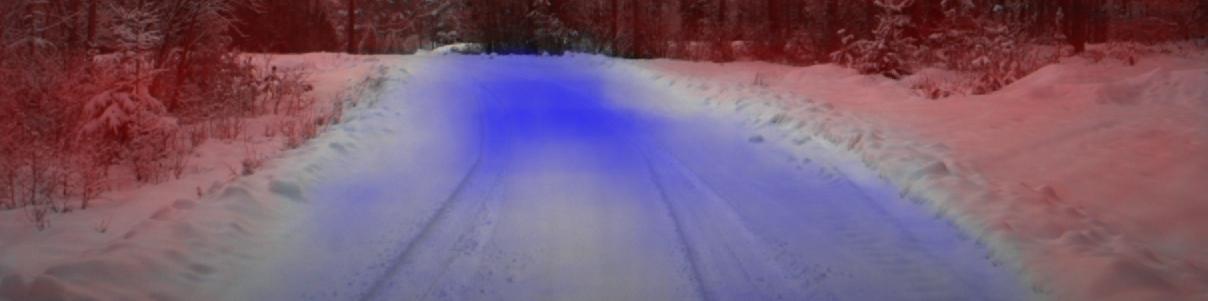} \\
\includegraphics[width=0.2\linewidth]{testi/1554370548031407510.jpg} & \includegraphics[width=0.2\linewidth]{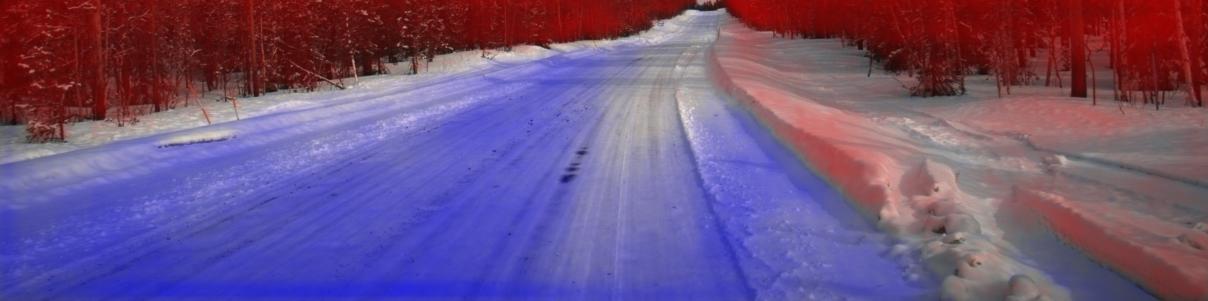} & \includegraphics[width=0.2\linewidth]{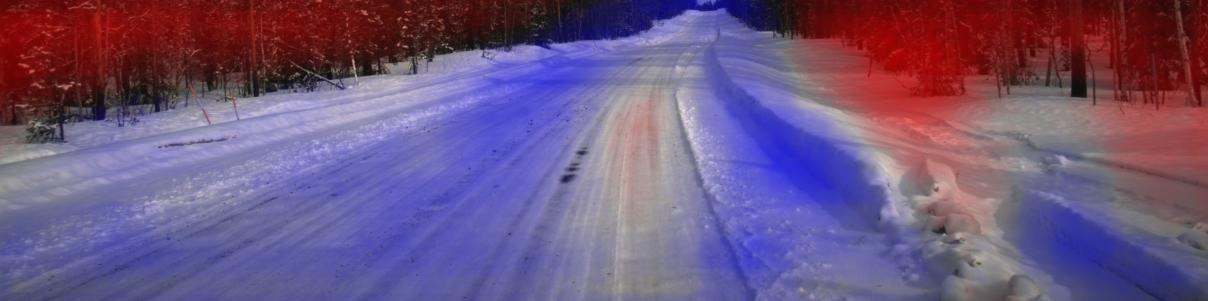} & \includegraphics[width=0.2\linewidth]{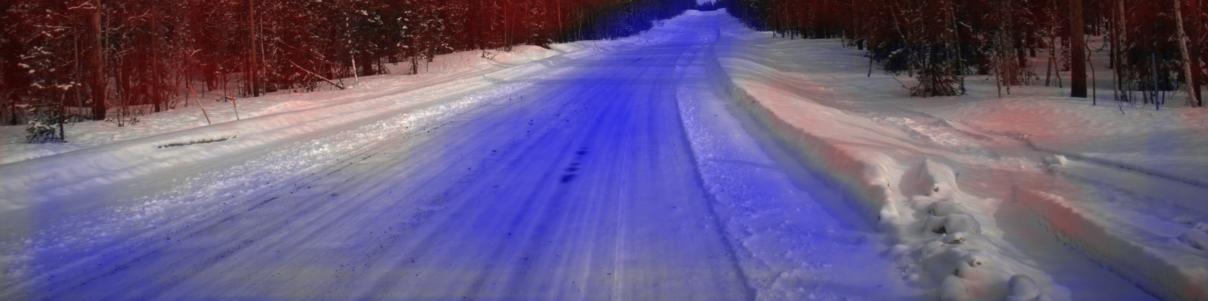} & \includegraphics[width=0.2\linewidth]{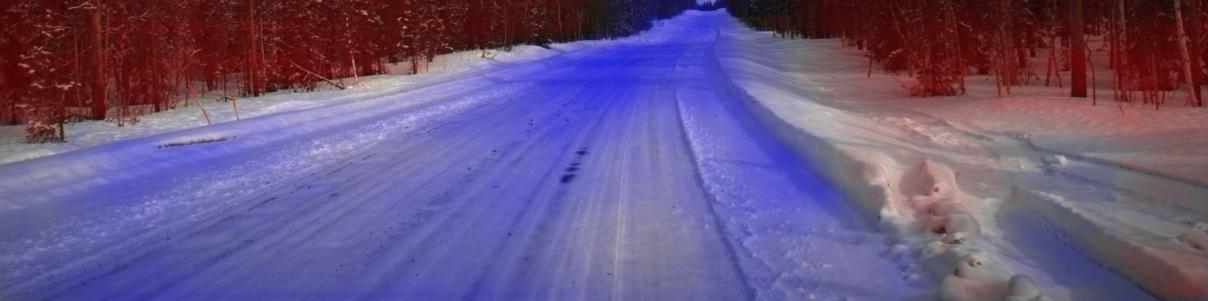} \\
\includegraphics[width=0.2\linewidth]{testi/1554377435123666072.jpg} & \includegraphics[width=0.2\linewidth]{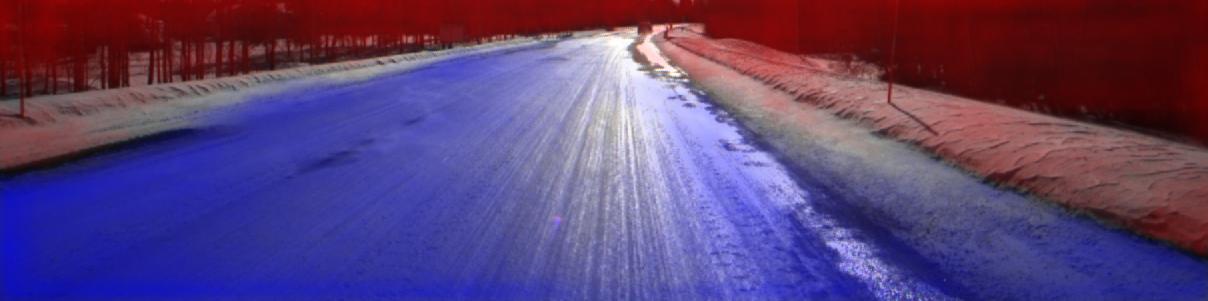} & \includegraphics[width=0.2\linewidth]{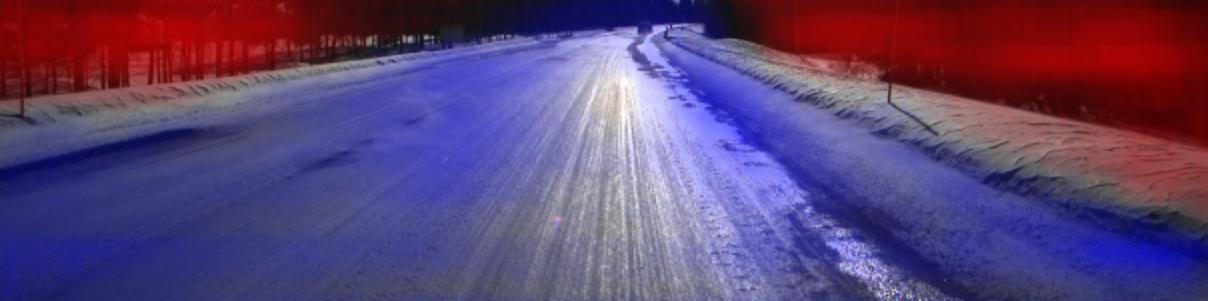} & \includegraphics[width=0.2\linewidth]{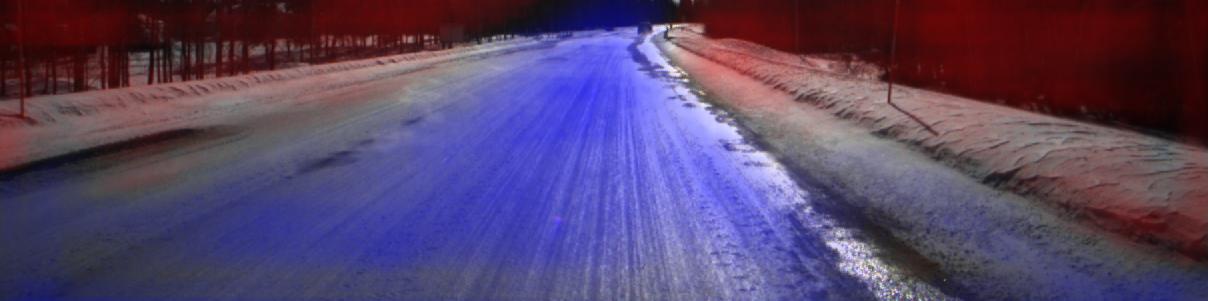} & \includegraphics[width=0.2\linewidth]{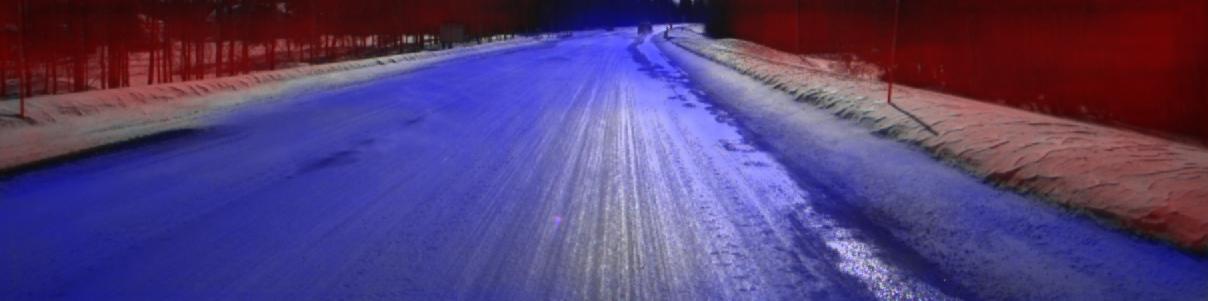} 
\end{tabular}
}
\caption{Example steering feature maps with FPN ResNet based multi-task model overlaid on test set samples. Red areas indicate positive feature values and blue areas indicate negative values. Steering feature values are normalized to the colour scale.}\label{fig:steerfeature_examples2}
\end{figure}

\begin{figure}[p]
\centering
\makebox[\textwidth][c]{
\begin{tabular}{ccccc}
Original image & Channel 1 & Channel 2 & Channel 3 & Channel 4 \\
\includegraphics[width=0.2\linewidth]{testi/1539761851635811558.jpg} & \includegraphics[width=0.2\linewidth]{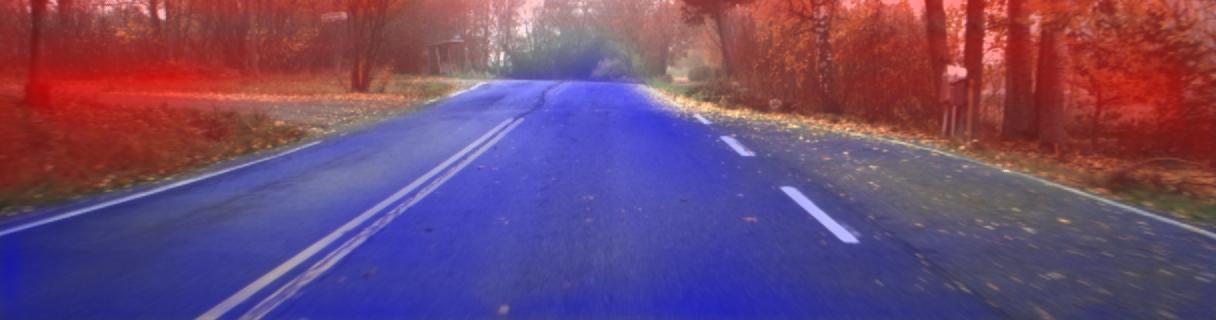} & \includegraphics[width=0.2\linewidth]{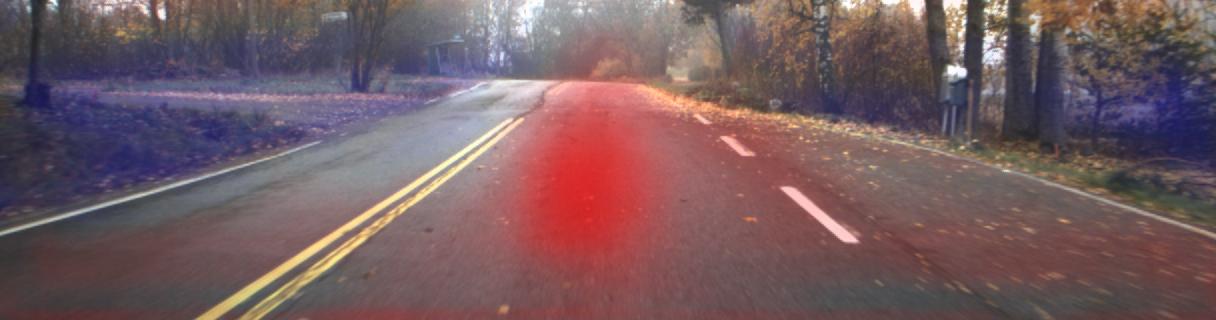} & \includegraphics[width=0.2\linewidth]{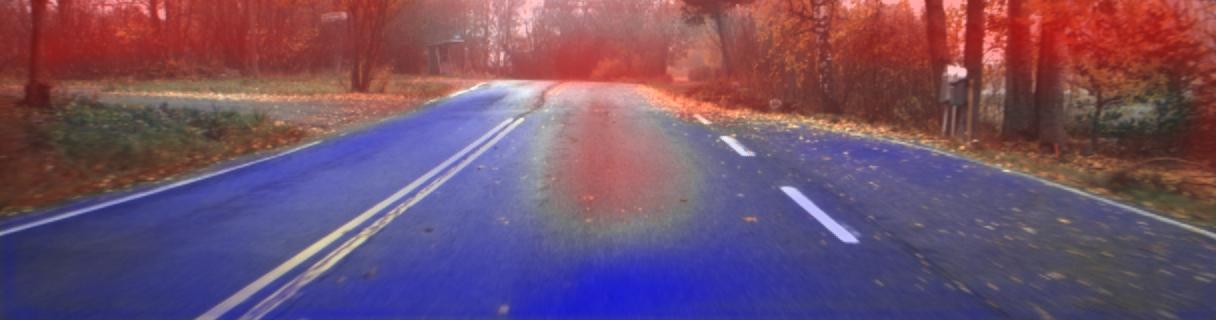} & \includegraphics[width=0.2\linewidth]{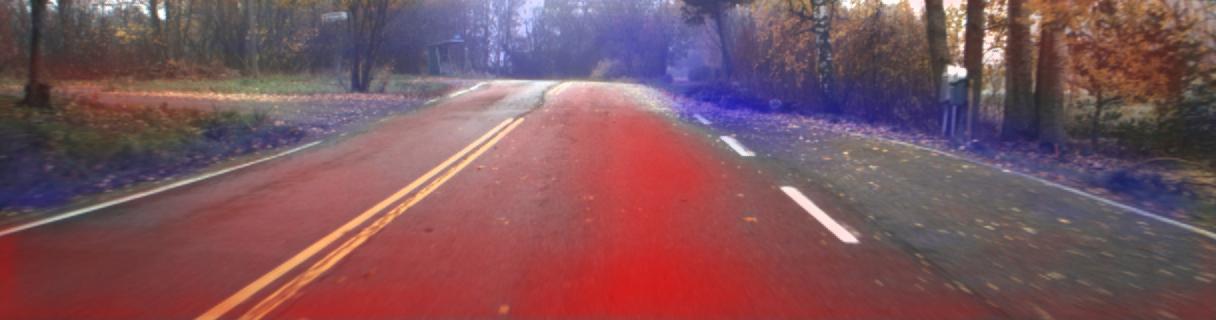} \\
\includegraphics[width=0.2\linewidth]{testi/1539765583750738068.jpg} & \includegraphics[width=0.2\linewidth]{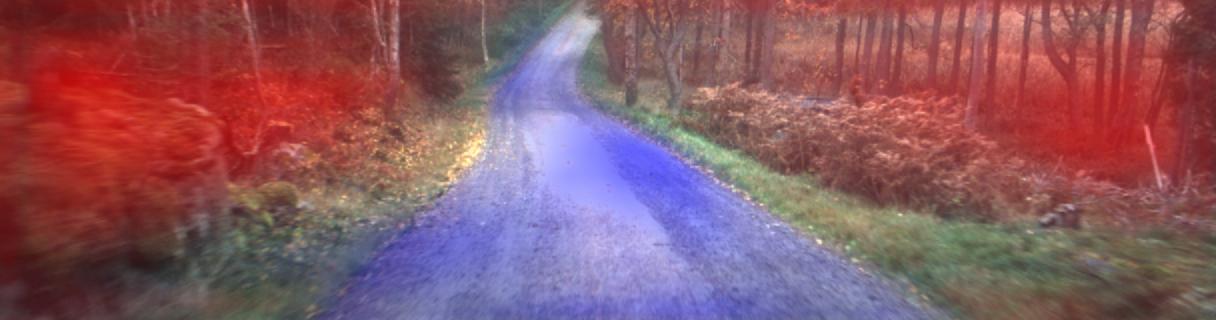} & \includegraphics[width=0.2\linewidth]{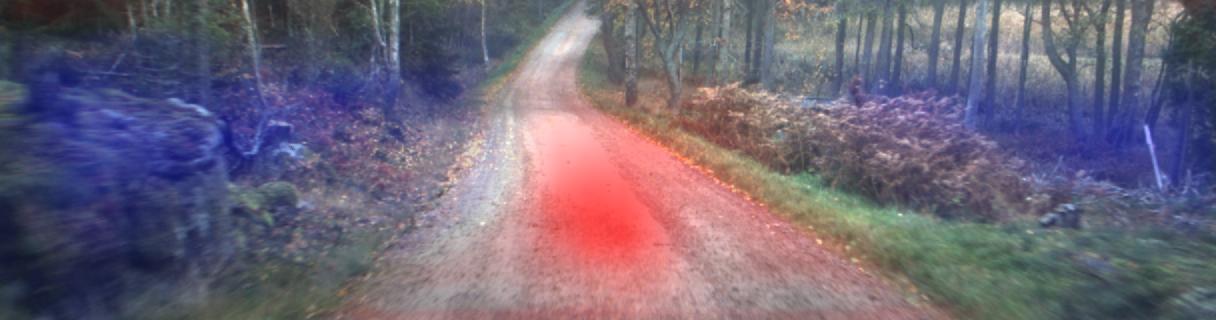} & \includegraphics[width=0.2\linewidth]{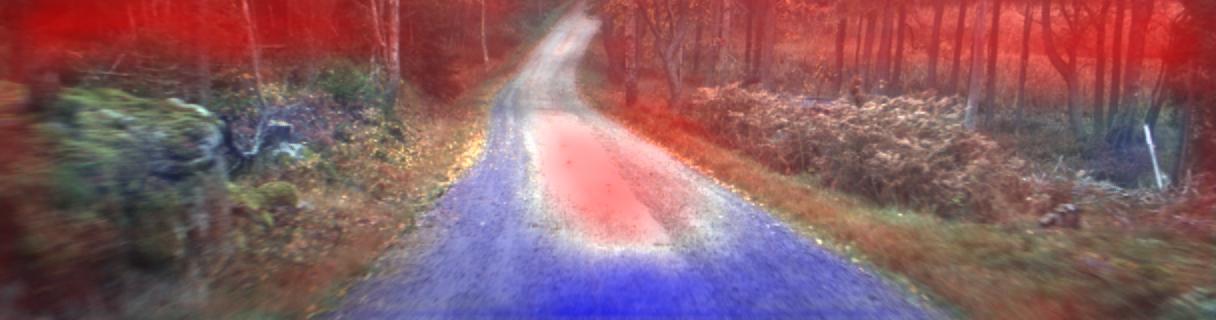} & \includegraphics[width=0.2\linewidth]{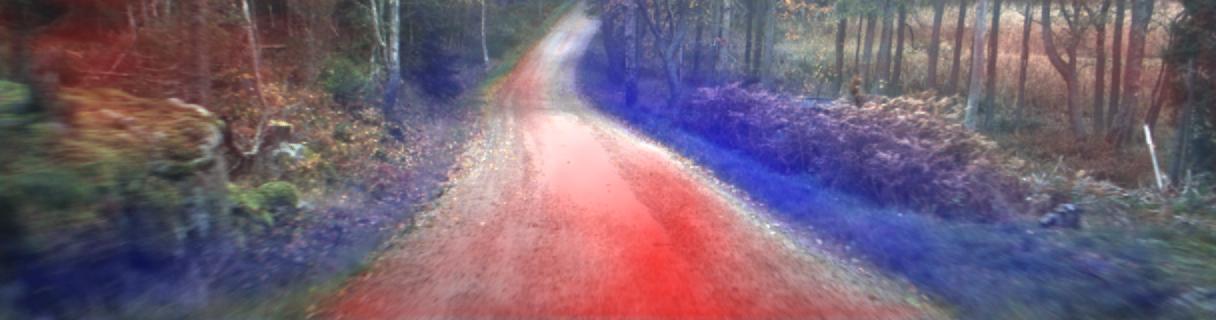} \\
\includegraphics[width=0.2\linewidth]{testi/1547210340470251952.jpg} & \includegraphics[width=0.2\linewidth]{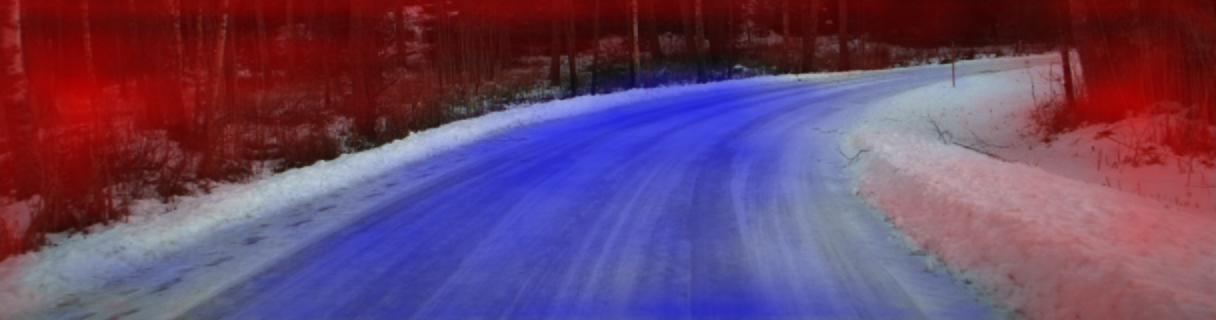} & \includegraphics[width=0.2\linewidth]{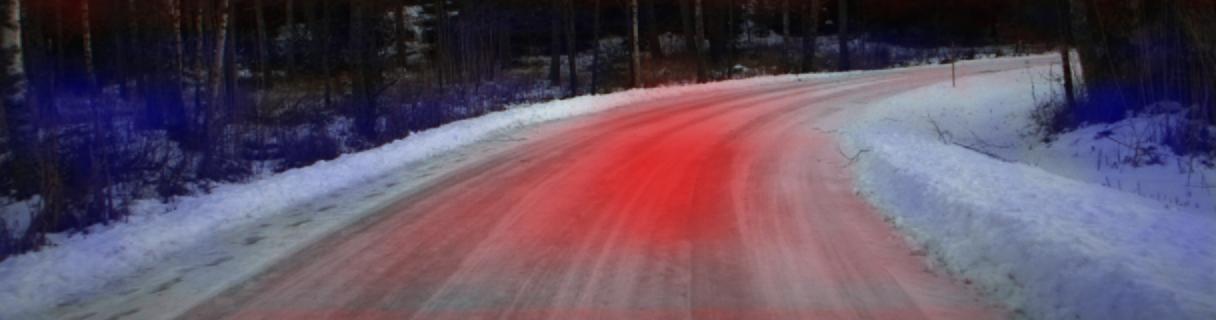} & \includegraphics[width=0.2\linewidth]{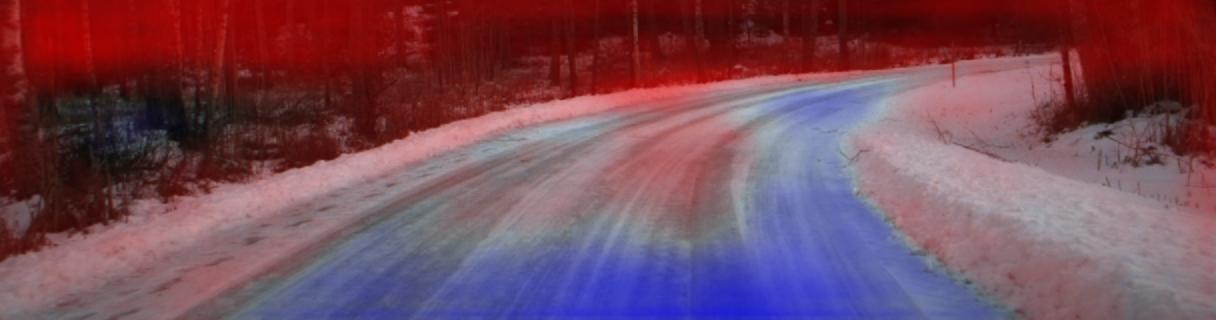} & \includegraphics[width=0.2\linewidth]{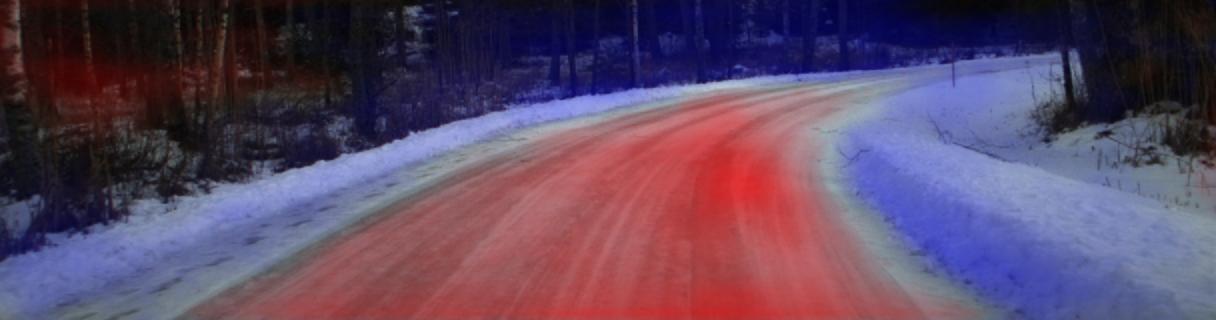} \\
\includegraphics[width=0.2\linewidth]{testi/1547211606777793387.jpg} & \includegraphics[width=0.2\linewidth]{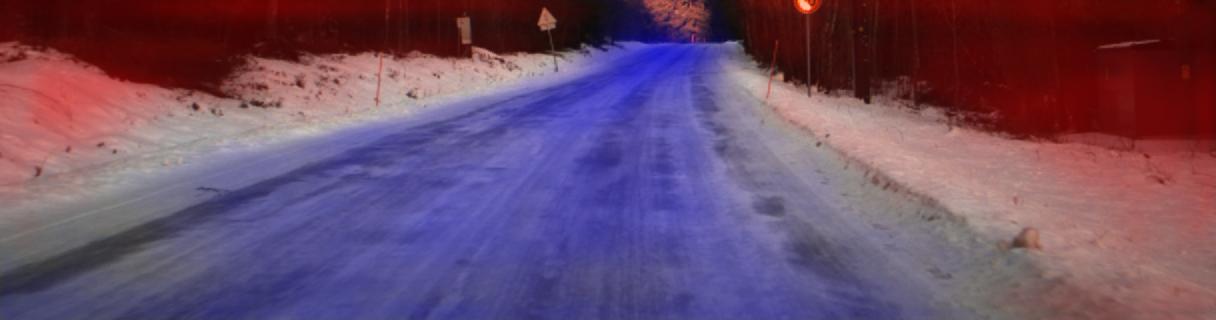} & \includegraphics[width=0.2\linewidth]{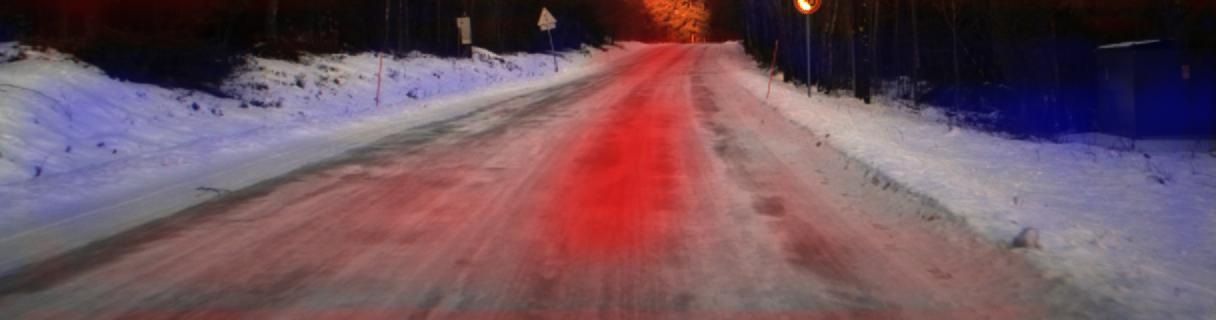} & \includegraphics[width=0.2\linewidth]{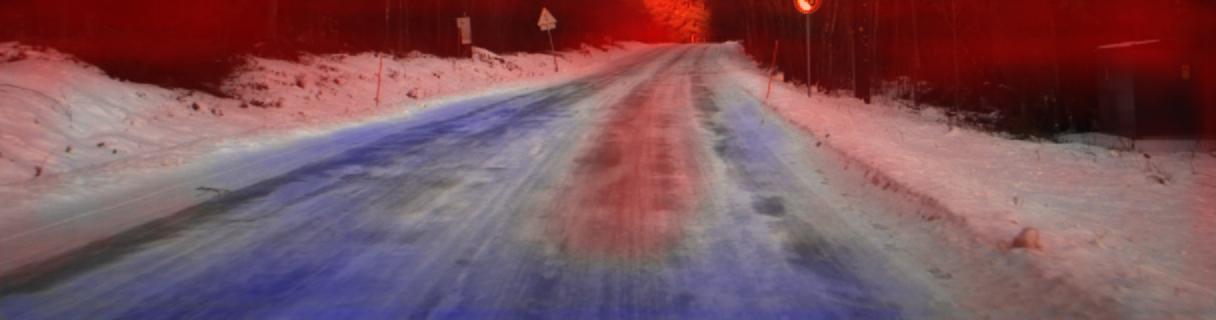} & \includegraphics[width=0.2\linewidth]{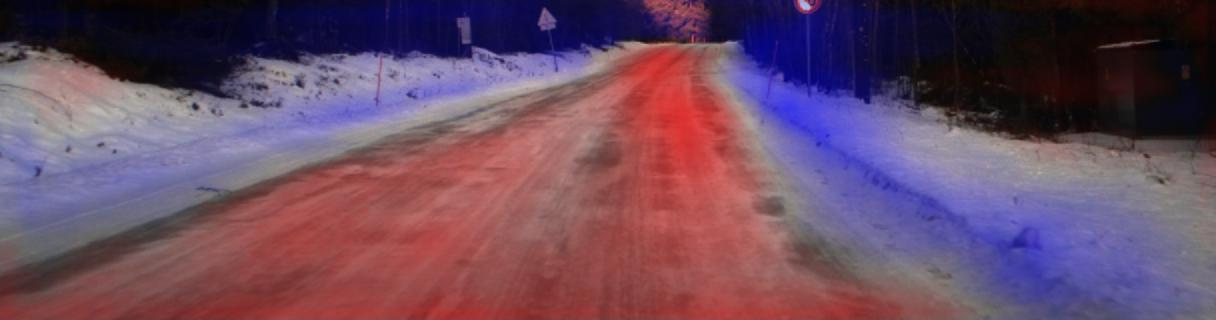} \\
\includegraphics[width=0.2\linewidth]{testi/1547463486843833753.jpg} & \includegraphics[width=0.2\linewidth]{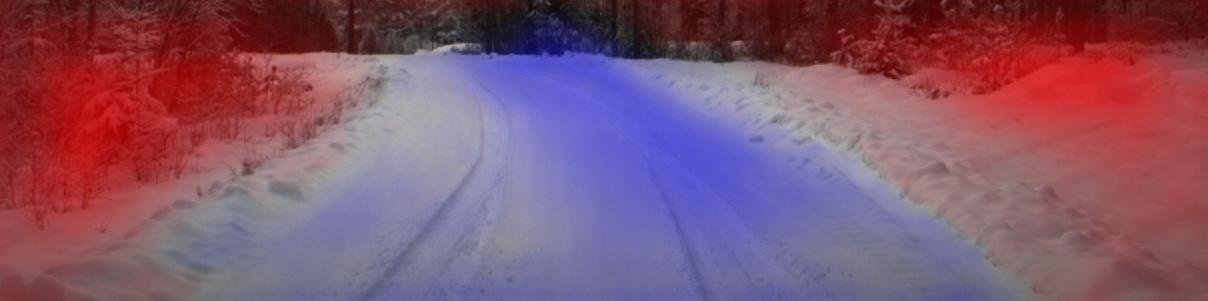} & \includegraphics[width=0.2\linewidth]{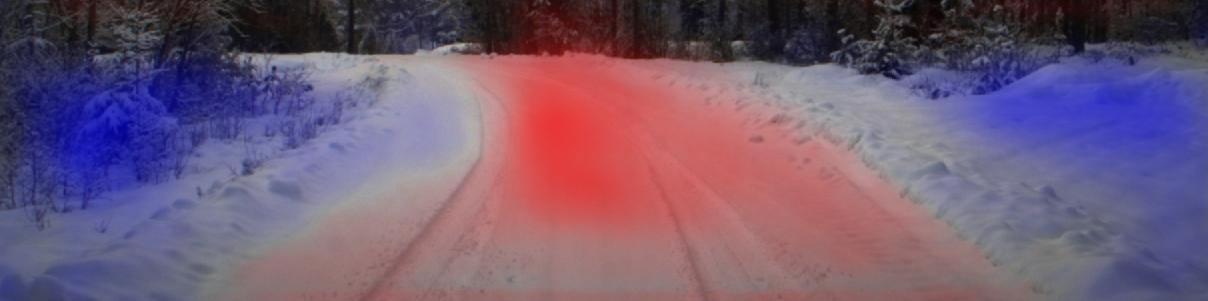} & \includegraphics[width=0.2\linewidth]{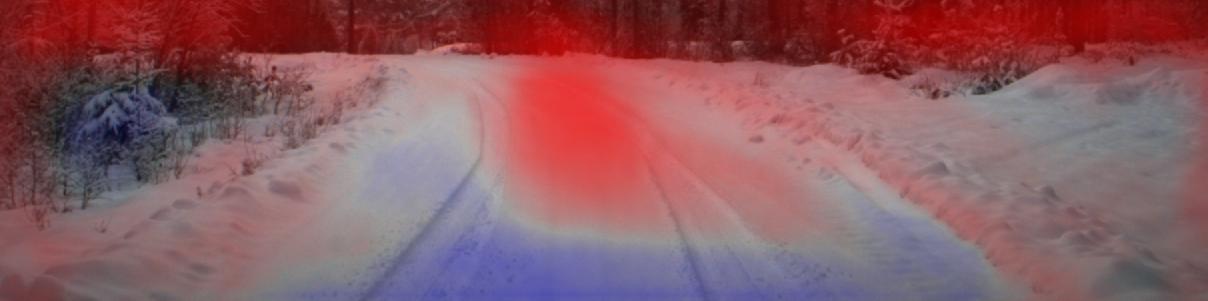} & \includegraphics[width=0.2\linewidth]{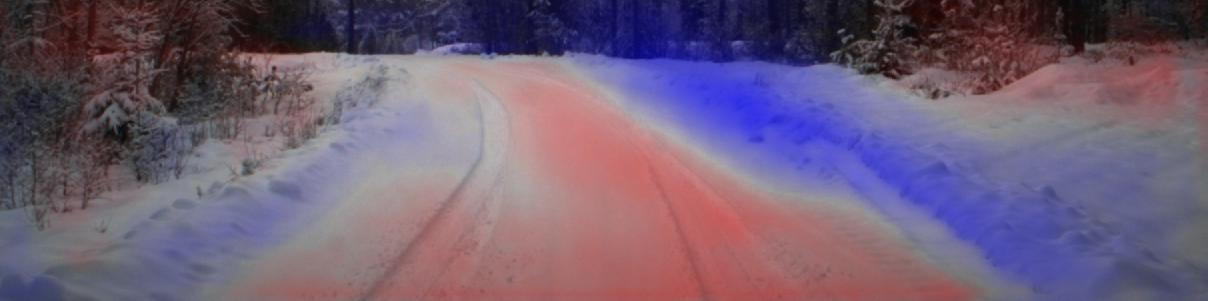} \\
\includegraphics[width=0.2\linewidth]{testi/1554370548031407510.jpg} & \includegraphics[width=0.2\linewidth]{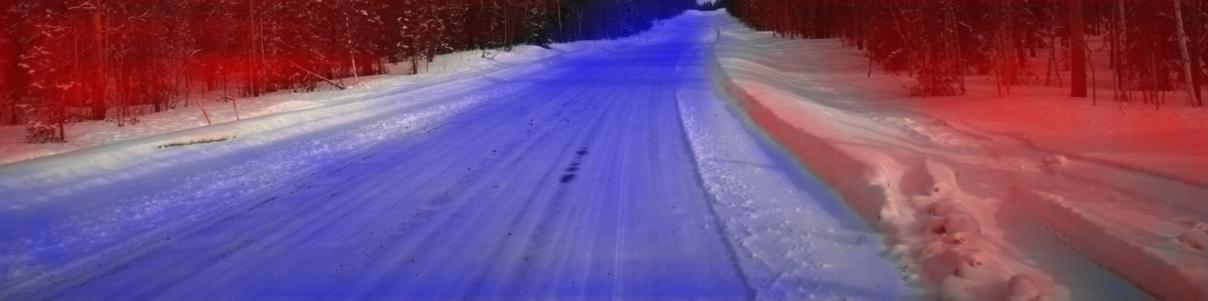} & \includegraphics[width=0.2\linewidth]{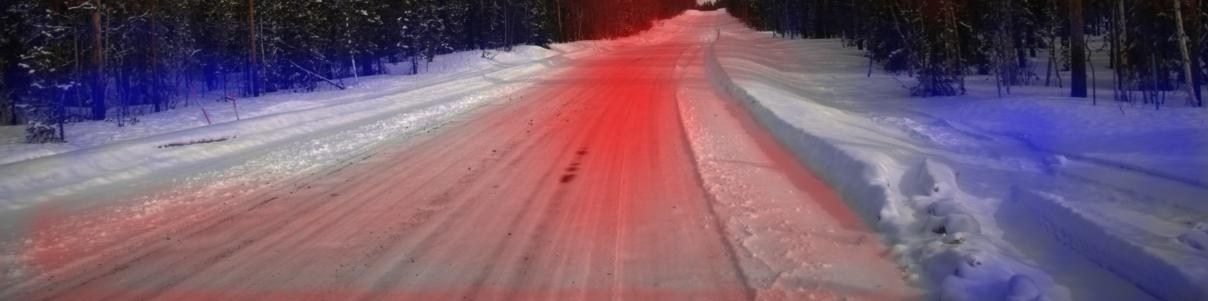} & \includegraphics[width=0.2\linewidth]{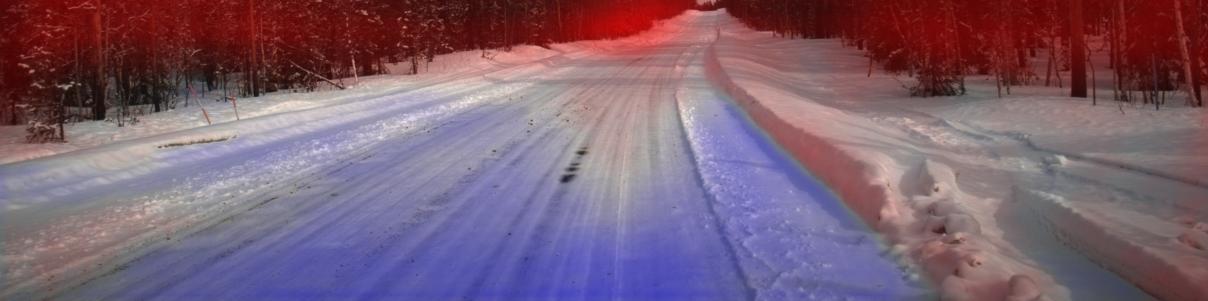} & \includegraphics[width=0.2\linewidth]{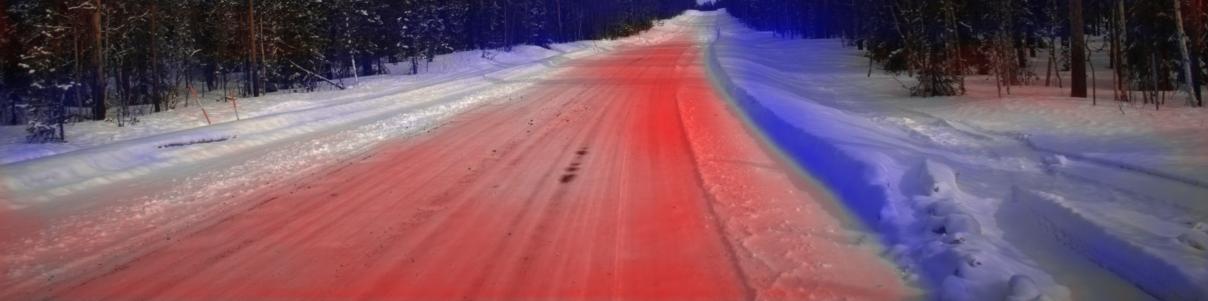} \\
\includegraphics[width=0.2\linewidth]{testi/1554377435123666072.jpg} & \includegraphics[width=0.2\linewidth]{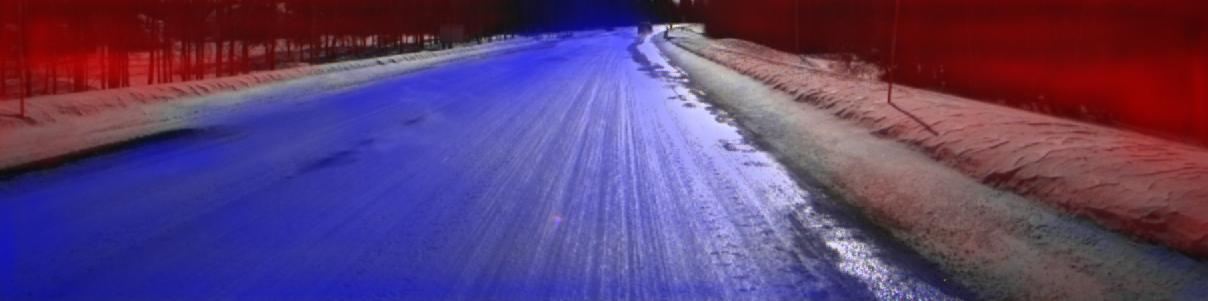} & \includegraphics[width=0.2\linewidth]{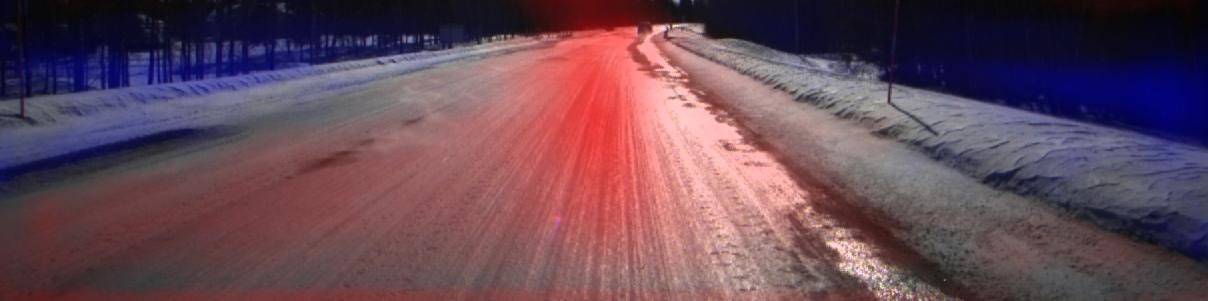} & \includegraphics[width=0.2\linewidth]{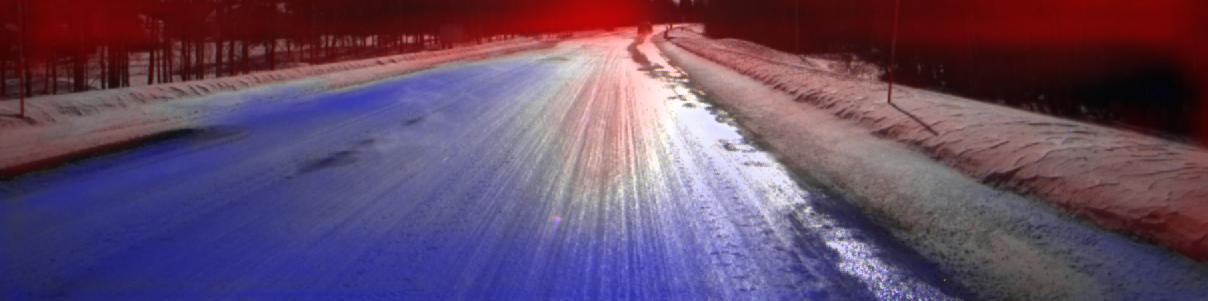} & \includegraphics[width=0.2\linewidth]{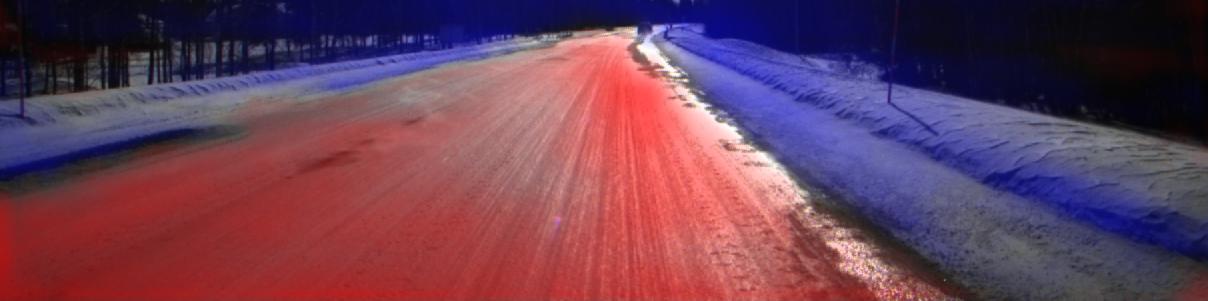} 
\end{tabular}
}
\caption{Example steering feature maps with HRNet based multi-task model overlaid on test set samples. Red areas indicate positive feature values and blue areas indicate negative values. Steering feature values are normalized to the colour scale.}\label{fig:steerfeature_examples}
\end{figure}

We present examples of steering feature masks in~\cref{fig:steerfeature_examples2,fig:steerfeature_examples}. These are used both for steering wheel angle prediction and for steering feature based road area segmentation. We observe that the features correlate well with the road area but have clearly weighted areas that correlate with the patterns indicating road curvature. For example, the negative blue areas in channel 2 correlate with lane sides in~\cref{fig:steerfeature_examples2}.

\section{Segmentation Masks with HRNet Backbone Based Models}

\begin{figure}[t]
\centering
\makebox[\textwidth][c]{
\begin{tabular}{ccccc}
Original image & Ground truth & ST model & TL model & \textbf{Proposed model} \\
\includegraphics[width=0.2\linewidth]{testi/1539761851635811558.jpg} & \includegraphics[width=0.2\linewidth]{testi/1539761851635811558_gt.jpg} & \includegraphics[width=0.2\linewidth]{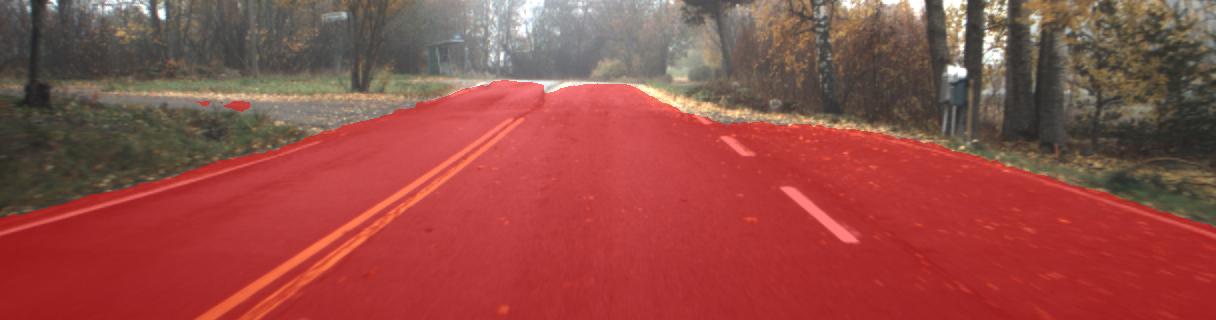} & \includegraphics[width=0.2\linewidth]{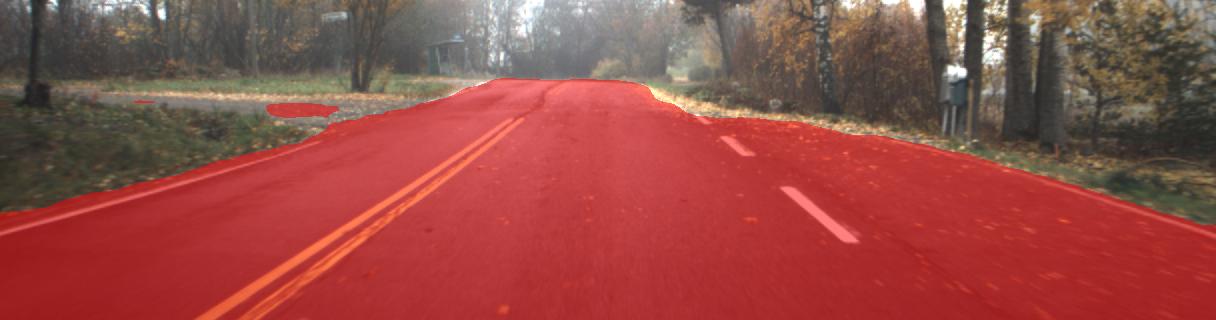} & \includegraphics[width=0.2\linewidth]{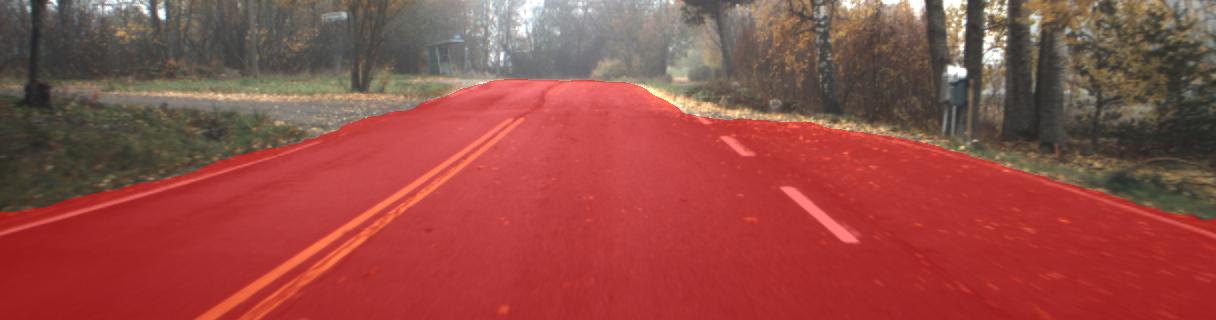} \\
\includegraphics[width=0.2\linewidth]{testi/1539765583750738068.jpg} & \includegraphics[width=0.2\linewidth]{testi/1539765583750738068_gt.jpg} & \includegraphics[width=0.2\linewidth]{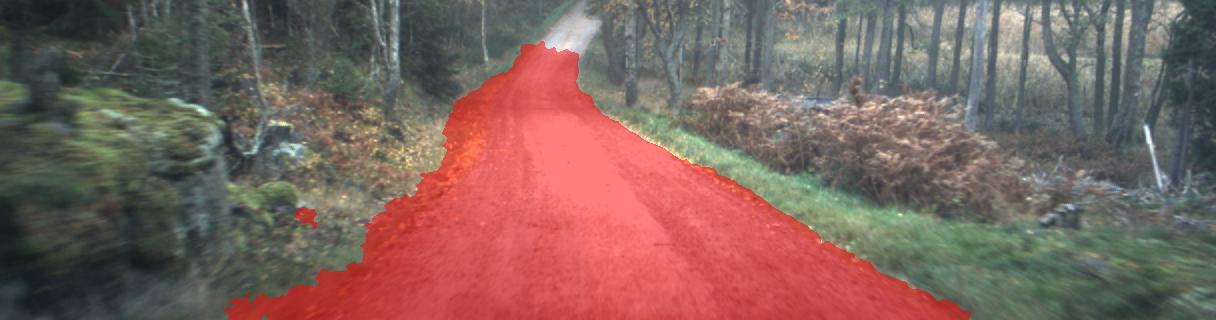} & \includegraphics[width=0.2\linewidth]{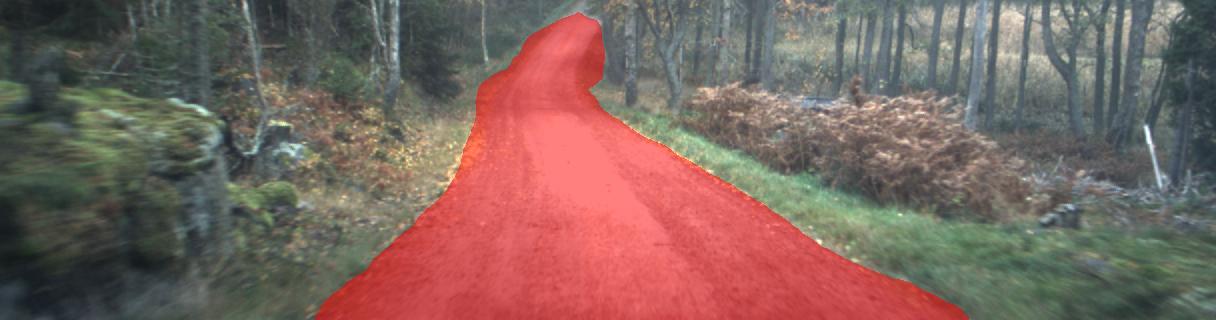} & \includegraphics[width=0.2\linewidth]{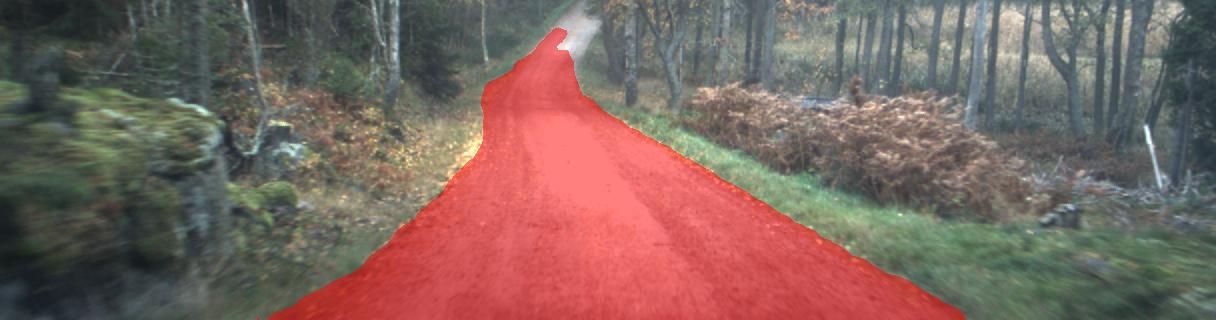} \\
\includegraphics[width=0.2\linewidth]{testi/1547210340470251952.jpg} & \includegraphics[width=0.2\linewidth]{testi/1547210340470251952_gt.jpg} & \includegraphics[width=0.2\linewidth]{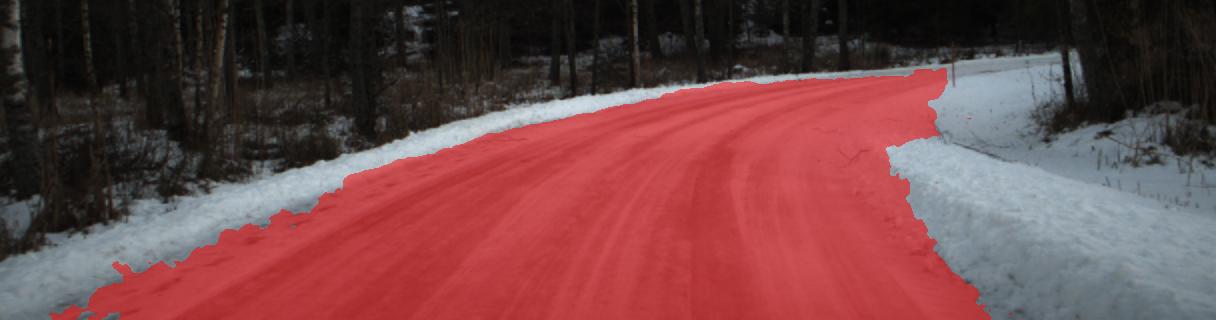} & \includegraphics[width=0.2\linewidth]{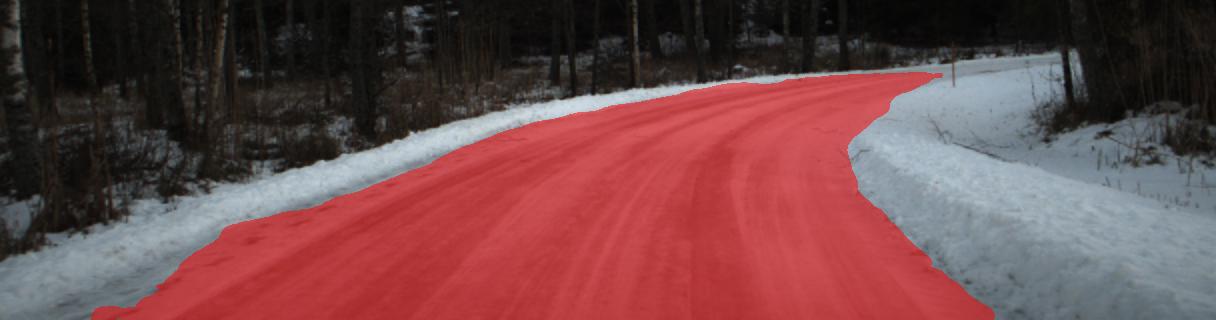} & \includegraphics[width=0.2\linewidth]{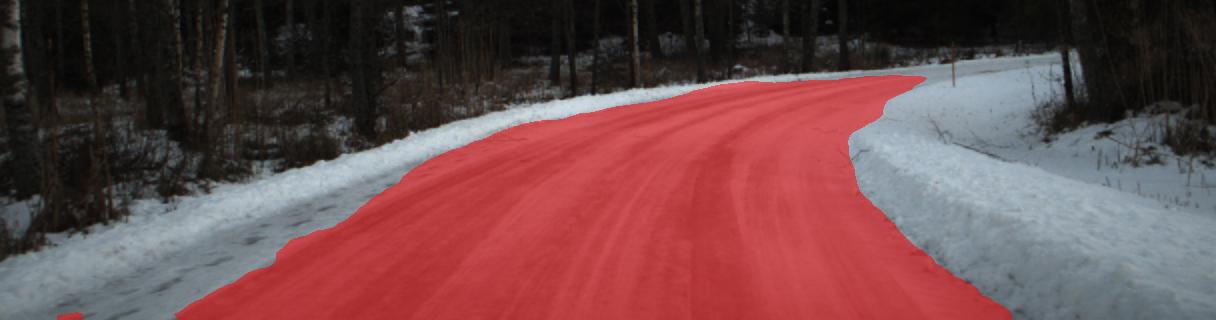} \\
\includegraphics[width=0.2\linewidth]{testi/1547463486843833753.jpg} & \includegraphics[width=0.2\linewidth]{testi/1547463486843833753_gt.jpg} & \includegraphics[width=0.2\linewidth]{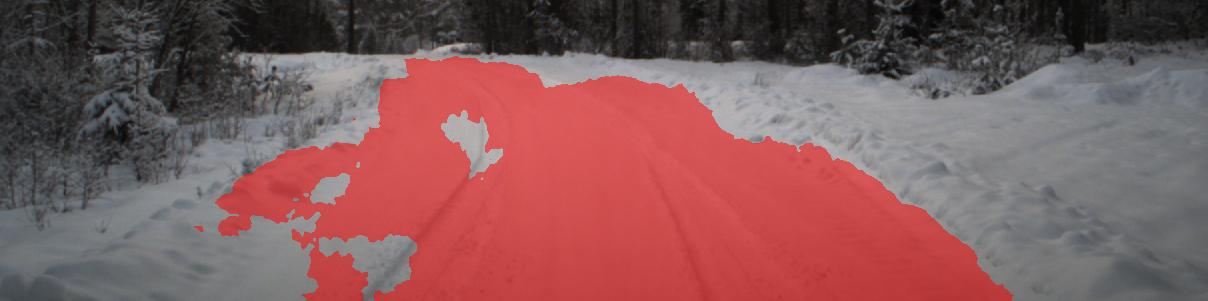} & \includegraphics[width=0.2\linewidth]{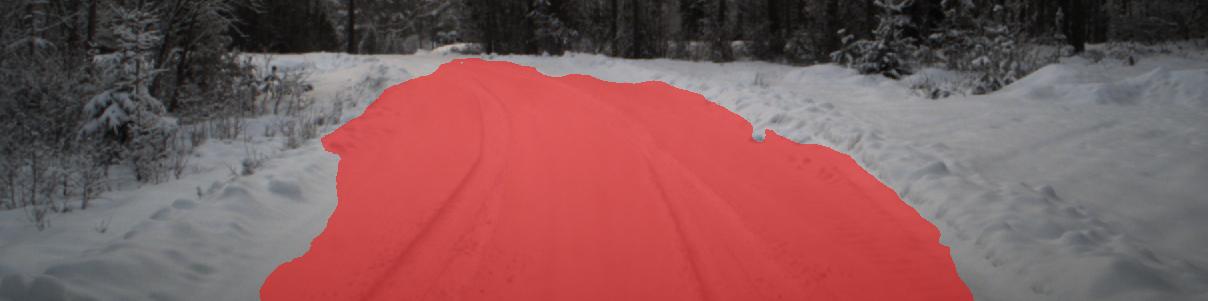} & \includegraphics[width=0.2\linewidth]{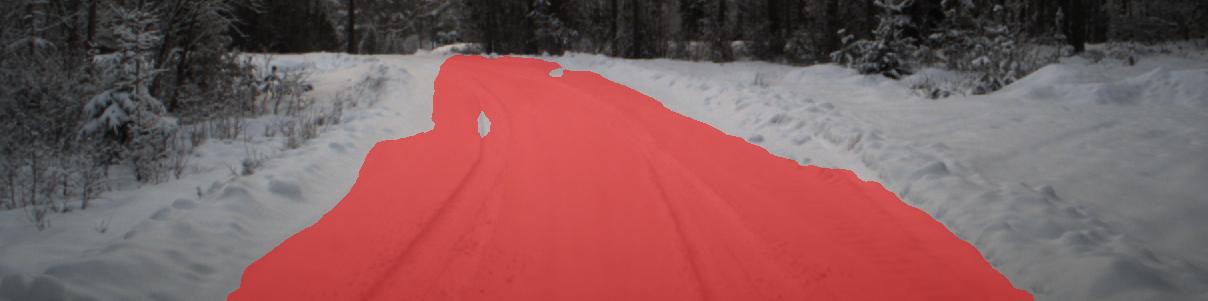} \\
\includegraphics[width=0.2\linewidth]{testi/1551953922753198847.jpg} & \includegraphics[width=0.2\linewidth]{testi/1551953922753198847_gt.jpg} & \includegraphics[width=0.2\linewidth]{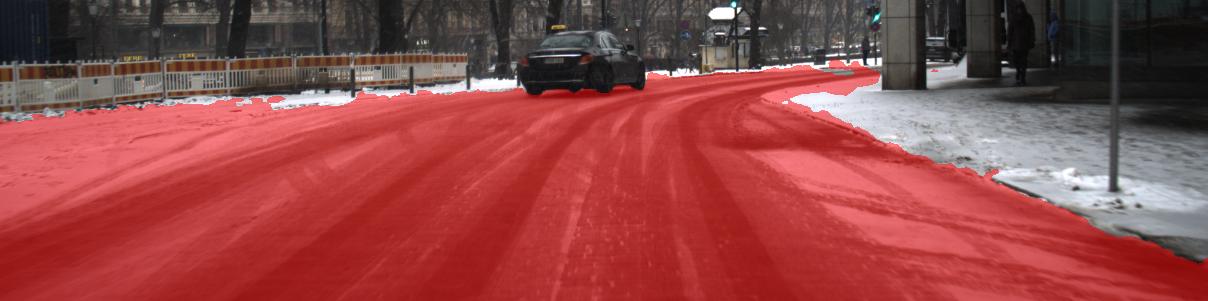} & \includegraphics[width=0.2\linewidth]{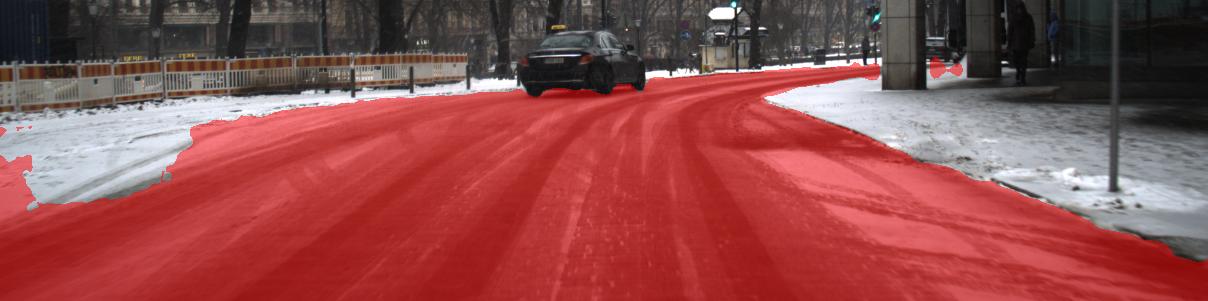} & \includegraphics[width=0.2\linewidth]{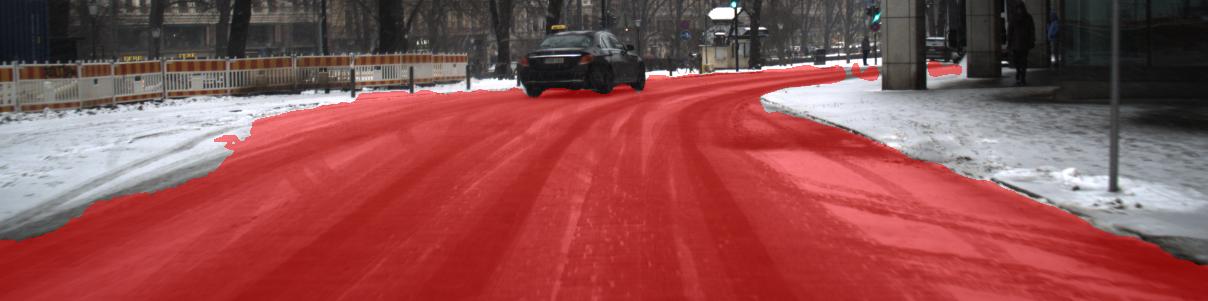} \\
\includegraphics[width=0.2\linewidth]{testi/1554217843481744409.jpg} & \includegraphics[width=0.2\linewidth]{testi/1554217843481744409_gt.jpg} & \includegraphics[width=0.2\linewidth]{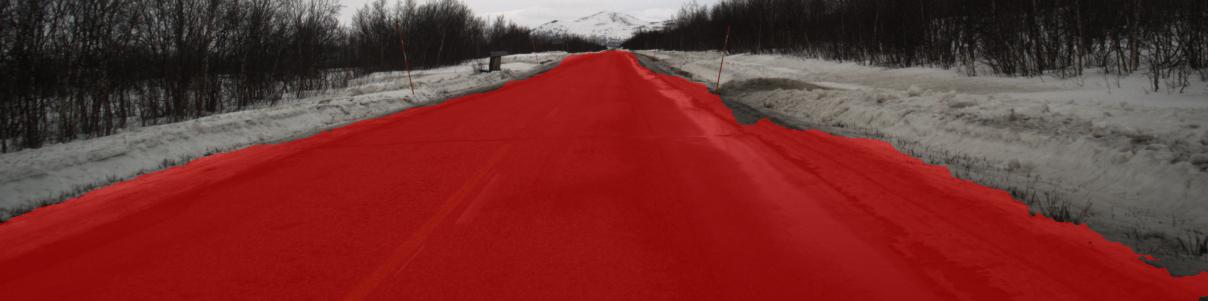} & \includegraphics[width=0.2\linewidth]{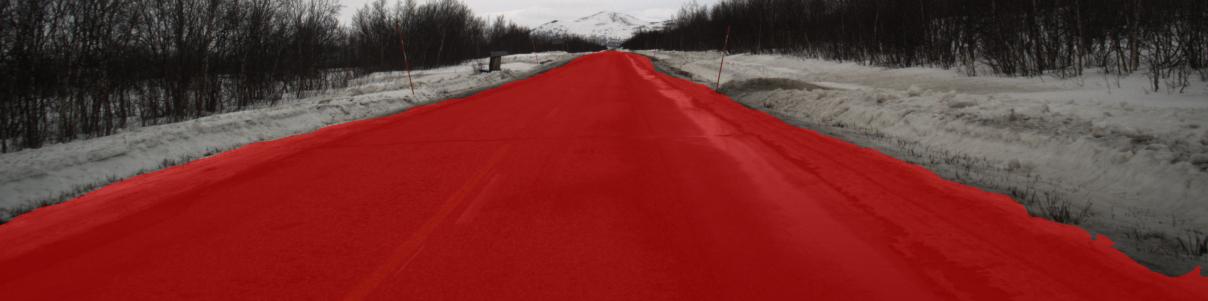} & \includegraphics[width=0.2\linewidth]{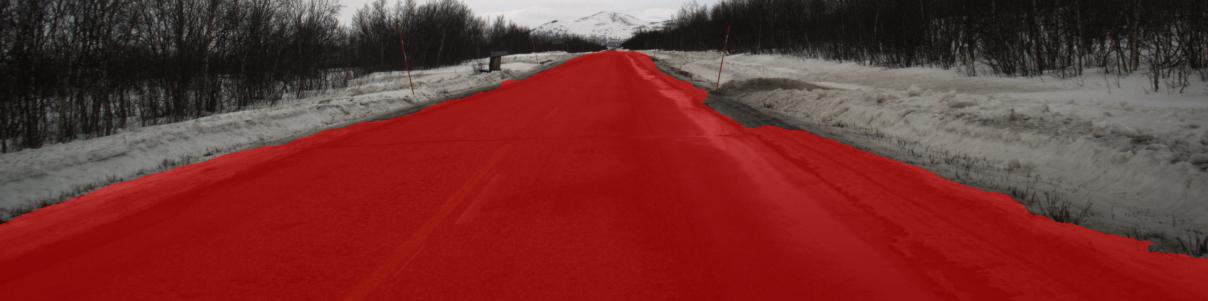} \\
\includegraphics[width=0.2\linewidth]{testi/1554306708725417938.jpg} & \includegraphics[width=0.2\linewidth]{testi/1554306708725417938_gt.jpg} & \includegraphics[width=0.2\linewidth]{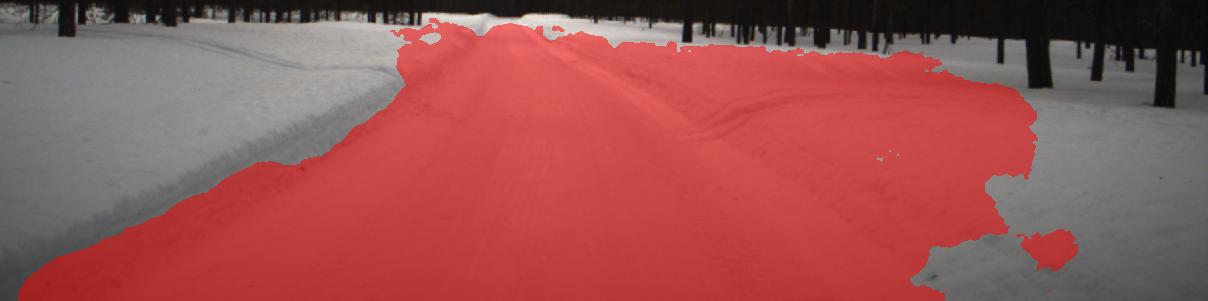} & \includegraphics[width=0.2\linewidth]{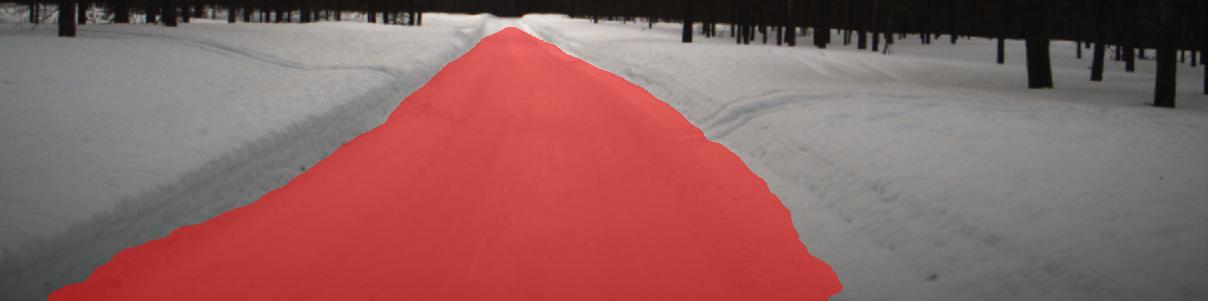} & \includegraphics[width=0.2\linewidth]{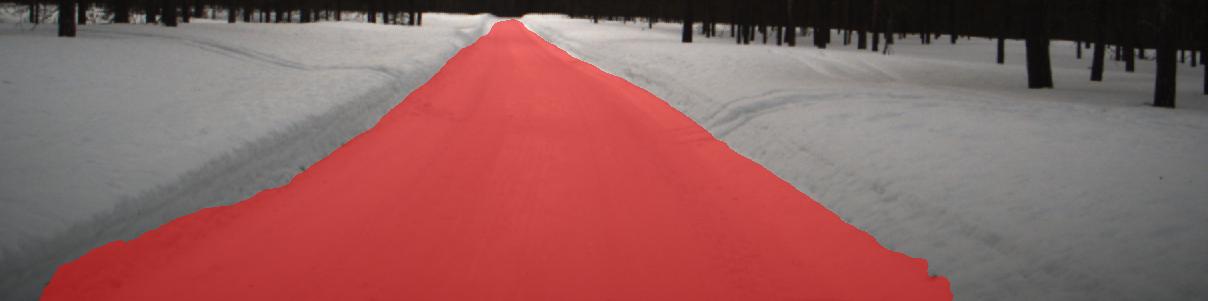} \\
\includegraphics[width=0.2\linewidth]{testi/1554321339770876097.jpg} & \includegraphics[width=0.2\linewidth]{testi/1554321339770876097_gt.jpg} & \includegraphics[width=0.2\linewidth]{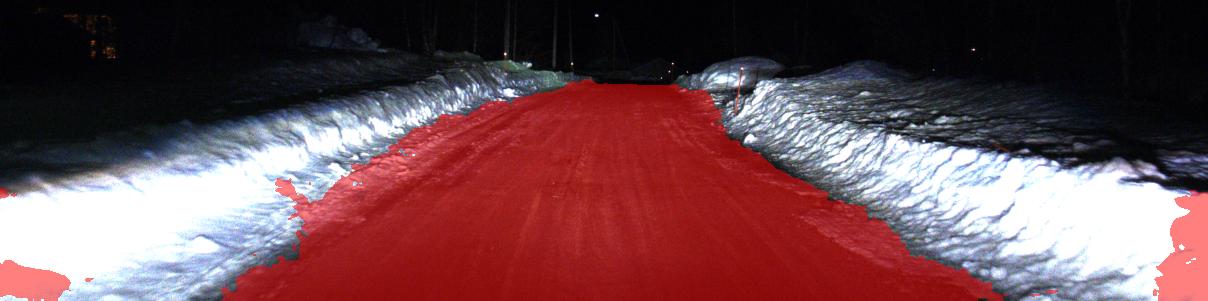} & \includegraphics[width=0.2\linewidth]{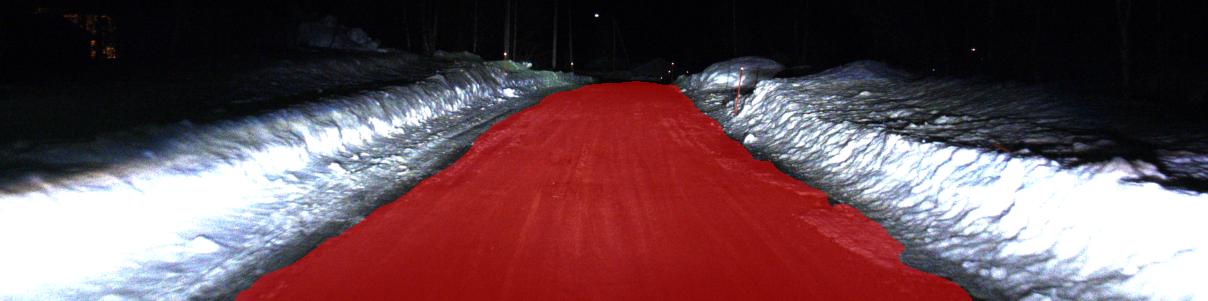} & \includegraphics[width=0.2\linewidth]{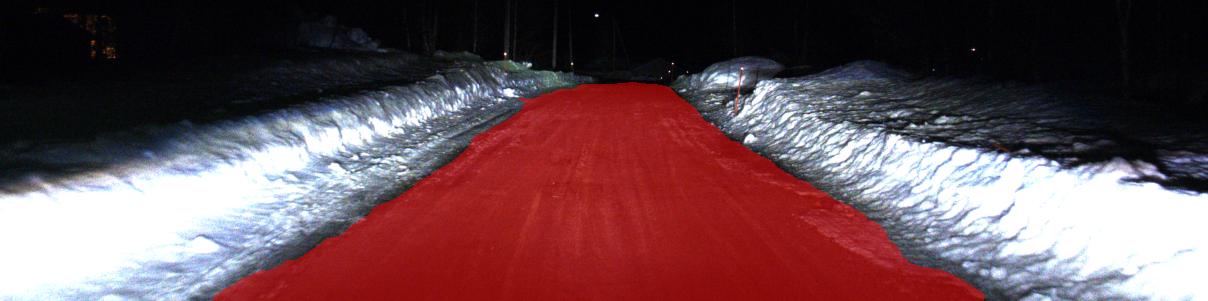} \\
\includegraphics[width=0.2\linewidth]{testi/1554370548031407510.jpg} & \includegraphics[width=0.2\linewidth]{testi/1554370548031407510_gt.jpg} & \includegraphics[width=0.2\linewidth]{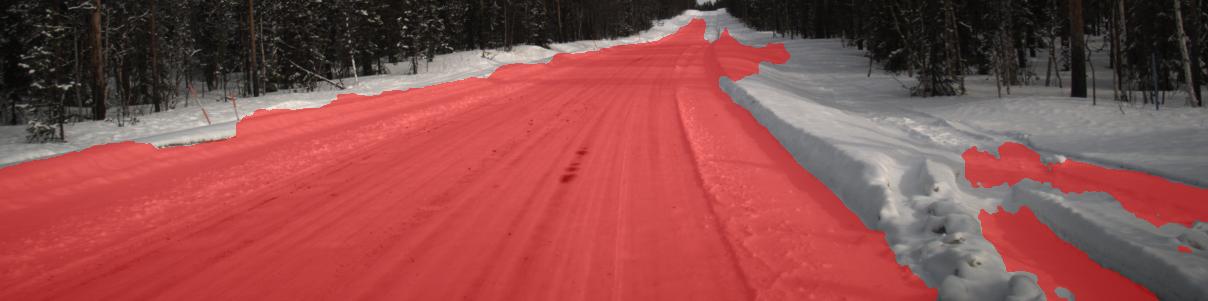} & \includegraphics[width=0.2\linewidth]{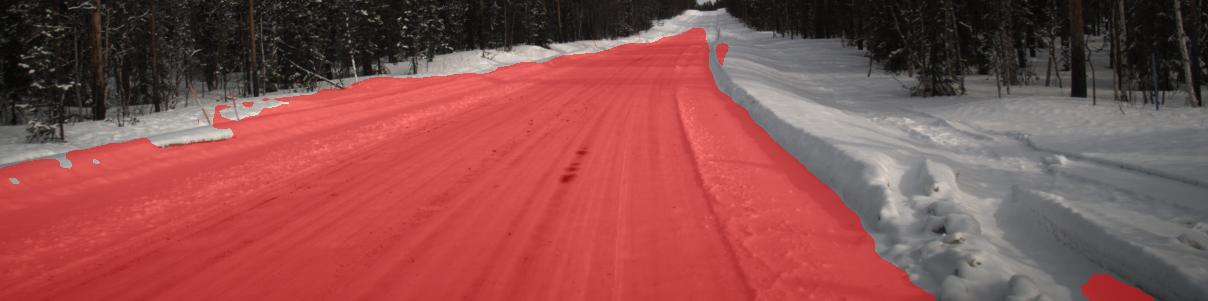} & \includegraphics[width=0.2\linewidth]{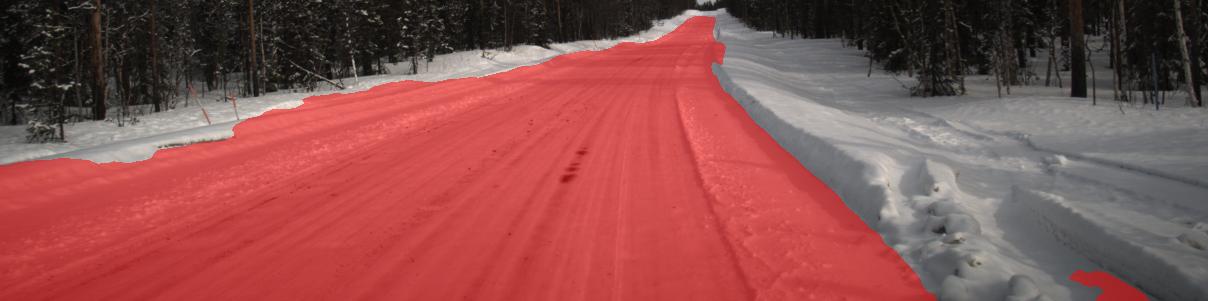} \\
\includegraphics[width=0.2\linewidth]{testi/1554377435123666072.jpg} & \includegraphics[width=0.2\linewidth]{testi/1554377435123666072_gt.jpg} & \includegraphics[width=0.2\linewidth]{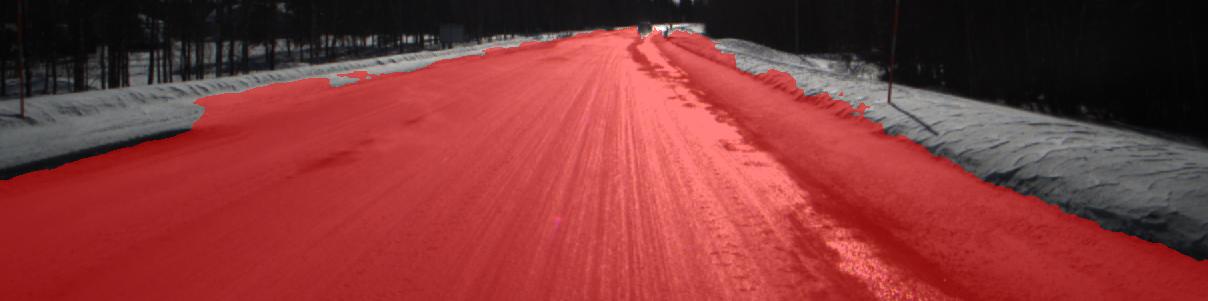} & \includegraphics[width=0.2\linewidth]{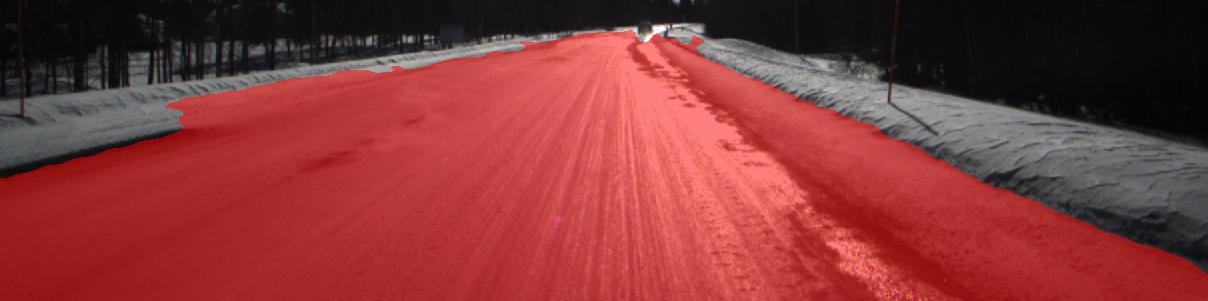} & \includegraphics[width=0.2\linewidth]{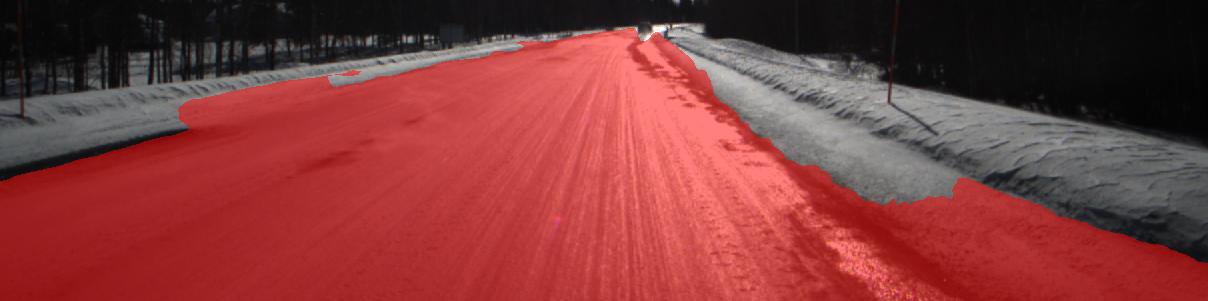} \\
\includegraphics[width=0.2\linewidth]{testi/1555327346406548673.jpg} & \includegraphics[width=0.2\linewidth]{testi/1555327346406548673_gt.jpg} & \includegraphics[width=0.2\linewidth]{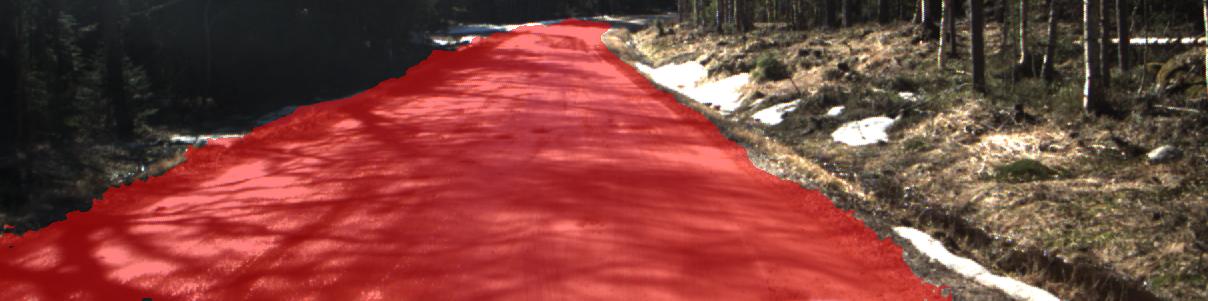} & \includegraphics[width=0.2\linewidth]{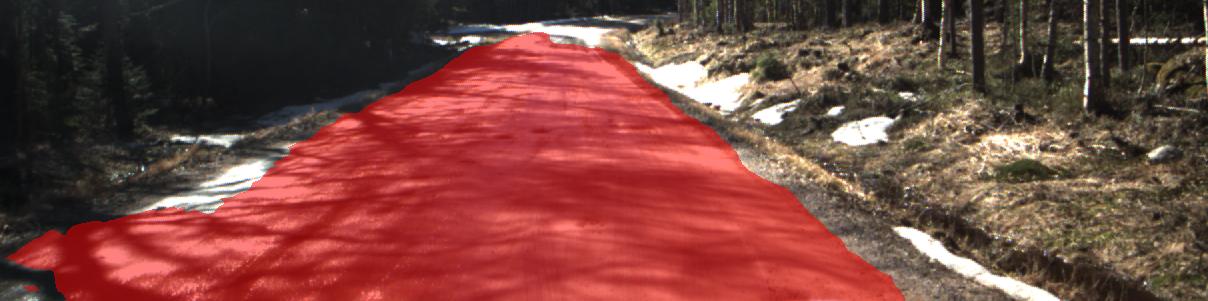} & \includegraphics[width=0.2\linewidth]{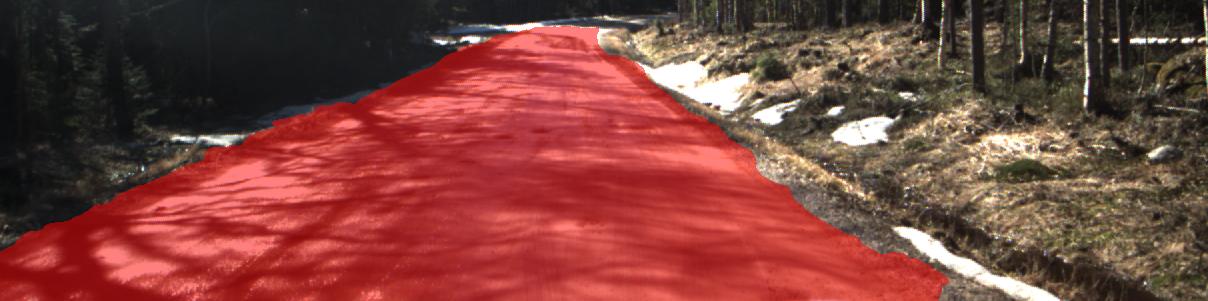}
\end{tabular}
}
\caption{Example road area segmentation masks evaluated with HRNet based models on test set samples.}\label{fig:segmentation_examples}
\end{figure}

Example segmentation masks with the HRNet backbone based models are in~\cref{fig:segmentation_examples}. We observe a bit smoother behavior when compared to the FPN ResNet based models, but no much difference between our multi-task method and the corresponding reference transfer learning model.

\section{Data Set Preprocessing}

The Mapillary data set provides segmentation masks for 124 semantic object categories for 25000 images. As our goal is to use Mapillary data set for training only road area segmentation, we combine classes road, road markings (zebra crossing and general), service lane, plain crosswalk, parking, rail track, manhole and pothole to a single road area class as these classes are drivable area in most of the traffic scenes. After that we remove the samples with less than 5\% of road area from the data set and crop 25\% of the image area from top of each image. In this way the training set images correspond mostly to the FGI autonomous steering data set images that have low height and are centered on the road, which is clearly present in most of the images. Mapillary images are scaled to size $768\times 1024$ and random cropping and scaling are also applied.

The FGI autonomous steering data set contains 28~hours of driving data recorded within different weather conditions on Southern Finland rural and urban roads and on the Western Lapland area. 
The data set was collected with 10~fps sampling frequency to obtain 1,010,000 images from the front-facing center RGB camera and the corresponding steering wheel angles from the drive-by-wire interface. Intersections and stops are excluded from the training dataset. 
The images have size $320\times 1216$ and no cropping is applied due to the steering task which is dependent on the fixed camera position. However, we apply horizontal image flipping in addition to inverting the steering wheel angle accordingly in order to augment the data to the left-lane driving scenario. This is done with half of the training data. 

Color augmentation and small random Gaussian blur are applied to both data sets during training. The road area class is weighted with value~$2.287$ during training to balance between the road and non-road classes. 

\section{Training Details}

\subsection{Training Parameters}

We trained our models with stochastic gradient descent (SGD) with learning rate~\num{2.5e-4}, Nesterov momentum~\num{0.9} and weight decay~\num{5e-4}. 
We trained the model with 100,000 steps each including a minibatch of 16 samples from the Mapillary data set and 32 samples from the FGI autonomous steering data set.  
We used more samples from the FGI autonomous steering data set due to the smaller size of the images and to have more supervision for the steering task, as it has only one scalar value for supervision on each image. 
We chose the model with highest validation accuracy which was evaluated every 1000th step during training. The discriminators are trained with Adam optimizers using learning rate~\num{1e-4} and parameters~$\beta_1=0.9$, $\beta_2=0.99$. 

\subsection{Backbones}

We repeated the experiments with two backbones: High-Resolution Network (HRNet) and Feature Pyramid Network applied to ResNet features (FPN ResNet). 
HRNet is used as the main backbone in model development 
and FPN ResNet is used as a less-accurate baseline to confirm the performance difference between our multi-task model and transfer learning method.
The FPN architecture was used to provide a multi-scale representation based on the feed-forward ResNet features, as the MTI-Net structure is dependent on multi-scale inference. We used only 128 channels in each pyramid layer to reduce GPU memory usage. Both backbones use pre-trained ImageNet weights.

\section{Model Architecture Details}

Here we explain the differences of our model to the original MTI-Net which are not covered in the paper. These mean only the architectures of steering feature segmentation heads and steering heads.

\subsection{Steering Feature Segmentation Head}

The steering feature segmentation heads are implemented as in~\cref{tab:steerseghead}.

\begin{table}[t]
\caption{Steering feature segmentation head.}\label{tab:steerseghead}
\centering
\begin{tabular}{|l|c|c|} \hline
Layer & Output channels & Kernel size \\ \hline
Input & 4 & \\ \hline
Conv2d & 4 & 3x3 \\ \hline
BatchNorm2d & 4 &  \\ \hline
ReLU & 4 &  \\ \hline
Conv2d & 1 & 3x3 \\ \hline
\end{tabular}
\end{table}

\subsection{Steering Heads}

Steering head architectures acting on different steering feature resolutions are in~\cref{tab:steerhead,tab:steerhead0,tab:steerhead1,tab:steerhead2,tab:steerhead3}. This architecture reached the highest validation mIoU in our non-exhaustive experiments for steering head architecture optimization. The steering heads act on different scales in the initial task prediction units of the MTI-Net, also one acts on the final output steering feature that is interpolated to the original image resolution.

\begin{table}[t]
\caption{Steering head in original scale.}\label{tab:steerhead}
\centering
\begin{tabular}{l|ccc|}
\cline{2-4}
                                  & \multicolumn{3}{c|}{SteerHead 1/1}                                                                   \\ \hline
\multicolumn{1}{|l|}{Layer}       & \multicolumn{1}{l|}{Output size}  & \multicolumn{1}{l|}{Kernel size} & \multicolumn{1}{l|}{Stride} \\ \hline
\multicolumn{1}{|l|}{Input}       & \multicolumn{1}{c|}{(4,320,1216)} & \multicolumn{1}{c|}{}            &                             \\ \cline{1-1}
\multicolumn{1}{|l|}{Conv2D}      & \multicolumn{1}{c|}{(64,158,606)} & \multicolumn{1}{c|}{5x5}         & 2                           \\ \cline{1-1}
\multicolumn{1}{|l|}{BatchNorm2D} & \multicolumn{1}{c|}{(64,158,606)} & \multicolumn{1}{c|}{}            &                             \\ \cline{1-1}
\multicolumn{1}{|l|}{ReLU}        & \multicolumn{1}{c|}{(64,158,606)} & \multicolumn{1}{c|}{}            &                             \\ \cline{1-1}
\multicolumn{1}{|l|}{MaxPool2D}   & \multicolumn{1}{c|}{(64,52,202)}  & \multicolumn{1}{c|}{3x3}         &                             \\ \cline{1-1}
\multicolumn{1}{|l|}{Conv2D}      & \multicolumn{1}{c|}{(64,24,99)}   & \multicolumn{1}{c|}{5x5}         & 2                           \\ \cline{1-1}
\multicolumn{1}{|l|}{BatchNorm2D} & \multicolumn{1}{c|}{(64,24,99)}   & \multicolumn{1}{c|}{}            &                             \\ \cline{1-1}
\multicolumn{1}{|l|}{ReLU}        & \multicolumn{1}{c|}{(64,24,99)}   & \multicolumn{1}{c|}{}            &                             \\ \cline{1-1}
\multicolumn{1}{|l|}{MaxPool2D}   & \multicolumn{1}{c|}{(64,12,49)}   & \multicolumn{1}{c|}{2x2}         &                             \\ \cline{1-1}
\multicolumn{1}{|l|}{Conv2D}      & \multicolumn{1}{c|}{(64,10,47)}   & \multicolumn{1}{c|}{3x3}         & 1                           \\ \cline{1-1}
\multicolumn{1}{|l|}{BatchNorm2D} & \multicolumn{1}{c|}{(64,10,47)}   & \multicolumn{1}{c|}{}            &                             \\ \cline{1-1}
\multicolumn{1}{|l|}{ReLU}        & \multicolumn{1}{c|}{(64,10,47)}   & \multicolumn{1}{c|}{}            &                             \\ \cline{1-1}
\multicolumn{1}{|l|}{MaxPool2D}   & \multicolumn{1}{c|}{(64,5,23)}    & \multicolumn{1}{c|}{2x2}         &                             \\ \cline{1-1}
\multicolumn{1}{|l|}{Conv2D}      & \multicolumn{1}{c|}{(64,3,21)}    & \multicolumn{1}{c|}{3x3}         & 1                           \\ \cline{1-1}
\multicolumn{1}{|l|}{BatchNorm2D} & \multicolumn{1}{c|}{(64,3,21)}    & \multicolumn{1}{c|}{}            &                             \\ \cline{1-1}
\multicolumn{1}{|l|}{ReLU}        & \multicolumn{1}{c|}{(64,3,21)}    & \multicolumn{1}{c|}{}            &                             \\ \cline{1-1}
\multicolumn{1}{|l|}{Conv2D}      & \multicolumn{1}{c|}{(64,1,19)}    & \multicolumn{1}{c|}{3x3}         & 1                           \\ \cline{1-1}
\multicolumn{1}{|l|}{BatchNorm2D} & \multicolumn{1}{c|}{(64,1,19)}    & \multicolumn{1}{l|}{}            & \multicolumn{1}{l|}{}       \\ \cline{1-1}
\multicolumn{1}{|l|}{ReLU}        & \multicolumn{1}{c|}{(64,1,19)}    & \multicolumn{1}{l|}{}            & \multicolumn{1}{l|}{}       \\ \cline{1-1}
\multicolumn{1}{|l|}{Flatten}     & \multicolumn{1}{c|}{1216}         & \multicolumn{1}{l|}{}            & \multicolumn{1}{l|}{}       \\ \cline{1-1}
\multicolumn{1}{|l|}{Linear}      & \multicolumn{1}{c|}{25}           & \multicolumn{1}{l|}{}            & \multicolumn{1}{l|}{}       \\ \cline{1-1}
\multicolumn{1}{|l|}{ReLU}        & \multicolumn{1}{c|}{25}           & \multicolumn{1}{l|}{}            & \multicolumn{1}{l|}{}       \\ \cline{1-1}
\multicolumn{1}{|l|}{Linear}      & \multicolumn{1}{c|}{1}            & \multicolumn{1}{l|}{}            & \multicolumn{1}{l|}{}       \\ \hline
\end{tabular}
\end{table}

\begin{table}[h]
\caption{Steering head in 1/4 scale.}\label{tab:steerhead0}
\centering
\begin{tabular}{l|ccc|}
\cline{2-4}
                                  & \multicolumn{3}{c|}{SteerHead 1/4}                                                                   \\ \hline
\multicolumn{1}{|l|}{Layer}       & \multicolumn{1}{l|}{Output size}  & \multicolumn{1}{l|}{Kernel size} & \multicolumn{1}{l|}{Stride} \\ \hline
\multicolumn{1}{|l|}{Input}       & \multicolumn{1}{c|}{(4,80,304)} & \multicolumn{1}{c|}{}            &                             \\ \cline{1-1}
\multicolumn{1}{|l|}{Conv2D}      & \multicolumn{1}{c|}{(64,38,150)} & \multicolumn{1}{c|}{5x5}         & 2                           \\ \cline{1-1}
\multicolumn{1}{|l|}{BatchNorm2D} & \multicolumn{1}{c|}{(64,38,150)} & \multicolumn{1}{c|}{}            &                             \\ \cline{1-1}
\multicolumn{1}{|l|}{ReLU}        & \multicolumn{1}{c|}{(64,38,150)} & \multicolumn{1}{c|}{}            &                             \\ \cline{1-1}
\multicolumn{1}{|l|}{MaxPool2D}   & \multicolumn{1}{c|}{(64,19,75)}  & \multicolumn{1}{c|}{2x2}         &                             \\ \cline{1-1}
\multicolumn{1}{|l|}{Conv2D}      & \multicolumn{1}{c|}{(64,8,36)}   & \multicolumn{1}{c|}{5x5}         & 2                           \\ \cline{1-1}
\multicolumn{1}{|l|}{BatchNorm2D} & \multicolumn{1}{c|}{(64,8,36)}   & \multicolumn{1}{c|}{}            &                             \\ \cline{1-1}
\multicolumn{1}{|l|}{ReLU}        & \multicolumn{1}{c|}{(64,8,36)}   & \multicolumn{1}{c|}{}            &                             \\ \cline{1-1}
\multicolumn{1}{|l|}{MaxPool2D}   & \multicolumn{1}{c|}{(64,4,18)}   & \multicolumn{1}{c|}{2x2}         &                             \\ \cline{1-1}
\multicolumn{1}{|l|}{Conv2D}      & \multicolumn{1}{c|}{(64,2,16)}   & \multicolumn{1}{c|}{3x3}         & 1                           \\ \cline{1-1}
\multicolumn{1}{|l|}{BatchNorm2D} & \multicolumn{1}{c|}{(64,2,16)}   & \multicolumn{1}{c|}{}            &                             \\ \cline{1-1}
\multicolumn{1}{|l|}{ReLU}        & \multicolumn{1}{c|}{(64,2,16)}   & \multicolumn{1}{c|}{}            &                             \\ \cline{1-1}
\multicolumn{1}{|l|}{Conv2D}      & \multicolumn{1}{c|}{(64,1,14)}    & \multicolumn{1}{c|}{2x3}         & 1                           \\ \cline{1-1}
\multicolumn{1}{|l|}{BatchNorm2D} & \multicolumn{1}{c|}{(64,1,14)}    & \multicolumn{1}{c|}{}            &                             \\ \cline{1-1}
\multicolumn{1}{|l|}{ReLU}        & \multicolumn{1}{c|}{(64,1,14)}    & \multicolumn{1}{c|}{}            &                             \\ \cline{1-1}
\multicolumn{1}{|l|}{Flatten}     & \multicolumn{1}{c|}{869}         & \multicolumn{1}{l|}{}            & \multicolumn{1}{l|}{}       \\ \cline{1-1}
\multicolumn{1}{|l|}{Linear}      & \multicolumn{1}{c|}{25}           & \multicolumn{1}{l|}{}            & \multicolumn{1}{l|}{}       \\ \cline{1-1}
\multicolumn{1}{|l|}{ReLU}        & \multicolumn{1}{c|}{25}           & \multicolumn{1}{l|}{}            & \multicolumn{1}{l|}{}       \\ \cline{1-1}
\multicolumn{1}{|l|}{Linear}      & \multicolumn{1}{c|}{1}            & \multicolumn{1}{l|}{}            & \multicolumn{1}{l|}{}       \\ \hline
\end{tabular}
\end{table}

\begin{table}[h]
\caption{Steering head in 1/8 scale.}\label{tab:steerhead1}
\centering
\begin{tabular}{l|ccc|}
\cline{2-4}
                                  & \multicolumn{3}{c|}{SteerHead 1/8}                                                                   \\ \hline
\multicolumn{1}{|l|}{Layer}       & \multicolumn{1}{l|}{Output size}  & \multicolumn{1}{l|}{Kernel size} & \multicolumn{1}{l|}{Stride} \\ \hline
\multicolumn{1}{|l|}{Input}       & \multicolumn{1}{c|}{(4,40,152)} & \multicolumn{1}{c|}{}            &                             \\ \cline{1-1}
\multicolumn{1}{|l|}{Conv2D}      & \multicolumn{1}{c|}{(64,18,74)} & \multicolumn{1}{c|}{5x5}         & 2                           \\ \cline{1-1}
\multicolumn{1}{|l|}{BatchNorm2D} & \multicolumn{1}{c|}{(64,18,74)} & \multicolumn{1}{c|}{}            &                             \\ \cline{1-1}
\multicolumn{1}{|l|}{ReLU}        & \multicolumn{1}{c|}{(64,18,74)} & \multicolumn{1}{c|}{}            &                             \\ \cline{1-1}
\multicolumn{1}{|l|}{MaxPool2D}   & \multicolumn{1}{c|}{(64,9,24)}  & \multicolumn{1}{c|}{2x3}         &                             \\ \cline{1-1}
\multicolumn{1}{|l|}{Conv2D}      & \multicolumn{1}{c|}{(64,7,22)}   & \multicolumn{1}{c|}{3x3}         & 1                           \\ \cline{1-1}
\multicolumn{1}{|l|}{BatchNorm2D} & \multicolumn{1}{c|}{(64,7,22)}   & \multicolumn{1}{c|}{}            &                             \\ \cline{1-1}
\multicolumn{1}{|l|}{ReLU}        & \multicolumn{1}{c|}{(64,7,22)}   & \multicolumn{1}{c|}{}            &                             \\ \cline{1-1}
\multicolumn{1}{|l|}{Conv2D}      & \multicolumn{1}{c|}{(64,5,20)}   & \multicolumn{1}{c|}{3x3}         & 1                           \\ \cline{1-1}
\multicolumn{1}{|l|}{BatchNorm2D} & \multicolumn{1}{c|}{(64,5,20)}   & \multicolumn{1}{c|}{}            &                             \\ \cline{1-1}
\multicolumn{1}{|l|}{ReLU}        & \multicolumn{1}{c|}{(64,5,20)}   & \multicolumn{1}{c|}{}            &                             \\ \cline{1-1}
\multicolumn{1}{|l|}{Conv2D}      & \multicolumn{1}{c|}{(64,3,18)}    & \multicolumn{1}{c|}{3x3}         & 1                           \\ \cline{1-1}
\multicolumn{1}{|l|}{BatchNorm2D} & \multicolumn{1}{c|}{(64,3,18)}    & \multicolumn{1}{c|}{}            &                             \\ \cline{1-1}
\multicolumn{1}{|l|}{ReLU}        & \multicolumn{1}{c|}{(64,3,18)}    & \multicolumn{1}{c|}{}            &                             \\ \cline{1-1}
\multicolumn{1}{|l|}{Conv2D}      & \multicolumn{1}{c|}{(64,1,16)}    & \multicolumn{1}{c|}{3x3}         & 1                           \\ \cline{1-1}
\multicolumn{1}{|l|}{BatchNorm2D} & \multicolumn{1}{c|}{(64,1,16)}    & \multicolumn{1}{l|}{}            & \multicolumn{1}{l|}{}       \\ \cline{1-1}
\multicolumn{1}{|l|}{ReLU}        & \multicolumn{1}{c|}{(64,1,16)}    & \multicolumn{1}{l|}{}            & \multicolumn{1}{l|}{}       \\ \cline{1-1}
\multicolumn{1}{|l|}{Flatten}     & \multicolumn{1}{c|}{1024}         & \multicolumn{1}{l|}{}            & \multicolumn{1}{l|}{}       \\ \cline{1-1}
\multicolumn{1}{|l|}{Linear}      & \multicolumn{1}{c|}{25}           & \multicolumn{1}{l|}{}            & \multicolumn{1}{l|}{}       \\ \cline{1-1}
\multicolumn{1}{|l|}{ReLU}        & \multicolumn{1}{c|}{25}           & \multicolumn{1}{l|}{}            & \multicolumn{1}{l|}{}       \\ \cline{1-1}
\multicolumn{1}{|l|}{Linear}      & \multicolumn{1}{c|}{1}            & \multicolumn{1}{l|}{}            & \multicolumn{1}{l|}{}       \\ \hline
\end{tabular}
\end{table}

\begin{table}[h]
\caption{Steering head in 1/16 scale.}\label{tab:steerhead2}
\centering
\begin{tabular}{l|ccc|}
\cline{2-4}
                                  & \multicolumn{3}{c|}{SteerHead 1/16}                                                                   \\ \hline
\multicolumn{1}{|l|}{Layer}       & \multicolumn{1}{l|}{Output size}  & \multicolumn{1}{l|}{Kernel size} & \multicolumn{1}{l|}{Stride} \\ \hline
\multicolumn{1}{|l|}{Input}       & \multicolumn{1}{c|}{(4,20,76)} & \multicolumn{1}{c|}{}            &                             \\ \cline{1-1}
\multicolumn{1}{|l|}{Conv2D}      & \multicolumn{1}{c|}{(64,8,36)} & \multicolumn{1}{c|}{5x5}         & 2                           \\ \cline{1-1}
\multicolumn{1}{|l|}{BatchNorm2D} & \multicolumn{1}{c|}{(64,8,36)} & \multicolumn{1}{c|}{}            &                             \\ \cline{1-1}
\multicolumn{1}{|l|}{ReLU}        & \multicolumn{1}{c|}{(64,8,36)} & \multicolumn{1}{c|}{}            &                             \\ \cline{1-1}
\multicolumn{1}{|l|}{Conv2D}      & \multicolumn{1}{c|}{(64,6,34)}   & \multicolumn{1}{c|}{3x3}         & 1                           \\ \cline{1-1}
\multicolumn{1}{|l|}{BatchNorm2D} & \multicolumn{1}{c|}{(64,6,34)}   & \multicolumn{1}{c|}{}            &                             \\ \cline{1-1}
\multicolumn{1}{|l|}{ReLU}        & \multicolumn{1}{c|}{(64,6,34)}   & \multicolumn{1}{c|}{}            &                             \\ \cline{1-1}
\multicolumn{1}{|l|}{Conv2D}      & \multicolumn{1}{c|}{(64,4,32)}   & \multicolumn{1}{c|}{3x3}         & 1                           \\ \cline{1-1}
\multicolumn{1}{|l|}{BatchNorm2D} & \multicolumn{1}{c|}{(64,4,32)}   & \multicolumn{1}{c|}{}            &                             \\ \cline{1-1}
\multicolumn{1}{|l|}{ReLU}        & \multicolumn{1}{c|}{(64,4,32)}   & \multicolumn{1}{c|}{}            &                             \\ \cline{1-1}
\multicolumn{1}{|l|}{Conv2D}      & \multicolumn{1}{c|}{(64,2,30)}    & \multicolumn{1}{c|}{3x3}         & 1                           \\ \cline{1-1}
\multicolumn{1}{|l|}{BatchNorm2D} & \multicolumn{1}{c|}{(64,2,30)}    & \multicolumn{1}{c|}{}            &                             \\ \cline{1-1}
\multicolumn{1}{|l|}{ReLU}        & \multicolumn{1}{c|}{(64,2,30)}    & \multicolumn{1}{c|}{}            &                             \\ \cline{1-1}
\multicolumn{1}{|l|}{Conv2D}      & \multicolumn{1}{c|}{(64,1,28)}    & \multicolumn{1}{c|}{2x3}         & 1                           \\ \cline{1-1}
\multicolumn{1}{|l|}{BatchNorm2D} & \multicolumn{1}{c|}{(64,1,28)}    & \multicolumn{1}{l|}{}            & \multicolumn{1}{l|}{}       \\ \cline{1-1}
\multicolumn{1}{|l|}{ReLU}        & \multicolumn{1}{c|}{(64,1,28)}    & \multicolumn{1}{l|}{}            & \multicolumn{1}{l|}{}       \\ \cline{1-1}
\multicolumn{1}{|l|}{Flatten}     & \multicolumn{1}{c|}{1792}         & \multicolumn{1}{l|}{}            & \multicolumn{1}{l|}{}       \\ \cline{1-1}
\multicolumn{1}{|l|}{Linear}      & \multicolumn{1}{c|}{25}           & \multicolumn{1}{l|}{}            & \multicolumn{1}{l|}{}       \\ \cline{1-1}
\multicolumn{1}{|l|}{ReLU}        & \multicolumn{1}{c|}{25}           & \multicolumn{1}{l|}{}            & \multicolumn{1}{l|}{}       \\ \cline{1-1}
\multicolumn{1}{|l|}{Linear}      & \multicolumn{1}{c|}{1}            & \multicolumn{1}{l|}{}            & \multicolumn{1}{l|}{}       \\ \hline
\end{tabular}
\end{table}

\begin{table}[h]
\caption{Steering head in 1/32 scale.}\label{tab:steerhead3}
\centering
\begin{tabular}{l|ccc|}
\cline{2-4}
                                  & \multicolumn{3}{c|}{SteerHead 1/32}                                                                   \\ \hline
\multicolumn{1}{|l|}{Layer}       & \multicolumn{1}{l|}{Output size}  & \multicolumn{1}{l|}{Kernel size} & \multicolumn{1}{l|}{Stride} \\ \hline
\multicolumn{1}{|l|}{Input}       & \multicolumn{1}{c|}{(4,10,38)} & \multicolumn{1}{c|}{}            &                             \\ \cline{1-1}
\multicolumn{1}{|l|}{Conv2D}      & \multicolumn{1}{c|}{(64,8,36)} & \multicolumn{1}{c|}{3x3}         & 1                           \\ \cline{1-1}
\multicolumn{1}{|l|}{BatchNorm2D} & \multicolumn{1}{c|}{(64,8,36)} & \multicolumn{1}{c|}{}            &                             \\ \cline{1-1}
\multicolumn{1}{|l|}{ReLU}        & \multicolumn{1}{c|}{(64,8,36)} & \multicolumn{1}{c|}{}            &                             \\ \cline{1-1}
\multicolumn{1}{|l|}{Conv2D}      & \multicolumn{1}{c|}{(64,6,34)}   & \multicolumn{1}{c|}{3x3}         & 1                           \\ \cline{1-1}
\multicolumn{1}{|l|}{BatchNorm2D} & \multicolumn{1}{c|}{(64,6,34)}   & \multicolumn{1}{c|}{}            &                             \\ \cline{1-1}
\multicolumn{1}{|l|}{ReLU}        & \multicolumn{1}{c|}{(64,6,34)}   & \multicolumn{1}{c|}{}            &                             \\ \cline{1-1}
\multicolumn{1}{|l|}{Conv2D}      & \multicolumn{1}{c|}{(64,4,32)}   & \multicolumn{1}{c|}{3x3}         & 1                           \\ \cline{1-1}
\multicolumn{1}{|l|}{BatchNorm2D} & \multicolumn{1}{c|}{(64,4,32)}   & \multicolumn{1}{c|}{}            &                             \\ \cline{1-1}
\multicolumn{1}{|l|}{ReLU}        & \multicolumn{1}{c|}{(64,4,32)}   & \multicolumn{1}{c|}{}            &                             \\ \cline{1-1}
\multicolumn{1}{|l|}{Conv2D}      & \multicolumn{1}{c|}{(64,2,30)}    & \multicolumn{1}{c|}{3x3}         & 1                           \\ \cline{1-1}
\multicolumn{1}{|l|}{BatchNorm2D} & \multicolumn{1}{c|}{(64,2,30)}    & \multicolumn{1}{c|}{}            &                             \\ \cline{1-1}
\multicolumn{1}{|l|}{ReLU}        & \multicolumn{1}{c|}{(64,2,30)}    & \multicolumn{1}{c|}{}            &                             \\ \cline{1-1}
\multicolumn{1}{|l|}{Conv2D}      & \multicolumn{1}{c|}{(64,1,28)}    & \multicolumn{1}{c|}{2x3}         & 1                           \\ \cline{1-1}
\multicolumn{1}{|l|}{BatchNorm2D} & \multicolumn{1}{c|}{(64,1,28)}    & \multicolumn{1}{l|}{}            & \multicolumn{1}{l|}{}       \\ \cline{1-1}
\multicolumn{1}{|l|}{ReLU}        & \multicolumn{1}{c|}{(64,1,28)}    & \multicolumn{1}{l|}{}            & \multicolumn{1}{l|}{}       \\ \cline{1-1}
\multicolumn{1}{|l|}{Flatten}     & \multicolumn{1}{c|}{1792}         & \multicolumn{1}{l|}{}            & \multicolumn{1}{l|}{}       \\ \cline{1-1}
\multicolumn{1}{|l|}{Linear}      & \multicolumn{1}{c|}{25}           & \multicolumn{1}{l|}{}            & \multicolumn{1}{l|}{}       \\ \cline{1-1}
\multicolumn{1}{|l|}{ReLU}        & \multicolumn{1}{c|}{25}           & \multicolumn{1}{l|}{}            & \multicolumn{1}{l|}{}       \\ \cline{1-1}
\multicolumn{1}{|l|}{Linear}      & \multicolumn{1}{c|}{1}            & \multicolumn{1}{l|}{}            & \multicolumn{1}{l|}{}       \\ \hline
\end{tabular}
\end{table}